\documentclass{article}

\usepackage[OT1]{fontenc}
\usepackage{iclr2027_conference,times}

\usepackage{amsmath,amssymb,amsfonts,bm,mathtools}
\usepackage{microtype}
\usepackage{booktabs}
\usepackage{multirow}
\usepackage{graphicx}
\usepackage{algorithm}
\usepackage{algpseudocode}
\usepackage{hyperref}
\usepackage{url}
\usepackage{xcolor}
\hypersetup{colorlinks=true,linkcolor=blue,citecolor=blue,urlcolor=blue}

\newcommand{\R}{\mathbb{R}}
\newcommand{\E}{\mathbb{E}}

\newcommand{\N}{\mathcal{N}}
\newcommand{\dd}{\mathrm{d}}
\newcommand{\I}{\mathbf{I}}
\newcommand{\norm}[1]{\left\lVert #1 \right\rVert}

\newcommand{\given}{\,|\,}

\newcommand{\diag}{\operatorname{diag}}

\title{AECSF: Adaptive Ensemble Conditional Score Filtering for High-Dimensional Nonlinear Data Assimilation}

\author{%
\textbf{Yangwen Zhang\textsuperscript{1}, Shiwei Ni\textsuperscript{1}, Xiaoping Zhang\textsuperscript{2}, Xiaofei Guan\textsuperscript{1}, and Lili Ju\textsuperscript{3}}\\[0.7em]
\small
\textsuperscript{1}School of Mathematical Sciences, Tongji University, Shanghai 200092, China\\
\textsuperscript{2}School of Mathematics and Statistics, Wuhan University, Wuhan, 430072, Hubei, China\\
\textsuperscript{3}Department of Mathematics, University of South Carolina, Columbia, South Carolina 29208, USA%
}
\date{}

\iclrfinalcopy

\begin{document}

\maketitle
\lhead{}
\begin{center}
\small
\textbf{Correspondence:} Xiaofei Guan (\href{mailto:guanxf@tongji.edu.cn}{guanxf@tongji.edu.cn})\par
\textbf{Author emails:} Yangwen Zhang (\href{mailto:2311747@tongji.edu.cn}{2311747@tongji.edu.cn}); Shiwei Ni (\href{mailto:2543480712@qq.com}{2543480712@qq.com}).\par
Xiaoping Zhang (\href{mailto:xpzhang.math@whu.edu.cn}{xpzhang.math@whu.edu.cn}); Lili Ju (\href{mailto:ju@math.sc.edu}{ju@math.sc.edu}).
\end{center}

\begin{abstract}
In high-dimensional filters, particle weights can concentrate on a few members, while Gaussian updates can miss non-Gaussian posterior structure. A diffusion filter updates a forecast ensemble by gradually denoising particles in light of a new observation. Each noisy particle can correspond to several system states, and these states can have very different likelihoods. A score correction based on a single state can therefore misrepresent the posterior. We propose the Adaptive Ensemble Conditional Score Filter (AECSF), which uses the conditional Tweedie identity to express the required score through the mean state given the particle and the observation. All reverse particles share a set of proposals, but each assigns its own weights to estimate its conditional mean. As denoising proceeds, the reverse particles guide which proposals are resampled. We characterize when fixed proposals yield the exact noisy posterior score and bound the contribution of conditional-mean errors to the final sampling error. In controlled Gaussian experiments, AECSF samples the posterior more accurately than the tested likelihood-guidance rules. With 20 forecast members in 10,000-dimensional Lorenz--96, AECSF has lower RMSE and CRPS than the tested tempered SMC and kernel-density filters. On Kuramoto--Sivashinsky, the analysis ensembles also improve subsequent forecasts over the tested baselines at comparable cost. For Navier--Stokes flow, we learn a forecast prior that captures spatial dependence in vorticity and use it to reconstruct the full field from sparse velocity measurements.
\end{abstract}

\section{Introduction}
\label{sec:introduction}

Reliable prediction of complex dynamical systems depends on accurate state estimation and uncertainty quantification. Data assimilation estimates the evolving system state by combining model forecasts with noisy observations \citep{reich2015probabilistic,sanzalonso2023data}. In Bayesian filtering, each new observation updates the forecast distribution to a posterior distribution representing the remaining uncertainty about the system state. An ensemble provides a practical representation of these distributions and carries this uncertainty into subsequent forecasts. For high-dimensional nonlinear systems, however, a limited ensemble may poorly represent the posterior, while increasing its size raises the cost of propagating and updating its members. The challenge is therefore to improve posterior inference at a computational cost compatible with repeated assimilation.

Ensemble Kalman and particle methods address this challenge in different ways. Ensemble Kalman methods use sample covariances for efficient updates, but Gaussian approximations and linear regression can miss non-Gaussian posterior structure \citep{evensen1994sequential,hunt2007efficient}. Particle methods allow more general distributions, but importance weights can concentrate on few members \citep{liu1998sequential,doucet2001sequential}, especially in high dimensions \citep{snyder2008obstacles,bengtsson2008curse}. These limitations motivate alternative updates that can represent non-Gaussian posteriors at practical ensemble sizes.

Score-based generative methods offer another way to construct posterior ensembles. They generate samples through reverse diffusion, which progressively transforms random noise into samples of a target distribution \citep{song2021scorebased}. This transformation is driven by score functions, the gradients of log densities at successive noise levels. For Bayesian filtering, these scores must reflect both the forecast distribution and the new observation. Neural approaches learn score models for trajectories or filtering distributions \citep{rozet2023scorebased,bao2024scorefilter}. The Ensemble Score Filter (EnSF) instead estimates the forecast score directly from the current forecast ensemble and adds an observation-dependent correction \citep{bao2024ensemble}. EnSF has been applied to surface quasigeostrophic dynamics \citep{bao2025sqg}, stochastic partial differential equations \citep{huynh2026spde}, and data-driven dynamical forecasting \citep{tang2026scoreenhanced}. Although EnSF avoids neural score training, obtaining accurate observation-conditioned scores at each diffusion noise level remains challenging.

The observation likelihood describes what a sensor would measure at a given system state. A reverse particle, however, represents a state perturbed by artificial noise. At a given noise level, several system states can be compatible with that particle, even when they imply different sensor readings. For a given noisy particle, we average the observation likelihood over the forecast distribution of the underlying state conditional on that particle. EnSF evaluates its likelihood-gradient correction at the noisy particle itself \citep{bao2024ensemble}. Diffusion posterior sampling evaluates the likelihood at a single denoised state estimate \citep{chung2023diffusion}. Both approaches avoid integrating over the possible states, but the resulting score can be biased. A filter with a small forecast ensemble must still account for uncertainty about which system state gave rise to each noisy particle.

We propose the Adaptive Ensemble Conditional Score Filter (AECSF) for this setting. Under Gaussian noising, the conditional Tweedie identity relates the required score to the mean system state given both the noisy reverse particle and the observation \citep{boys2023tweedie,peng2024posteriorcovariance,patsenker2025injecting}. In the training-free implementation, the forecast members supply an initial set of proposals. Sharing this ensemble avoids drawing a separate set of proposals for every reverse particle. Each particle weights the proposals according to its own noisy state. As denoising progresses, the states that matter for one particle can change, and fixed proposals can miss them. AECSF uses the current reverse particles to guide proposal selection during the same run. We make three contributions.

\begingroup
\setlength{\topsep}{3pt}
\setlength{\partopsep}{0pt}
\setlength{\itemsep}{2pt}
\setlength{\parsep}{0pt}
\setlength{\parskip}{0pt}
\begin{itemize}
\item We propose AECSF, a filter with a training-free implementation and an analytically tractable score estimator based on the conditional Tweedie identity. Each reverse particle estimates its own conditional mean from the same weighted proposals, and the evolving particles guide proposal updates within the same reverse-diffusion run.

\item We characterize when a fixed proposal measure gives the exact noisy posterior score. For adaptive proposals, we establish consistency of conditional-mean estimates and reverse-sampling endpoints relative to a numerical reference. We also bound the contribution of conditional-mean errors to the final sampling error.

\item We evaluate AECSF on controlled Gaussian sampling problems, 10,000-dimensional Lorenz--96 with 20 members, and a 1,024-dimensional Kuramoto--Sivashinsky system. It improves posterior-sampling accuracy over the tested guidance rules and achieves competitive filtering accuracy at the reported costs. We also test the same estimator with a learned forecast prior for sparse Navier--Stokes observations.
\end{itemize}
\endgroup

\section{Related Work}
\label{sec:related_work}

Ensemble score and flow filters construct analysis updates from forecast
samples. EnSF estimates the forecast score and adds a likelihood correction
\citep{bao2024ensemble}; EnFF uses Monte Carlo likelihood weighting or
guidance based on local likelihood linearization
\citep{transue2025flowmatching}. IEnSF repeats reverse diffusion, updating
its score approximation between runs \citep{zhang2025iensf}.
Binder et al. obtain a conditional score from a joint KDE of forecast
states and simulated observations, with centers fixed within each
analysis \citep{binder2026closedform}. AECSF instead adapts shared
proposal locations and integration weights within one reverse run,
estimating the conditional means needed by the current reverse particles.

Learned generative models supply priors or transitions for assimilation.
SDA combines learned local trajectory scores to generate trajectories
consistent with observations \citep{rozet2023scorebased}. FlowDAS learns
state transitions through stochastic interpolants and uses Monte Carlo
likelihood guidance \citep{chen2025flowdas}; DAISI maps forecast members
into a fixed learned prior's latent states to initialize conditional
generation \citep{andrae2026daisi}. Latent-EnSF learns shared state and
observation representations, while LD-EnSF learns latent dynamics to
reduce forecast cost \citep{si2025latentensf,xiao2026ldensf}. AECSF estimates its
conditional score from weighted proposals and the observation likelihood. In our
sparse assimilation experiment, a learned model supplies the forecast prior
from which we construct those proposals.

Tempered SMC uses intermediate targets, resampling and moves to manage
weight degeneracy \citep{delmoral2006sequential,gilks2001following}.
With pretrained diffusion emulators, SURGE weights and resamples guided
diffusion paths using a Girsanov correction \citep{wei2026surge}, while
FA-APF uses predictive-likelihood weights to select ancestors and an
approximate observation-conditioned sampler to generate descendants
\citep{savary2026generative}. In AECSF, resampling and movement adapt the
proposals used to estimate the score. If the proposals stay fixed and only
their likelihood weights change, the temperature factors cancel. The resulting
score is obtained by first weighting the fixed proposals by the full observation
likelihood and then applying Gaussian noising to the resulting empirical
distribution. This parallels EnFF's fixed-ensemble result \citep{transue2025flowmatching}.
With guided resampling, the current reverse particles influence which proposals
are selected. Importance weights correct for that selection. The updated
proposals then supply a conditional mean for each of the same reverse particles,
whose final states form the analysis ensemble.

\section{The Filtering Problem and the Conditional Score}
\label{sec:problem_main}

For state \(x_n\in\R^d\) and observation \(y_n\in\R^{m_n}\), let \(p_n\) be the transition kernel and \(g_n\) the likelihood. The forecast and analysis densities condition on \(y_{1:n-1}\) and \(y_{1:n}\), respectively:
\begin{align}
    x_n&\sim p_n(\cdot\given x_{n-1}),\qquad y_n\sim g_n(\cdot\given x_n),
    \label{eq:main_state_model}\\
    \pi_n^f(x_n)&=\int p_n(x_n\given x_{n-1})\pi_{n-1}^a(x_{n-1})\,\dd x_{n-1},\nonumber\\
    \pi_n^a(x_n)&\propto g_n(y_n\given x_n)\pi_n^f(x_n).
    \label{eq:main_analysis_recursion}
\end{align}
Within a single assimilation cycle, let \(Z\) denote the system state before we add artificial noise. At diffusion time \(\tau\in[0,1]\), the noisy state \(X_\tau\) has the conditional law below.
\begin{equation}
\begin{aligned}
    Z&\sim\pi_n^f,\qquad Y_n\given Z\sim g_n(\cdot\given Z),\\
    X_\tau\given Z&\sim\N(a_\tau Z,\sigma_\tau^2\I_d).
\end{aligned}
\label{eq:main_vp_forward}
\end{equation}
Here \(a_\tau>0\) scales the signal and \(\sigma_\tau^2>0\) is the noise variance. Given \(Z\), the diffusion noise is independent of the observation. Conditioning on \(Y_n=y_n\) gives \(Z\) the analysis density \(\pi_n^a\). Adding the Gaussian noise then gives \(X_\tau\) the density \(q_{n,\tau}^a\), whose score drives reverse diffusion:
\begin{equation}
    s_{n,\tau}^a(x;y_n)=\nabla_x\log q_{n,\tau}^a(x\given y_n).
    \label{eq:main_target_score}
\end{equation}
The conditional Tweedie identity \citep{efron2011tweedies,boys2023tweedie} expresses this score through the mean state conditioned on the noisy particle and observation:
\begin{align}
    D_{n,\tau}(x;y_n)&=\E[Z\given X_\tau=x,Y_n=y_n],
    \label{eq:main_conditional_mean}\\
    s_{n,\tau}^a(x;y_n)
    &=-\frac{x}{\sigma_\tau^2}
    +\frac{a_\tau}{\sigma_\tau^2}D_{n,\tau}(x;y_n).
    \label{eq:main_conditional_tweedie}
\end{align}
Thus score construction reduces to conditional-mean estimation; Appendix~\ref{subsec:conditional_tweedie_derivation} gives the derivation.

\section{The Adaptive Ensemble Conditional Score Filter}
\label{sec:ecsf_main}

At cycle \(n\), AECSF takes a forecast ensemble
\(\mathcal X_n^f=\{x_n^{f,j}\}_{j=1}^J\), an observation \(y_n\), and a
differentiable likelihood. It evolves \(J\) reverse particles into the
analysis ensemble \(\mathcal X_n^a\), using \(M\) weighted proposals shared
by all reverse particles.

At each cycle, we use \(\mathcal X_n^f\) to approximate the forecast prior
\(\pi_n^f\) by \(\widetilde\pi_n^f\). In the training-free implementation,
we fit a diagonal Gaussian \(\widetilde\pi_n^f=\N(m_f,\diag(v_f))\),
where \(m_f\) is the ensemble mean and \(v_f\) contains the regularized
variance at each coordinate. We start with one proposal per forecast
member (\(M=J\)). Table~\ref{tab:l96_proposal_capacity} compares
different values of \(M\) at fixed \(J\).

With sparse observations, the prior relates what the sensors measure to
the state elsewhere in the field. To learn these dependencies, we
\hyperref[app:ns_flow_training]{train a conditional normalizing flow} on paired
forecast ensembles and full fields from simulations independent of the
assimilation runs. At each cycle, features \(c_n\) summarize the current
forecast ensemble \(\mathcal X_n^f\): its mean pattern, its spread, and how
deviations from the mean vary together at neighbouring locations.
Conditioned on these features, the learned density \(p_\theta(z\mid c_n)\) describes how the coordinates of
the full field \(z\) vary together before we incorporate the new observation
(Appendix~\ref{app:ns_flow_training}, Eq.~\eqref{eq:ns_flow_density_score}).

We combine either forecast prior with the tempered likelihood to obtain the proposal
target in \eqref{eq:main_tempered_proposal}. For the fitted Gaussian, we use its
prior score to move proposals in state space, as in \eqref{eq:rho_tempered_move}.
With the learned prior, we update a standard Gaussian variable \(u\), whose log
prior density has gradient \(-u\). The forecast ensemble sets how we scale \(u\)
before decoding it. The trained decoder then determines how a change in \(u\)
affects the vorticity field and hence the predicted sensor velocities.
With either prior, the updated proposals
give each reverse particle its conditional mean \eqref{eq:main_ecsf_estimator}
and the score for its next update \eqref{eq:main_estimated_score}.

\newlength{\aecsfTextFloatSep}
\newlength{\aecsfFloatSep}
\newlength{\aecsfInTextSep}
\newlength{\aecsfAboveCaptionSkip}
\setlength{\aecsfTextFloatSep}{\textfloatsep}
\setlength{\aecsfFloatSep}{\floatsep}
\setlength{\aecsfInTextSep}{\intextsep}
\setlength{\aecsfAboveCaptionSkip}{\abovecaptionskip}
\setlength{\textfloatsep}{8pt plus 2pt minus 1pt}
\setlength{\floatsep}{8pt plus 2pt minus 1pt}
\setlength{\intextsep}{8pt plus 2pt minus 1pt}
\setlength{\abovecaptionskip}{6pt}
\begin{figure}[t]
\centering
\includegraphics[width=\textwidth]{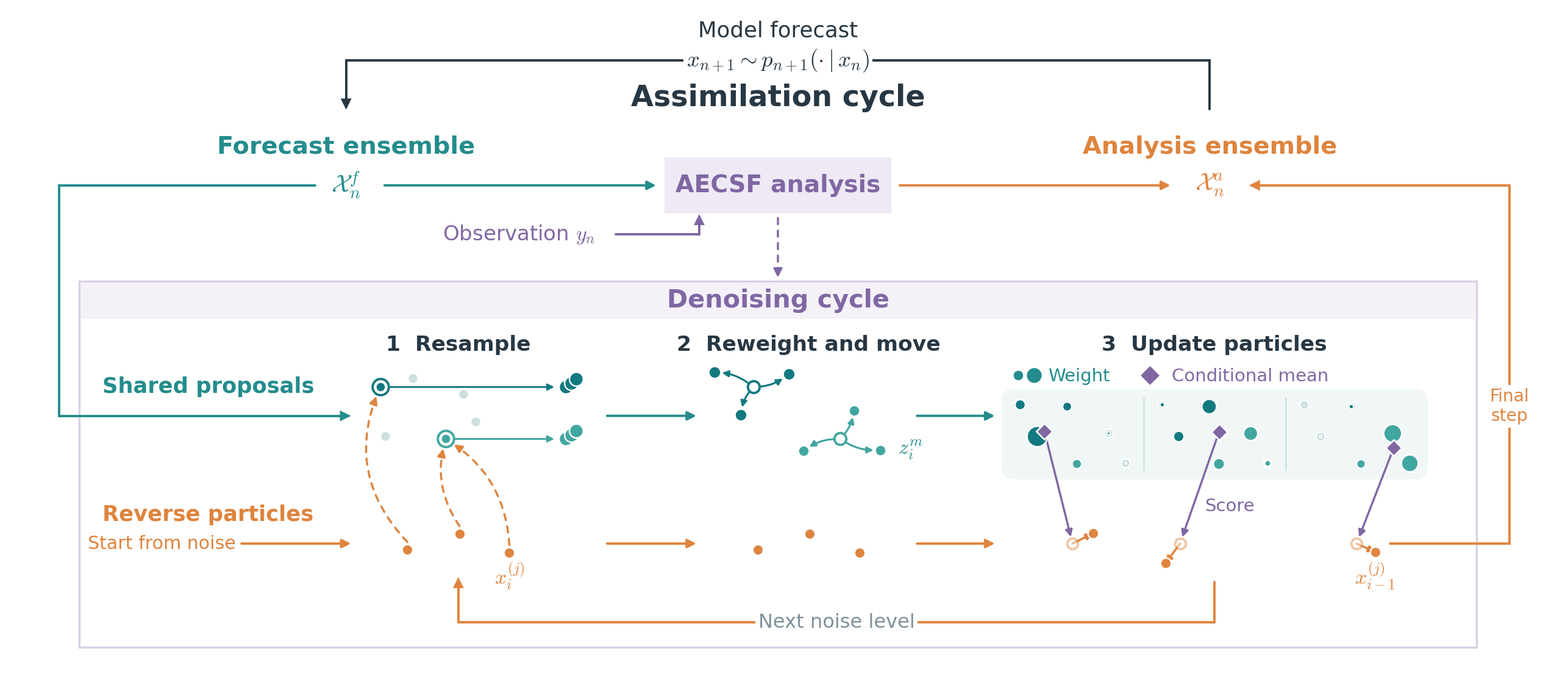}
\caption{AECSF with guided resampling. The outer loop propagates the analysis
ensemble using the model transition in \eqref{eq:main_state_model}.
\textbf{1. Resample.} The reverse particles guide proposal selection
\eqref{eq:refresh_resampling} (dashed orange arrows) using the target
weights \eqref{eq:incremental_target_weights}.
\textbf{2. Reweight and move.} Correct the weights of the resampled
proposals \eqref{eq:ancestor_correction} and move them using
\eqref{eq:rho_tempered_move}.
\textbf{3. Update particles.} The three panels show the same proposal
locations. Dot sizes show the weights \(c_{j,m}\) for a different reverse
particle in each panel. Purple diamonds mark the resulting conditional
means \eqref{eq:main_ecsf_estimator}. The scores
\eqref{eq:main_estimated_score} drive the reverse update
\eqref{eq:main_reverse_update}.}
\label{fig:aecsf_interaction}
\end{figure}

\subsection{Estimating the conditional score with shared proposals}
\label{subsec:main_score_estimation}

For a reverse particle at \(x\), several system states \(z\) may be compatible with both \(x\) and the observation \(y_n\). Their conditional density determines the mean in \eqref{eq:main_conditional_mean}:
\begin{equation}
    p(z\given X_\tau=x,Y_n=y_n)
    \propto\phi_{\sigma_\tau^2\I_d}(x-a_\tau z)\pi_n^f(z)g_n(y_n\given z).
    \label{eq:main_conditional_law}
\end{equation}
Here, \(\phi_{\sigma_\tau^2\I_d}\) is the centered Gaussian density with covariance \(\sigma_\tau^2\I_d\). The three factors favor system states compatible with the noisy state, plausible under the forecast, and consistent with the observation.

To share proposals across reverse particles, we include part of the observation likelihood in a common reference distribution. As in annealed importance sampling and tempered SMC \citep{neal2001annealed,delmoral2006sequential}, an exponent \(\rho\) controls how much of the likelihood enters this distribution:
\begin{equation}
    \pi_{n,\rho}(z)\propto\pi_n^f(z)g_n(y_n\given z)^\rho,
    \qquad 0\leq\rho\leq1.
    \label{eq:main_tempered_proposal}
\end{equation}
It interpolates between the forecast and posterior and gives the factorization
\begin{equation}
    p(z\given X_\tau=x,Y_n=y_n)
    \propto\pi_{n,\rho}(z)\phi_{\sigma_\tau^2\I_d}(x-a_\tau z)
    g_n(y_n\given z)^{1-\rho}.
    \label{eq:main_tempered_factorization}
\end{equation}
The full likelihood is retained through \(g_n^\rho g_n^{1-\rho}=g_n\).
Increasing \(\rho\) shifts more observation information into the shared
reference and leaves less for the particle-specific weights.

On the reverse grid \(1=\tau_{N_\tau}>\cdots>\tau_0=0\), write
\(x_i^{(j)}=x_{\tau_i}^{(j)}\) for reverse particle \(j\) at step \(i\).
The first factor in \eqref{eq:main_tempered_factorization} is the tempered
distribution. We construct it using the fitted or learned forecast prior
\(\widetilde\pi_n^f\) in \eqref{eq:main_tempered_proposal} and approximate it
with \(M\) shared proposals. After any scheduled adaptation at step \(i\),
their empirical distribution places probability mass \(W_{i,m}\) at each
proposal location \(z_i^m\), with \(\sum_m W_{i,m}=1\):
\begin{equation}
    \nu_{n,i}^M=\sum_{m=1}^M W_{i,m}\delta_{z_i^m}.
    \label{eq:main_candidate_measure}
\end{equation}
Every reverse particle starts from these same weights. To estimate the
conditional mean for particle \(j\), we multiply them by the other two
factors in \eqref{eq:main_tempered_factorization}.
The Gaussian kernel \(K_i(x,z)=\phi_{\sigma_{\tau_i}^2\I_d}(x-a_{\tau_i}z)\)
measures how well each proposal agrees with that particle's location
\(x=x_i^{(j)}\). Multiplying the kernel by the remaining likelihood gives
\(F_i(x,z)=K_i(x,z)g_n(y_n\given z)^{1-\rho_i}\).
The estimated conditional mean is a weighted average of the proposals,
\begin{equation}
    \widehat D_{n,i}^M(x_i^{(j)};y_n)=\sum_{m=1}^M c_{j,m}z_i^m,
    \qquad \text{where}\quad
    c_{j,m}=\frac{W_{i,m}F_i(x_i^{(j)},z_i^m)}
       {\sum_{\ell=1}^M W_{i,\ell}F_i(x_i^{(j)},z_i^\ell)}.
\label{eq:main_ecsf_estimator}
\end{equation}
The weighted average in \eqref{eq:main_ecsf_estimator} estimates the mean
of the clean state given particle \(j\)'s noisy location \(x_i^{(j)}\)
and the observation \(y_n\). Under Gaussian noising, Tweedie's identity
\eqref{eq:main_conditional_tweedie} expresses the score as the displacement
from this location to the scaled conditional mean, divided by the noise
variance. Using the ensemble estimate for that mean gives particle \(j\)
the score for its next reverse step,
\begin{equation}
    \widehat s_{n,\tau_i}^a(x_i^{(j)};y_n)
    =-\frac{x_i^{(j)}}{\sigma_{\tau_i}^2}
    +\frac{a_{\tau_i}}{\sigma_{\tau_i}^2}
      \widehat D_{n,i}^M(x_i^{(j)};y_n).
    \label{eq:main_estimated_score}
\end{equation}

\subsection{Updating the shared proposals}
\label{subsec:main_proposal_updates}

After a reverse particle takes a step using this score, the proposals that
contributed most to its conditional mean may no longer be the most relevant
at its new location. During resampling, the current
reverse particles guide which proposals are selected. The reverse particles
stay in place while the proposals are reweighted and moved. Each particle
then uses the updated proposals to estimate the score for its next step
(Figure~\ref{fig:aecsf_interaction}).

\paragraph{Updating the target weights.}
We initialize the \(M\) shared proposals with equal weights \(1/M\).
In the training-free schedule, we carry these weights from one step to the
next and update them as the temperature \(\rho_i=1-\tau_i\) increases.
Resampling and movement follow separate schedules. Let \((z_m^-,W_m^-)\)
be the proposals and weights carried into this step from temperature
\(\rho^-\). We multiply each weight by the incremental likelihood and
normalize over proposals \citep{delmoral2006sequential}:
\begin{equation}
    \omega_{i,m}=\frac{W_m^-g_n(y_n\given z_m^-)^{\rho_i-\rho^-}}
    {\sum_\ell W_\ell^-g_n(y_n\given z_\ell^-)^{\rho_i-\rho^-}}.
    \label{eq:incremental_target_weights}
\end{equation}

\paragraph{Reverse particle guided resampling.}
The target weight \(\omega_{i,m}\) tells us how much mass proposal \(m\)
carries in the tempered distribution. Two proposals with the same target
weight can contribute differently to the conditional mean at a given reverse
particle. In step~1 of Figure~\ref{fig:aecsf_interaction}, we therefore
multiply the target weights by \(K_i\) to favor proposals whose scaled
locations are closer to particle \(j\). Normalizing over proposals gives the
probability \(r_{i,jm}\) that particle \(j\) assigns to proposal \(m\).
The remaining likelihood enters later, when the particle estimates its
conditional mean in \eqref{eq:main_ecsf_estimator}.
Since all particles share one proposal ensemble, we average their
probabilities for each proposal to obtain its resampling probability \(s_{i,m}\):
\begin{equation}
    r_{i,jm}=\frac{K_i(x_i^{(j)},z_m^-)\omega_{i,m}}
    {\sum_\ell K_i(x_i^{(j)},z_\ell^-)\omega_{i,\ell}},
    \qquad s_{i,m}=\frac1J\sum_{j=1}^Jr_{i,jm}.
    \label{eq:refresh_resampling}
\end{equation}
Equation~\eqref{eq:refresh_resampling} determines how often each proposal is
copied. A proposal favored by several reverse particles may be selected more
often than its target weight warrants. If all copies carried equal weight,
that proposal would then contribute too much to the shared ensemble.

We correct for this in step~2 of Figure~\ref{fig:aecsf_interaction}.
Let \(A_m\) be the ancestor of copy \(m\), drawn by stratified resampling
from \(s_i\). We assign the copy its ancestor's target weight
\(\omega_{i,A_m}\) divided by its selection probability \(s_{i,A_m}\),
then normalize these weights over all copies \citep{pitt1999auxiliary,liu1998sequential}:
\begin{equation}
    z_m^+=z_{A_m}^-,\qquad
    v_m=\frac{\omega_{i,A_m}}{s_{i,A_m}},\qquad
    W_m^+=\frac{v_m}{\sum_\ell v_\ell}.
    \label{eq:ancestor_correction}
\end{equation}
Appendix~\ref{subsec:active_refresh} gives the correction identity and
its support condition. When resampling is skipped, we keep the
locations and set \(W_m^+=\omega_{i,m}\).

\paragraph{Moving the proposals.}
Resampling can leave several copies at one location. Unadjusted Langevin moves let
the copies explore nearby states under the tempered target \citep{gilks2001following,roberts1996langevin}.
For the state-space move, we fit a diagonal Gaussian to \(\mathcal X_n^f\), regularize
its variances, and hold it fixed during analysis. Its score
\(\widehat s_n^f(z)=\nabla_z\log\widetilde\pi_n^f(z)\) and the tempered likelihood
gradient drive the following step of size \(h\), starting at \(z_m^{(0)}=z_m^+\).
In both Langevin updates, we draw \(\xi\) independently for each proposal at each step.
\begin{equation}
\begin{aligned}
    z_m^{(\ell+1)}=z_m^{(\ell)}
    &+\frac h2\left[\widehat s_n^f(z_m^{(\ell)})
       +\rho_i\nabla_z\log g_n(y_n\given z_m^{(\ell)})\right]\\
    &+\sqrt h\,\xi,\qquad\xi\sim\N(0,\I_d).
\end{aligned}
\label{eq:rho_tempered_move}
\end{equation}
We recompute the likelihood after each move.

With the learned prior, we take the same Langevin step in the flow's
Gaussian input coordinates \(u\). Writing \(\omega(u,b)\) for the full
vorticity field, the update at temperature \(\rho\) is given by \eqref{eq:ns_latent_move}:
\[
u^+=u+\tfrac h2\bigl[-u+\rho\,\nabla_u\log g_n(y_n\mid\omega(u,b))\bigr]
       +\sqrt h\,\xi,\qquad \xi\sim\N(0,\I_{\dim u}).
\]
Here \(b\) sets the low-frequency perturbation added to the decoded field
in \eqref{eq:ns_forecast_prior} to represent uncertainty in large-scale
vorticity structures. We hold \(b\) fixed while differentiating the log
likelihood through the decoder with respect to \(u\). The Gaussian prior
score \(-u\) takes the place of \(\widehat s_n^f(z)\) in
\eqref{eq:rho_tempered_move}. After each block of moves,
we draw \(b\) from its Gaussian conditional distribution using
\eqref{eq:ns_low_frequency_draw} and decode the updated proposals.
In both implementations, a move changes the proposal locations and preserves
the shared weights. Each reverse particle combines those weights with the
kernel and remaining likelihood at the new locations to estimate its own
conditional mean in \eqref{eq:main_ecsf_estimator}.

\subsection{Reverse sampling and recursive filtering}
\label{subsec:main_reverse_filtering}

The forecast score in \eqref{eq:rho_tempered_move} guides proposal moves.
The conditional score in \eqref{eq:main_estimated_score} advances each
reverse particle (step~3 in Figure~\ref{fig:aecsf_interaction}).
Let \(f_\tau\) and \(r_\tau\) be the drift coefficient
and diffusion amplitude of the Gaussian noising process
(Appendix~\ref{subsec:reverse_recursive}). With \(\Delta_i=\tau_i-\tau_{i-1}\),
the training-free implementation uses the Euler--Maruyama reverse update
\begin{equation}
\begin{aligned}
    x_{i-1}^{(j)}=x_i^{(j)}
    &-\Delta_i\left[f_{\tau_i}x_i^{(j)}
    -r_{\tau_i}^2\widehat s_{n,\tau_i}^a(x_i^{(j)};y_n)\right]\\
    &+\sqrt{\Delta_i r_{\tau_i}^2}\,\xi_i^{(j)},
    \qquad\xi_i^{(j)}\sim\N(0,\I_d).
\end{aligned}
\label{eq:main_reverse_update}
\end{equation}
The learned-prior NS experiment uses the same conditional score in the
probability-flow ODE update \eqref{eq:ns_reverse_ode_step}. The final
reverse particles form the analysis ensemble for the next forecast.

\subsection{Exactness and approximation error}
\label{subsec:main_error_analysis}

We first identify the distribution represented by the estimated score, then
relate conditional-mean error to the final analysis distribution.

\paragraph{The score of a fixed proposal ensemble.}
Applying the remaining likelihood factor to the weighted proposals gives
\begin{equation}
\widehat\eta_{n,i}^M(\dd z)=
\frac{g_n(y_n\given z)^{1-\rho_i}\nu_{n,i}^M(\dd z)}
{\nu_{n,i}^M(g_n(y_n\given\cdot)^{1-\rho_i})}.
\label{eq:main_empirical_surrogate}
\end{equation}
With proposal locations and weights held fixed during differentiation,
the unclipped estimate in \eqref{eq:main_estimated_score} is exactly the
score of the Gaussian mixture
\begin{equation}
\begin{aligned}
\widehat q_{n,i}^M(x;y_n)&=
\int\phi_{\sigma_{\tau_i}^2\I_d}(x-a_{\tau_i}z)
\widehat\eta_{n,i}^M(\dd z),\\
\widehat s_{n,\tau_i}^a(x;y_n)&=\nabla_x\log\widehat q_{n,i}^M(x;y_n).
\end{aligned}
\label{eq:main_frozen_bank_score}
\end{equation}
Each proposal defines a Gaussian component, weighted by its integration
weight and remaining likelihood.
Under the conditions in Appendix~\ref{subsec:surrogate_target_condition},
this mixture and the Gaussian-noised posterior have the same score everywhere
if and only if the fixed proposal measure equals the tempered target \(\pi_{n,\rho_i}\).

\paragraph{From conditional means to scores.}
Let $D_\nu$ denote the conditional mean estimated from a proposal
measure $\nu$. The conditional Tweedie identity gives
\begin{equation}
\widehat s_{\nu,n,\tau}^a(x;y_n)-s_{n,\tau}^a(x;y_n)
=\frac{a_\tau}{\sigma_\tau^2}[D_\nu-D_{n,\tau}(x;y_n)].
\label{eq:main_score_mean_error}
\end{equation}
Score error equals conditional-mean error multiplied by $a_\tau/\sigma_\tau^2$.
Appendix~\ref{subsec:weighted_moment_error} bounds the conditional-mean
error through weighted moments and normalizing mass.

\paragraph{Error in the analysis distribution.}
To isolate the error from using finitely many proposals, we compare AECSF with
a numerical reference. Both start with the same weighted proposals and use the
same likelihood increments and move kernels. At each stage, the reference
propagates the entire proposal distribution through the kernel and integrates
over it to obtain the conditional means. These means drive its reverse
particles on the same Euler grid as AECSF.
Let \(k\) count completed Euler steps from \(\tau=1\), with \(K\) steps in
total. Let \(X_k^j\) and \(Y_k^j\) be the actual and reference particles,
coupled with the same noise.
Let $I$ select a member uniformly and independently.
Assume the reference Euler drift maps are $\beta_k$-Lipschitz
and the required second moments are finite.

\begin{samepage}
Theorem C.4 in Appendix~\ref{app:adaptive_candidate_theory} gives
\begin{equation}
\begin{aligned}
W_2\!\left(\mathcal L(X_K^I),\pi_n^a\right)
\le{}&\left(\prod_{k=1}^K\beta_k\right)e_0
+\sum_{k=1}^K w_kq_k
+W_2\!\left(\mathcal L(Y_K^I),\pi_n^a\right),
\end{aligned}
\label{eq:main_reverse_endpoint_bound}
\end{equation}
The first two terms bound the discrepancy between the actual and reference
endpoints. The last term measures how far the reference itself is from the
posterior. Here \(e_0\) is the root-mean-square initial distance between paired
particles. Each \(q_k\) is the root-mean-square score difference evaluated at
the actual particles before step \(k\). Equation~\eqref{eq:main_score_mean_error}
expresses that difference through the conditional means. The weights \(w_k\)
account for step size and subsequent error propagation (Theorem C.4).
\par
\end{samepage}

The reference retains the forecast approximation, the numerical proposal moves,
the reverse discretization, and the endpoint noise.
The same appendix gives fixed-grid consistency results and treats clipped scores.

\section{Experiments}
\label{sec:numerics}

We evaluate posterior sampling, recursive filtering, and subsequent prediction
using RMSE, componentwise CRPS, spread--skill ratio (SSR), and computational
cost. Paired methods share forecast information or physical inputs and
initialization; recursive forecasts then evolve from each method's analyses.
Appendix~\ref{app:experimental_protocols} defines metrics, protocols, and
timing intervals.

\subsection{Testing sampling accuracy on Gaussian posteriors}
\label{sec:controlled_score_recovery}

We compare the final samples with known Gaussian posteriors to test how proposal
resampling affects sampling accuracy. We use ten fully observed problems in
\(d=128\), with \(M=J=20\) and four repetitions each. Comparing against the
posterior under the forecast prior fitted to the common ensemble isolates sampling
error. Using the data-generating prior also captures error from fitting the prior. We retain the endpoint
noise in both references. AECSF has lower endpoint energy discrepancies than AECSF
without resampling and than guidance evaluated at either the noisy state or the
denoised forecast mean (Figure~\ref{fig:controlled_score_endpoint_dimension},
Appendix~\ref{app:controlled_highdim_details}). Resampling also lowers the errors
in the mean and diagonal variances of individual output ensembles.

To see how resampling affects the estimates used during sampling, we also
examine the proposals and the score. Adding resampling reduces errors in the
proposal mean and variance when weighting and movement are already in place.
When we evaluate both score estimates at the same reverse-particle locations,
guided selection reduces score NMSE by 4.65\% relative to standard SMC resampling.
Against the fitted posterior, it reduces endpoint energy discrepancy by 31.7\%
when output particles are pooled and by 27.6\% when particles are paired across
independent repetitions (Appendix~\ref{app:gaussian_direct_diagnostics}).

\subsection{Lorenz--96: nonlinear observations at \texorpdfstring{\(d=10{,}000\)}{d=10,000}}

On Lorenz--96, we test how well AECSF tracks 10,000 state variables with 20
members and how guided proposal selection affects filtering accuracy. We use
full arctangent observations with noise SD 0.05 and evaluate cycles 50--199
of 200. Factorized
tempered SMC samples the product of the diagonal forecast prior and
coordinatewise likelihood. Its transport variant matches weighted marginal
means and variances in the output ensemble. The internal AECSF control uses
tempered ancestor weights and equal-weight descendants, retaining proposal
moves and reverse sampling.

\begin{table}[t]
\centering
\small
\setlength{\tabcolsep}{2pt}
\caption{Lorenz--96 results (\(d=10{,}000\), 20 forecast and output members). Means \(\pm\) sample SDs over ten runs, except Enhanced Binder (three). AECSF reports the original benchmark reference; complete paired comparisons appear in Table~\ref{tab:l96_full_comparisons}. SMC uses 320 internal particles; AECSF uses 20 proposals. The SMC baselines use a factorized target; transport includes marginal moment matching. Timing scopes: Appendix~\ref{app:experimental_protocols}. Dashes denote unreported metrics.}
\label{tab:l96_representative_results}
\begin{tabular}{lcccc}
\toprule
Method & RMSE \(\downarrow\) & CRPS \(\downarrow\) & SSR \((\to1)\) & \shortstack{Timed update/\\cycle (s)}\\
\midrule
EnSF & \(0.2418\pm0.0006\) & \(0.1432\pm0.0002\) & \(1.7366\pm0.0042\) & \(0.0485\pm0.0010\)\\
LETKF & \(0.1263\pm0.0116\) & \(0.0578\pm0.0015\) & \(0.6119\pm0.0604\) & \(0.5093\pm0.0014\)\\
Binder et al. & \(5.1323\pm0.0082\) & \(4.1428\pm0.0076\) & \(0.0002466\pm0.0001329\) & \(0.1507\pm0.0015\)\\
Enhanced Binder & \(1.4312\pm0.0143\) & \(0.7589\pm0.0028\) & \(0.0422\pm0.0007\) & ---\\
Tempered SMC & \(2.7330\pm0.0377\) & \(1.7974\pm0.0340\) & \(0.0125\pm0.0002\) & \(1.2507\pm0.0032\)\\
SMC + transport & \(2.6054\pm0.0265\) & \(1.6862\pm0.0240\) & \(0.0133\pm0.0001\) & \(1.2602\pm0.0031\)\\
AECSF (SMC resampling) & \(0.1708\pm0.0054\) & \(0.0840\pm0.0010\) & \(0.7768\pm0.0252\) & \(0.2739\pm0.0027\)\\
AECSF & \(0.1545\pm0.0039\) & \(0.0776\pm0.0007\) & \(0.8998\pm0.0232\) & \(0.3052\pm0.0139\)\\
\bottomrule
\end{tabular}
\end{table}

LETKF has the lowest RMSE and CRPS in the original comparison; AECSF improves
both over EnSF and has SSR closest to one. LETKF's timing uses GPU-parallel
local updates with CUDA graphs (\hyperref[app:l96_letkf_implementation]{Appendix~\ref*{app:l96_protocol}}).
AECSF gives lower RMSE and CRPS than either standalone tempered SMC filter.
To isolate the effect of the resampling rule, we keep AECSF's proposal moves
and reverse sampler fixed. Switching from standard SMC resampling to guided
selection with weight correction reduces RMSE by 10.3\% and CRPS by 7.6\%,
with improvements in all ten runs
(Table~\ref{tab:l96_full_comparisons}).

The Binder comparison examines observation conditioning. Enhanced Binder
conditions the forecast-state KDE on the original likelihood, updating
within-kernel distributions and mixture weights before sampling. It closes
74.3\% of the original Binder--AECSF RMSE gap in the paired comparison,
but remains severely underdispersed, with global weights concentrating on one component
(Appendix~\ref{app:binder_transfer}).

\paragraph{Effects of resampling and moving proposals.}
To measure the effect of adding resampling, we keep the prior approximation,
reverse schedule, incremental weights, and 294 Langevin steps fixed. Adding
guided resampling with its weight correction reduces RMSE by 31.0\% and CRPS
by 28.1\%, at a 36.0\% increase in analysis time. Both errors fall in all ten
paired runs. When movement is removed, mean RMSE is near 5.12 both with guided
resampling and with fixed proposal locations and incremental weights only
(Figure~\ref{fig:l96_candidate_components} in Appendix~\ref{app:l96_protocol}).

A separate comparison asks whether the same time is better spent moving output
particles directly under the posterior. At approximately matched analysis time,
AECSF reduces RMSE by 9.9\% and CRPS by 17.3\% relative to direct posterior ULA.
Both errors are lower in all ten paired runs (Table~\ref{tab:l96_direct_cost}).
Table~\ref{tab:l96_direct_posterior} compares the methods at equal step counts.

\subsection{Kuramoto--Sivashinsky: nonlinear observations and ensemble scaling}

\begin{figure}[!t]
\centering
\includegraphics[width=\textwidth]{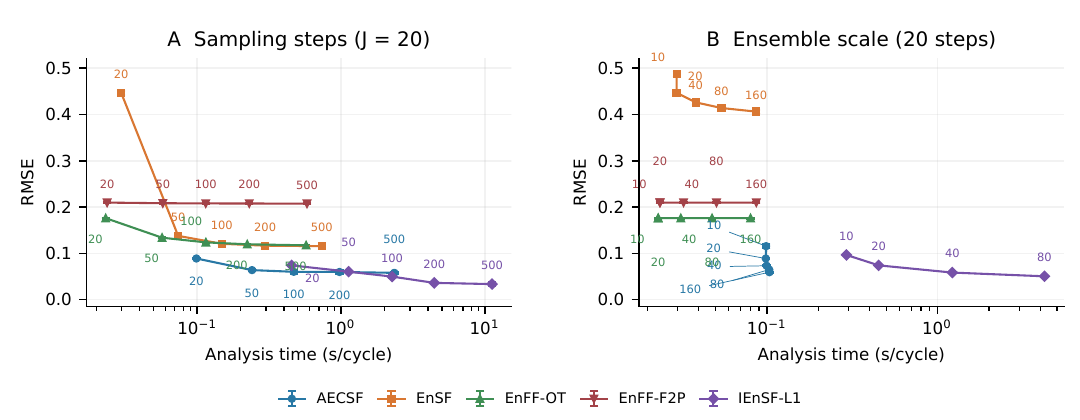}
\caption{KS analysis accuracy versus analysis time. (A) Sampling steps at \(J=20\);
(B) ensemble size at 20 steps. Labels indicate \(T\) (\(N_t\) for IEnSF-L1)
in A and \(J\) in B. Points/bars show means/sample SDs over five paired runs;
lines connect increasing settings.}
\label{fig:ks_cost_scale}
\end{figure}

On the 1,024-dimensional KS system, we vary step count and ensemble size separately
to measure their effects on accuracy, spread, and cost. We compare AECSF with
EnSF, EnFF-OT, EnFF-F2P, and IEnSF-L1 \citep{transue2025flowmatching,zhang2025iensf}
over five paired runs using full arctangent observations with noise SD 0.1.
With 20 members and about 0.46 seconds per analysis, AECSF has lower RMSE
than IEnSF-L1 (0.0601 versus 0.0740). At about 2.3 seconds, their respective
RMSEs are 0.0579 and 0.0498. CRPS follows the same ordering at both budgets.
At 20 steps, increasing AECSF's ensemble from 10 to 160 members lowers RMSE
from 0.1155 to 0.0593 while analysis time remains 0.099--0.104 seconds
(Figure~\ref{fig:ks_cost_scale}). At shared ensemble sizes, IEnSF-L1 has
lower RMSE but higher CRPS and cost. As AECSF's ensemble grows, its spread
increases slightly while its RMSE falls. Together these changes raise its SSR
from 1.60 to 3.33. IEnSF-L1's SSR rises from 2.32 to 4.66 over its tested range.

\begin{figure}[t]
\centering
\includegraphics[width=\textwidth]{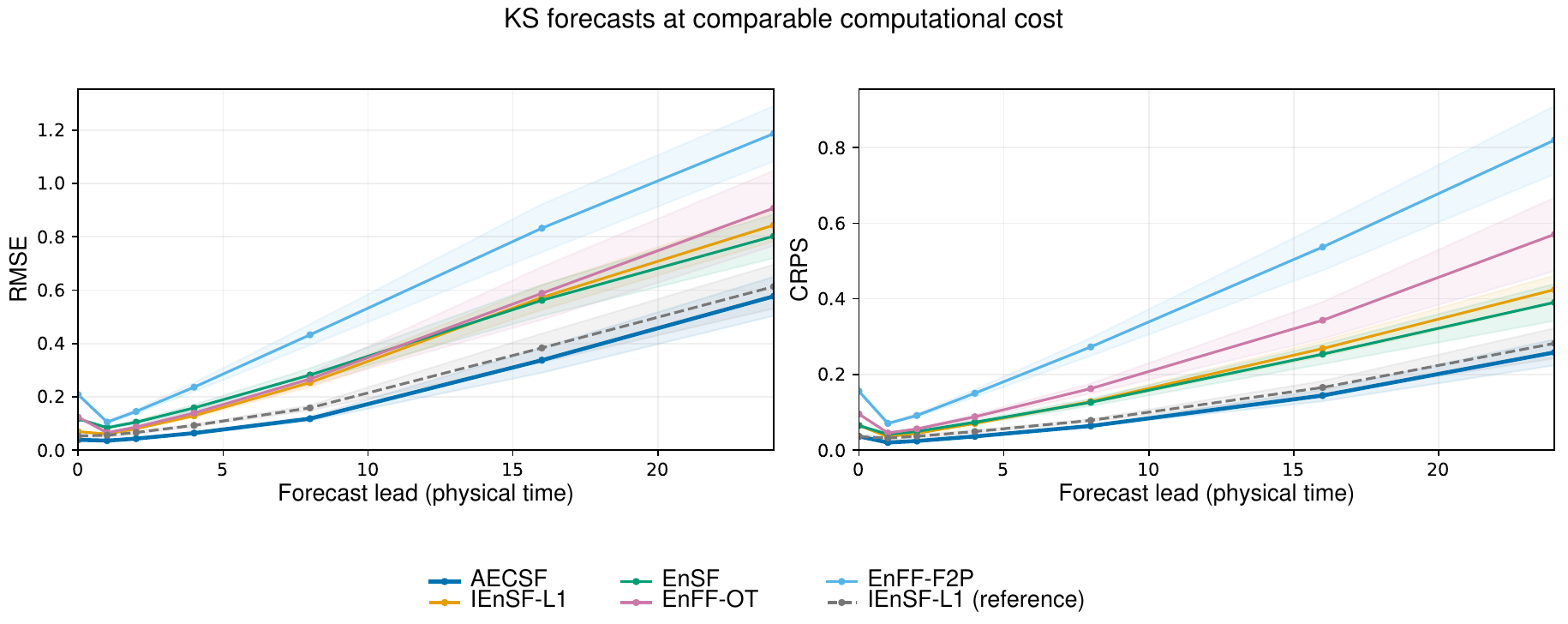}
\caption{KS forecast RMSE (left) and CRPS (right) at comparable computational cost.
Lines/shading show means/sample SDs over ten paired runs.
Dashed: higher-cost IEnSF-L1 reference.}
\label{fig:ks_forecast_comparable_cost}
\end{figure}

To assess how well the analyses predict later states, we propagate the analysis
ensembles without further observations. At a forecast horizon of 8 physical time units, AECSF
has the lowest RMSE and CRPS in all ten paired runs
(Figure~\ref{fig:ks_forecast_comparable_cost}). Averaged over leads 0--24,
its RMSE and CRPS are 40.75\% and 41.62\% lower than EnSF's.
Table~\ref{tab:ks_forecast_comparable_cost} lists the cost-comparison settings.
The dashed IEnSF-L1 curve uses a higher budget (\(J=20,N_t=100\)).

Costs include one assimilation cycle and the complete forecast to lead 8.
AECSF costs 8--17\% more than EnSF/EnFF. At lead 8, its central 90\%
intervals cover 99.83\% of outcomes, versus EnSF's 89.34\%, indicating
overly wide AECSF intervals.

\subsection{Sparse Navier--Stokes assimilation with a learned prior}
\label{subsec:ns_learned_main}

We assess flow reconstruction from sparse velocity observations by comparing AECSF
with forecast-only propagation from the same initial ensemble. Four paired
development trajectories span 50 cycles. The learned prior from
Section~\ref{sec:ecsf_main} combines a conditional flow with a
low-frequency Gaussian field \eqref{eq:ns_forecast_prior}
whose covariance is \hyperref[app:ns_low_frequency_covariance]{estimated from past innovations}.

\begin{table}[!htbp]
\centering
\small
\caption{Sparse Navier--Stokes assimilation on a $128\times128$ grid with 64 two-component velocity sensors, cycles 41--50. Means $\pm$ sample SDs across four trajectories. RMSE: velocity; CRPS and SSR: vorticity.}
\label{tab:ns_learned_thirty}
\begin{tabular}{lccc}
\toprule
Method & RMSE $\downarrow$ & CRPS $\downarrow$ & SSR ($\to1$)\\
\midrule
Forecast-only & \(0.58664\pm0.02599\) & \(1.48911\pm0.05970\) & \(1.0326\pm0.0343\)\\
AECSF & \(\mathbf{0.03676\pm0.00652}\) & \(\mathbf{0.11870\pm0.02152}\) & \(0.8601\pm0.1740\)\\
LETKF & \(0.07537\pm0.00983\) & \(0.28058\pm0.04357\) & \(2.6467\pm0.2127\)\\
EnSF & \(0.76133\pm0.07986\) & \(2.77979\pm0.24413\) & \(0.0488\pm0.0038\)\\
EnFF-OT & \(0.77696\pm0.04337\) & \(2.86202\pm0.14024\) & \(0.0028\pm0.0001\)\\
EnFF-F2P & \(0.44263\pm0.04114\) & \(1.19144\pm0.11279\) & \(1.1420\pm0.0663\)\\
\bottomrule
\end{tabular}
\end{table}

\begin{figure}[!htbp]
\centering
\includegraphics[width=\textwidth]{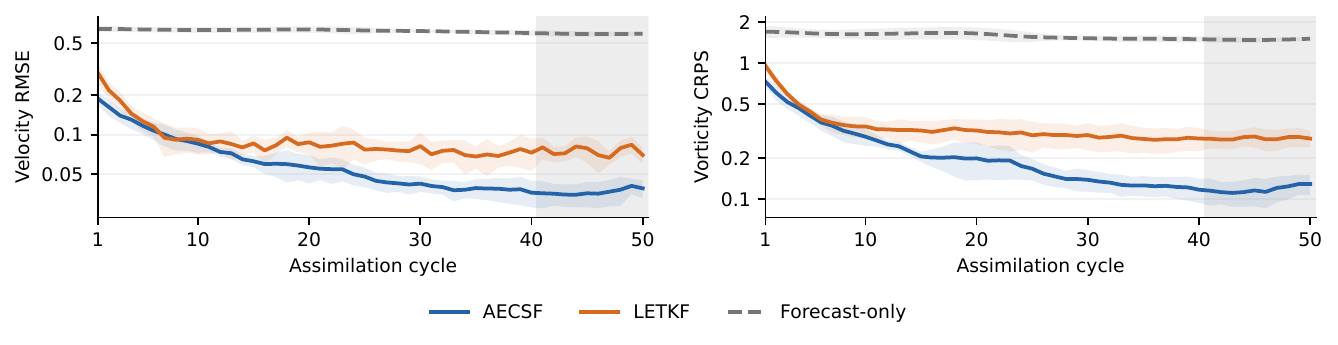}
\caption{NS accuracy over 50 cycles: velocity RMSE (left), vorticity CRPS (right). Lines/bands: means/sample SDs across four trajectories. Vertical log scales; gray: cycles 41--50.}
\label{fig:ns_thirty_accuracy}
\end{figure}

Over cycles 41--50, both AECSF and LETKF reduce errors relative to forecast-only propagation. AECSF has the lowest velocity RMSE and vorticity CRPS of the six methods (Table~\ref{tab:ns_learned_thirty}). Relative to LETKF, these errors fall by 51.2\% and 57.7\%, respectively, and both fall on every trajectory when averaged over this window. The AECSF ensemble remains underdispersed on two of the four trajectories. Figure~\ref{fig:ns_thirty_accuracy} shows the full 50-cycle record. Figures~\ref{fig:ns_thirty_low}--\ref{fig:ns50_vorticity_residuals} in the appendix show the low-frequency errors and the reconstructed flow fields.

\section{Discussion and conclusion}
\label{sec:conclusion}

A small forecast ensemble must support many different estimates during denoising, because each reverse particle needs a conditional mean at its current location. AECSF shares proposals across particles and adapts them as these locations change. Our analysis relates errors in the estimated means to the discrepancy between the final samples and a numerical reference.

The Lorenz--96 comparisons show why both proposal selection and movement matter. Guiding selection with the reverse particles and correcting the descendant weights improves on standard resampling. Removing movement gives large errors even with guided selection. AECSF also has lower errors than both independent tempered SMC filters. At comparable analysis cost, it improves on direct posterior ULA. The KDE comparison reveals a different difficulty. Conditioning Binder's forecast KDE on the original likelihood closes much of the RMSE gap, yet most of the mixture weight still falls on a single component and the ensemble remains underdispersed.

On Kuramoto--Sivashinsky, AECSF improves on EnFF-OT at comparable analysis cost, while its accuracy relative to IEnSF-L1 depends on the sampling budget. Increasing the AECSF ensemble lowers RMSE with little change in measured analysis time, but also raises the spread--skill ratio. When propagated without further observations, the tested AECSF ensembles give lower subsequent forecast RMSE and CRPS at comparable cost. For sparse Navier--Stokes observations, we keep the conditional-mean estimator and use a learned forecast prior to represent spatial dependence. Over 50 cycles, AECSF with the learned prior had lower average velocity RMSE and vorticity CRPS than forecast-only propagation and LETKF. The overly wide KS ensembles and the narrow ensembles on two Navier--Stokes trajectories point to ensemble calibration as a priority alongside accuracy.

\clearpage
\setlength{\textfloatsep}{\aecsfTextFloatSep}
\setlength{\floatsep}{\aecsfFloatSep}
\setlength{\intextsep}{\aecsfInTextSep}
\setlength{\abovecaptionskip}{\aecsfAboveCaptionSkip}
\subsection*{AI use statement}

In this work, we used  generative AI tools to assist with implementation and experimental design, mathematical derivations and arguments, and drafting and polishing portions of the manuscript. The authors reviewed all AI-assisted content, independently verified the mathematical arguments and references, and tested the implementation using analytical cases and controlled experiments. The authors take full responsibility for the final content of the paper.

\subsection*{Reproducibility statement}

The main text presents the estimator and principal findings; the appendices provide the derivations, algorithms, model and observation equations, evaluation metrics, configuration-selection history and evaluation seeds, hyperparameters, and supplementary diagnostics. All benchmark and learned-prior training data are generated from simulated dynamical models; no external observational dataset is required. The code is available upon request, and it will be made publicly available after acceptance.

\subsection*{Acknowledgments}

This work is supported by the National Natural Science Foundation of China (Nos. 12671506 and 12271409), the  Basic Research Project of CNPC (No. 2023ZZ05), and the Fundamental Research Funds for the Central Universities.

\bibliography{ctsf_references}

@article{evensen1994sequential,
  author  = {Evensen, Geir},
  title   = {Sequential Data Assimilation with a Nonlinear Quasi-Geostrophic Model Using Monte Carlo Methods to Forecast Error Statistics},
  journal = {Journal of Geophysical Research: Oceans},
  year    = {1994},
  volume  = {99},
  number  = {C5},
  pages   = {10143--10162},
  doi     = {10.1029/94JC00572}
}

@article{hunt2007efficient,
  author  = {Hunt, Brian R. and Kostelich, Eric J. and Szunyogh, Istvan},
  title   = {Efficient Data Assimilation for Spatiotemporal Chaos: A Local Ensemble Transform Kalman Filter},
  journal = {Physica D: Nonlinear Phenomena},
  year    = {2007},
  volume  = {230},
  number  = {1--2},
  pages   = {112--126},
  doi     = {10.1016/j.physd.2006.11.008}
}

@book{reich2015probabilistic,
  author    = {Reich, Sebastian and Cotter, Colin},
  title     = {Probabilistic Forecasting and Bayesian Data Assimilation},
  publisher = {Cambridge University Press},
  year      = {2015},
  doi       = {10.1017/CBO9781107706804}
}

@book{sanzalonso2023data,
  author    = {Sanz-Alonso, Daniel and Stuart, Andrew M. and Taeb, Armeen},
  title     = {Inverse Problems and Data Assimilation},
  publisher = {Cambridge University Press},
  year      = {2023},
  doi       = {10.1017/9781009414319}
}

@article{bao2024ensemble,
  author  = {Bao, Feng and Zhang, Zezhong and Zhang, Guannan},
  title   = {An Ensemble Score Filter for Tracking High-Dimensional Nonlinear Dynamical Systems},
  journal = {Computer Methods in Applied Mechanics and Engineering},
  year    = {2024},
  volume  = {432},
  pages   = {117447},
  doi     = {10.1016/j.cma.2024.117447},
  eprint  = {2309.00983},
  archivePrefix = {arXiv}
}

@inproceedings{xiao2026ldensf,
  author    = {Xiao, Pengpeng and Si, Phillip and Chen, Peng},
  title     = {{LD-EnSF}: Synergizing Latent Dynamics with Ensemble Score Filters for Fast Data Assimilation with Sparse Observations},
  booktitle = {The Fourteenth International Conference on Learning Representations},
  year      = {2026},
  url       = {https://openreview.net/forum?id=AWSVzzhbr7}
}

@article{transue2025flowmatching,
  author     = {Transue, Taos and Chen, Bohan and Takao, So and Wang, Bao},
  title      = {Flow Matching for Efficient and Scalable Data Assimilation},
  journal    = {SIAM/ASA Journal on Uncertainty Quantification},
  year       = {2026},
  note       = {Accepted for publication},
  url        = {https://arxiv.org/abs/2508.13313}
}

@article{zhang2025iensf,
  author     = {Zhang, Zezhong and Bao, Feng and Zhang, Guannan},
  title      = {{IEnSF}: Iterative Ensemble Score Filter for Reducing Error in Posterior Score Estimation in Nonlinear Data Assimilation},
  journal    = {Journal of Computational Physics},
  volume     = {568},
  pages      = {115357},
  year       = {2027},
  doi        = {10.1016/j.jcp.2026.115357}
}

@inproceedings{si2025latentensf,
  author    = {Si, Phillip and Chen, Peng},
  title     = {{Latent-EnSF}: A Latent Ensemble Score Filter for High-Dimensional Data Assimilation with Sparse Observation Data},
  booktitle = {The Thirteenth International Conference on Learning Representations},
  year      = {2025},
  url       = {https://openreview.net/forum?id=urcEYsZOBz}
}

@misc{chung2023diffusion,
  author    = {Chung, Hyungjin and Kim, Jeongsol and McCann, Michael Thompson and Klasky, Marc Louis and Ye, Jong Chul},
  title     = {Diffusion Posterior Sampling for General Noisy Inverse Problems},
  howpublished = {International Conference on Learning Representations},
  year      = {2023},
  url       = {https://openreview.net/forum?id=OnD9zGAGT0k},
  eprint    = {2209.14687},
  archivePrefix = {arXiv}
}

@article{snyder2008obstacles,
  author  = {Snyder, Chris and Bengtsson, Thomas and Bickel, Peter and Anderson, Jeff},
  title   = {Obstacles to High-Dimensional Particle Filtering},
  journal = {Monthly Weather Review},
  year    = {2008},
  volume  = {136},
  number  = {12},
  pages   = {4629--4640},
  doi     = {10.1175/2008MWR2529.1}
}

@article{delmoral2006sequential,
  author  = {Del Moral, Pierre and Doucet, Arnaud and Jasra, Ajay},
  title   = {Sequential Monte Carlo Samplers},
  journal = {Journal of the Royal Statistical Society Series B: Statistical Methodology},
  year    = {2006},
  volume  = {68},
  number  = {3},
  pages   = {411--436},
  doi     = {10.1111/j.1467-9868.2006.00553.x}
}

@article{gilks2001following,
  author  = {Gilks, Walter R. and Berzuini, Carlo},
  title   = {Following a Moving Target--Monte Carlo Inference for Dynamic Bayesian Models},
  journal = {Journal of the Royal Statistical Society Series B: Statistical Methodology},
  year    = {2001},
  volume  = {63},
  number  = {1},
  pages   = {127--146},
  doi     = {10.1111/1467-9868.00280}
}

@article{bengtsson2008curse,
  author  = {Bengtsson, Thomas and Bickel, Peter and Li, Bo},
  title   = {Curse-of-Dimensionality Revisited: Collapse of the Particle Filter in Very Large Scale Systems},
  journal = {Institute of Mathematical Statistics Collections},
  year    = {2008},
  volume  = {2},
  pages   = {316--334},
  doi     = {10.1214/193940307000000518},
  eprint  = {0805.3034},
  archivePrefix = {arXiv}
}

@book{doucet2001sequential,
  editor    = {Doucet, Arnaud and de Freitas, Nando and Gordon, Neil},
  title     = {Sequential Monte Carlo Methods in Practice},
  publisher = {Springer New York},
  year      = {2001},
  series    = {Information Science and Statistics},
  doi       = {10.1007/978-1-4757-3437-9},
  isbn      = {978-0-387-95146-1}
}

@article{neal2001annealed,
  author  = {Neal, Radford M.},
  title   = {Annealed Importance Sampling},
  journal = {Statistics and Computing},
  year    = {2001},
  volume  = {11},
  number  = {2},
  pages   = {125--139},
  doi     = {10.1023/A:1008923215028},
  eprint  = {physics/9803008},
  archivePrefix = {arXiv}
}

@article{liu1998sequential,
  author  = {Liu, Jun S. and Chen, Rong},
  title   = {Sequential Monte Carlo Methods for Dynamic Systems},
  journal = {Journal of the American Statistical Association},
  year    = {1998},
  volume  = {93},
  number  = {443},
  pages   = {1032--1044},
  doi     = {10.1080/01621459.1998.10473765}
}

@article{sivashinsky1977nonlinear,
  author  = {Sivashinsky, G. I.},
  title   = {Nonlinear Analysis of Hydrodynamic Instability in Laminar Flames--I. Derivation of Basic Equations},
  journal = {Acta Astronautica},
  year    = {1977},
  volume  = {4},
  number  = {11--12},
  pages   = {1177--1206},
  doi     = {10.1016/0094-5765(77)90096-0}
}

@misc{song2021scorebased,
  author    = {Song, Yang and Sohl-Dickstein, Jascha and Kingma, Diederik P. and Kumar, Abhishek and Ermon, Stefano and Poole, Ben},
  title     = {Score-Based Generative Modeling through Stochastic Differential Equations},
  howpublished = {International Conference on Learning Representations},
  year      = {2021},
  url       = {https://openreview.net/forum?id=PxTIG12RRHS},
  eprint    = {2011.13456},
  archivePrefix = {arXiv}
}

@article{efron2011tweedies,
  author  = {Efron, Bradley},
  title   = {Tweedie's Formula and Selection Bias},
  journal = {Journal of the American Statistical Association},
  year    = {2011},
  volume  = {106},
  number  = {496},
  pages   = {1602--1614},
  doi     = {10.1198/jasa.2011.tm11181}
}

@article{boys2023tweedie,
  author     = {Boys, Benjamin and Girolami, Mark and Pidstrigach, Jakiw and Reich, Sebastian and Mosca, Alan and Akyildiz, O. Deniz},
  title      = {Tweedie Moment Projected Diffusions for Inverse Problems},
  journal    = {Transactions on Machine Learning Research},
  year       = {2024},
  url        = {https://openreview.net/forum?id=4unJi0qrTE}
}

@inproceedings{peng2024posteriorcovariance,
  author     = {Peng, Xinyu and Zheng, Ziyang and Dai, Wenrui and Xiao, Nuoqian and Li, Chenglin and Zou, Junni and Xiong, Hongkai},
  title      = {Improving Diffusion Models for Inverse Problems Using Optimal Posterior Covariance},
  booktitle  = {Proceedings of the 41st International Conference on Machine Learning},
  series     = {Proceedings of Machine Learning Research},
  volume     = {235},
  pages      = {40347--40370},
  year       = {2024},
  publisher  = {PMLR},
  url        = {https://proceedings.mlr.press/v235/peng24h.html}
}

@inproceedings{patsenker2025injecting,
  author     = {Patsenker, Jonathan and Li, Henry and Ko, Myeongseob and Jia, Ruoxi and Kluger, Yuval},
  title      = {Injecting Measurement Information Yields a Fast and Noise-Robust Diffusion-Based Inverse Problem Solver},
  booktitle  = {Proceedings of the 29th International Conference on Artificial Intelligence and Statistics},
  series     = {Proceedings of Machine Learning Research},
  volume     = {300},
  pages      = {2323--2331},
  year       = {2026},
  publisher  = {PMLR},
  url        = {https://proceedings.mlr.press/v300/patsenker26a.html}
}

@article{bao2024scorefilter,
  author  = {Bao, Feng and Zhang, Zezhong and Zhang, Guannan},
  title   = {A Score-Based Filter for Nonlinear Data Assimilation},
  journal = {Journal of Computational Physics},
  year    = {2024},
  volume  = {514},
  pages   = {113207},
  doi     = {10.1016/j.jcp.2024.113207}
}

@misc{binder2026closedform,
  author        = {Binder, Brianna and Dasgupta, Agnimitra and Oberai, Assad},
  title         = {Closed-Form Conditional Diffusion Models for Data Assimilation},
  year          = {2026},
  eprint        = {2603.21291},
  archivePrefix = {arXiv},
  primaryClass  = {stat.ML},
  url           = {https://arxiv.org/abs/2603.21291}
}

@inproceedings{rozet2023scorebased,
  author    = {Rozet, Fran{\c{c}}ois and Louppe, Gilles},
  title     = {Score-Based Data Assimilation},
  booktitle = {Advances in Neural Information Processing Systems},
  year      = {2023},
  volume    = {36},
  pages     = {40521--40541},
  doi       = {10.52202/075280-1763},
  url       = {https://proceedings.neurips.cc/paper_files/paper/2023/hash/7f7fa581cc8a1970a4332920cdf87395-Abstract-Conference.html}
}

@article{agapiou2017importance,
  author  = {Agapiou, Sergios and Papaspiliopoulos, Omiros and Sanz-Alonso, Daniel and Stuart, Andrew M.},
  title   = {Importance Sampling: Intrinsic Dimension and Computational Cost},
  journal = {Statistical Science},
  year    = {2017},
  volume  = {32},
  number  = {3},
  pages   = {405--431},
  doi     = {10.1214/17-STS611}
}

@article{beskos2014stability,
  author  = {Beskos, Alexandros and Crisan, Dan and Jasra, Ajay},
  title   = {On the Stability of Sequential Monte Carlo Methods in High Dimensions},
  journal = {The Annals of Applied Probability},
  year    = {2014},
  volume  = {24},
  number  = {4},
  pages   = {1396--1445},
  doi     = {10.1214/13-AAP951}
}

@article{chatterjee2018sample,
  author  = {Chatterjee, Sourav and Diaconis, Persi},
  title   = {The Sample Size Required in Importance Sampling},
  journal = {The Annals of Applied Probability},
  year    = {2018},
  volume  = {28},
  number  = {2},
  pages   = {1099--1135},
  doi     = {10.1214/17-AAP1326}
}

@inproceedings{lorenz1996predictability,
  author    = {Lorenz, Edward N.},
  title     = {Predictability: A Problem Partly Solved},
  booktitle = {Seminar on Predictability},
  year      = {1996},
  volume    = {1},
  pages     = {1--18},
  publisher = {European Centre for Medium-Range Weather Forecasts},
  address   = {Reading, United Kingdom},
  url       = {https://www.ecmwf.int/en/elibrary/75462-predictability-problem-partly-solved}
}

@article{gneiting2007strictly,
  author  = {Gneiting, Tilmann and Raftery, Adrian E.},
  title   = {Strictly Proper Scoring Rules, Prediction, and Estimation},
  journal = {Journal of the American Statistical Association},
  year    = {2007},
  volume  = {102},
  number  = {477},
  pages   = {359--378},
  doi     = {10.1198/016214506000001437}
}

@article{tang2026scoreenhanced,
  author = {Tang, Jingqiao and Bausback, Ryan and Bao, Feng and Zhang, Guannan and Huynh, Phuoc-Toan},
  title = {A Score-Filter-Enhanced Data Assimilation Framework for Data-Driven Dynamical Systems},
  journal = {Numerical Methods for Partial Differential Equations},
  volume = {42},
  number = {5},
  pages = {e70138},
  year = {2026},
  doi = {10.1002/num.70138}
}

@article{bao2025sqg,
  author = {Bao, Feng and Chipilski, Hristo G. and Liang, Siming and Zhang, Guannan and Whitaker, Jeffrey S.},
  title = {Nonlinear Ensemble Filtering with Diffusion Models: Application to the Surface Quasigeostrophic Dynamics},
  journal = {Monthly Weather Review},
  volume = {153},
  number = {7},
  pages = {1155--1169},
  year = {2025},
  doi = {10.1175/MWR-D-24-0069.1}
}

@article{huynh2026spde,
  author = {Huynh, Phuoc-Toan and L{\'o}pez Fajardo, Ruth Y. and Zhang, Guannan and Ju, Lili and Bao, Feng},
  title = {A Score-Based Diffusion Model Approach for Adaptive Learning of Stochastic Partial Differential Equation Solutions},
  journal = {Journal of Computational Physics},
  volume = {556},
  pages = {114814},
  year = {2026},
  doi = {10.1016/j.jcp.2026.114814}
}

@inproceedings{chen2025flowdas,
  author    = {Chen, Siyi and Jia, Yixuan and Qu, Qing and Sun, He and Fessler, Jeffrey A.},
  title     = {{FlowDAS}: A Stochastic Interpolant-based Framework for Data Assimilation},
  booktitle = {Advances in Neural Information Processing Systems},
  volume    = {38},
  year      = {2025},
  doi       = {10.52202/085713-3445},
  url       = {https://proceedings.neurips.cc/paper_files/paper/2025/hash/951690f8dae19c76394404540b7ca734-Abstract-Conference.html}
}

@inproceedings{andrae2026daisi,
  author    = {Andrae, Martin and Wikingsson, Erik and Takao, So and Landelius, Tomas and Lindsten, Fredrik},
  title     = {{DAISI}: Data Assimilation with Inverse Sampling using Stochastic Interpolants},
  booktitle = {Proceedings of the 43rd International Conference on Machine Learning},
  series    = {Proceedings of Machine Learning Research},
  volume    = {306},
  pages     = {2787--2830},
  year      = {2026},
  publisher = {PMLR},
  url       = {https://proceedings.mlr.press/v306/andrae26a.html}
}

@inproceedings{wei2026surge,
  author    = {Wei, Lifu and Ren, Yinuo and Shi, Naichen and Lu, Yiping},
  title     = {{SURGE}: Approximation and Training Free Particle Filter for Diffusion Surrogate},
  booktitle = {Proceedings of the 43rd International Conference on Machine Learning},
  series    = {Proceedings of Machine Learning Research},
  volume    = {306},
  pages     = {134046--134074},
  year      = {2026},
  publisher = {PMLR},
  url       = {https://proceedings.mlr.press/v306/wei26w.html}
}

@inproceedings{savary2026generative,
  author    = {Savary, Thomas and Rozet, Fran{\c{c}}ois and Louppe, Gilles},
  title     = {Training-Free {Bayesian} Filtering with Generative Emulators},
  booktitle = {Proceedings of the 43rd International Conference on Machine Learning},
  series    = {Proceedings of Machine Learning Research},
  volume    = {306},
  pages     = {107889--107912},
  year      = {2026},
  publisher = {PMLR},
  url       = {https://proceedings.mlr.press/v306/savary26a.html}
}

@article{pitt1999auxiliary,
  author    = {Pitt, Michael K. and Shephard, Neil},
  title     = {Filtering via Simulation: Auxiliary Particle Filters},
  journal   = {Journal of the American Statistical Association},
  volume    = {94},
  number    = {446},
  pages     = {590--599},
  year      = {1999},
  doi       = {10.1080/01621459.1999.10474153}
}

@article{roberts1996langevin,
  author    = {Roberts, Gareth O. and Tweedie, Richard L.},
  title     = {Exponential Convergence of Langevin Distributions and Their Discrete Approximations},
  journal   = {Bernoulli},
  volume    = {2},
  number    = {4},
  pages     = {341--363},
  year      = {1996},
  doi       = {10.2307/3318418}
}
\bibliographystyle{iclr2027_conference}

\appendix

\section{Filtering model and notation}
\label{sec:problem_formulation}

The index \(n\) counts assimilation cycles. Within each cycle, diffusion time \(\tau\) runs backward during sampling. The transition \(p_n\) propagates the state \(x_n\in\R^d\), and the likelihood \(g_n(y_n\given x_n)\) describes the observation \(y_n\in\R^{m_n}\). Before assimilating \(y_n\), the forecast density \(\pi_n^f\) conditions on \(y_{1:n-1}\). After the update, the analysis density \(\pi_n^a\) also conditions on \(y_n\), as in \eqref{eq:main_analysis_recursion}. AECSF uses likelihood values to weight proposals and likelihood gradients to move them.

Within one assimilation step, \(Z\) denotes the system state before artificial noise is added.
The variable \(z\) denotes a possible value of the system state; \(z_i^m\) denotes a member of the proposal ensemble.
The joint model in \eqref{eq:main_vp_forward} defines the noisy state \(X_\tau\), whose conditional density given \(y_n\) is \(q_{n,\tau}^a\).
Its score \(s_{n,\tau}^a\) is determined by the conditional mean \(D_{n,\tau}\) through \eqref{eq:main_conditional_tweedie}.

For the main training-free benchmarks, the forecast ensemble \(\mathcal X_n^f\) initializes the proposal ensemble with \(M=J\) and one proposal at each forecast member. Table~\ref{tab:l96_proposal_capacity} studies a different proposal count, and Appendix~\ref{app:ns_learned_thirty} describes initialization for the learned-prior extension.
The \(J\) reverse particles \(x_\tau^{(j)}\) share the proposal ensemble but assign its members different normalized conditional weights.
Their final states form the analysis ensemble \(\mathcal X_n^a\).
The tempered density \(\pi_{n,\rho}\) specifies the sampling distribution assumed in the importance-sampling derivation.
Appendix~\ref{sec:ecsf} gives the derivation and implementation details; Appendix~\ref{sec:theory} states the assumptions, consistency under independent sampling, and conditional-mean error bounds.

\section{Deriving the conditional score and implementing AECSF}
\label{sec:ecsf}

\subsection{Exact conditional Tweedie identity}
\label{subsec:conditional_tweedie_derivation}

We first derive the score when the forecast distribution is known exactly. The result is a conditional form of Tweedie's identity \citep{efron2011tweedies,boys2023tweedie,peng2024posteriorcovariance,patsenker2025injecting}; the next subsection constructs the conditional-mean estimator from the proposal ensemble.

Consider the joint model
\begin{align}
    Z &\sim \pi_n^f,\label{eq:joint_x0}\\
    Y_n\given Z &\sim g_n(\cdot\given Z),\label{eq:joint_y}\\
    X_\tau\given Z &\sim \N(a_\tau Z,\sigma_\tau^2\I_d).\label{eq:joint_xtau}
\end{align}
The artificial diffusion noise is independent of the observation conditional on the system state \(Z\).
Conditioning on \(Y_n=y_n\) changes the law of \(Z\) from the forecast density \(\pi_n^f\) to the analysis density \(\pi_n^a\).
The corresponding conditional density of \(X_\tau\) is therefore the posterior after Gaussian noising, \(q_{n,\tau}^a(\cdot\mid y_n)\).

\paragraph{Proposition 1 (conditional Tweedie identity).}
Under the conditions in Appendix~\ref{subsec:theory_assumptions}, for any \(\tau\in(0,1]\) with \(a_\tau>0\) and \(\sigma_\tau>0\), the conditional score satisfies
\begin{equation}
    s_{n,\tau}^a(x;y_n)
    =
    -\frac{x}{\sigma_\tau^2}
    +
    \frac{a_\tau}{\sigma_\tau^2}D_{n,\tau}(x;y_n),
    \label{eq:conditional_tweedie_identity}
\end{equation}
where
\begin{equation}
    D_{n,\tau}(x;y_n):=\E[Z\given X_\tau=x,Y_n=y_n].
    \label{eq:denoising_mean_def}
\end{equation}
To see this, write the Gaussian-smoothed analysis density as
\begin{equation}
    q_{n,\tau}^a(x\mid y_n)
    =
    \frac{1}{C_n(y_n)}\int
    \phi_{\sigma_\tau^2\I_d}(x-a_\tau z)g_n(y_n\given z)\pi_n^f(z)\,\dd z,
    \label{eq:noisy_analysis_population}
\end{equation}
where \(C_n(y_n)\) is independent of \(x\). The integral gives the density of the noisy state \(X_\tau\) conditional on \(Y_n=y_n\). For a fixed noisy state \(x\), the conditional density of the system state is
\begin{equation}
    p(z\mid X_\tau=x,Y_n=y_n)
    =
    \frac{
    \phi_{\sigma_\tau^2\I_d}(x-a_\tau z)g_n(y_n\given z)\pi_n^f(z)
    }{
    \int \phi_{\sigma_\tau^2\I_d}(x-a_\tau u)g_n(y_n\given u)\pi_n^f(u)\,\dd u
    }.
    \label{eq:conditional_clean_state_density}
\end{equation}
The conditional mean \(D_{n,\tau}(x;y_n)\) in \eqref{eq:denoising_mean_def} is the first moment of \eqref{eq:conditional_clean_state_density}.
Differentiating \eqref{eq:noisy_analysis_population} under the integral and dividing by \(q_{n,\tau}^a(x\mid y_n)\) yields
\begin{equation}
    \nabla_x\log q_{n,\tau}^a(x\mid y_n)
    =
    -\frac{x}{\sigma_\tau^2}
    +
    \frac{a_\tau}{\sigma_\tau^2}
    \frac{
    \int z\,\phi_{\sigma_\tau^2\I_d}(x-a_\tau z)g_n(y_n\given z)\pi_n^f(z)\,\dd z
    }{
    \int \phi_{\sigma_\tau^2\I_d}(x-a_\tau z)g_n(y_n\given z)\pi_n^f(z)\,\dd z
    },
    \label{eq:tweedie_population_derivation}
\end{equation}
which is exactly \eqref{eq:conditional_tweedie_identity}.

\subsection{Estimating the conditional mean from the proposals}
\label{subsec:ecsf_score}

At reverse step \(i\), fix the proposal locations and weights in the empirical measure \(\nu_{n,i}^M\) from \eqref{eq:main_candidate_measure}. We can then evaluate the score in \eqref{eq:main_estimated_score} at any noisy state \(x\). The factor in \eqref{eq:main_ecsf_estimator} is
\begin{equation}
    F_i(x,z)
    =
    K_i(x,z)\,g_n(y_n\given z)^{1-\rho_i}.
\end{equation}
Whenever \(\nu_{n,i}^M(F_i(x,\cdot))>0\), its conditional-mean estimate is the weighted moment
\begin{equation}
    D_{\nu_{n,i}^M}(x;y_n)
    =
    \frac{\int z F_i(x,z)\,\nu_{n,i}^M(\dd z)}{\int F_i(x,z)\,\nu_{n,i}^M(\dd z)}
    =
    \sum_{m=1}^M c_m(x)z_i^m.
    \label{eq:empirical_bank_moment_ratio}
\end{equation}
For the reverse-particle queries, this is \(\widehat D_{n,i}^M\) in \eqref{eq:main_ecsf_estimator}. The corresponding log coefficients are
\begin{equation}
    b_{j,m}(\tau_i)
    =
    \log W_{i,m}-\frac{\norm{x_i^{(j)}-a_{\tau_i}z_i^m}^2}{2\sigma_{\tau_i}^2}
    +(1-\rho_i)\log g_n(y_n\given z_i^m),
    \label{eq:active_logweight}
\end{equation}
followed by log-sum-exp normalization over \(m\).
The resulting score estimate is
\begin{equation}
    \widehat s_{n,\tau_i}^{a}(x;y_n)
    =
    -\frac{x}{\sigma_{\tau_i}^2}
    +
    \frac{a_{\tau_i}}{\sigma_{\tau_i}^2}
    D_{\nu_{n,i}^M}(x;y_n).
    \label{eq:ecsf_score}
\end{equation}

The likelihood-tilted empirical measure \(\widehat\eta_{n,i}^M\) in \eqref{eq:main_empirical_surrogate} defines the Gaussian mixture
\begin{equation}
    \widehat q_{n,i}^M(x;y_n)
    =
    \int \phi_{\sigma_{\tau_i}^2\I_d}(x-a_{\tau_i}z)
    \widehat\eta_{n,i}^M(\dd z).
    \label{eq:empirical_noisy_surrogate}
\end{equation}
Holding the proposal ensemble fixed while differentiating with respect to \(x\) gives
\begin{equation}
    \nabla_x\log\widehat q_{n,i}^M(x;y_n)
    =
    -\frac{x}{\sigma_{\tau_i}^2}
    +
    \frac{a_{\tau_i}}{\sigma_{\tau_i}^2}
    D_{\nu_{n,i}^M}(x;y_n).
    \label{eq:frozen_bank_score_derivation}
\end{equation}

\subsection{Resampling and Langevin steps}
\label{subsec:active_refresh}

Stratified ancestor sampling uses independent offsets
\(U_\ell\sim\operatorname{Unif}((\ell-1)/M,\ell/M)\), followed by
\(A_\ell=\min\{m:\sum_{r=1}^m s_{i,r}\geq U_\ell\}\).

For the incoming proposal ensemble in Algorithm~\ref{alg:ecsf_analysis}, let \(r_j=(r_{i,j1},\ldots,r_{i,jM})\) be the ancestor-selection distribution for reverse particle \(j\) in \eqref{eq:refresh_resampling}. Its average \(s_i=J^{-1}\sum_j r_j\) minimizes the mean relative entropy:
\begin{equation}
    s_i
    =
    \operatorname*{arg\,min}_{p\in\Delta_M}
    \frac{1}{J}\sum_{j=1}^J\operatorname{KL}(r_j\Vert p),
    \qquad
    \Delta_M=\left\{p\in[0,1]^M:\sum_m p_m=1\right\}.
    \label{eq:refresh_kl_barycenter}
\end{equation}
Indeed,
\begin{equation}
    \frac{1}{J}\sum_{j=1}^J\operatorname{KL}(r_j\Vert p)
    =C+\operatorname{KL}(s_i\Vert p),
\end{equation}
where \(C\) is independent of \(p\). Thus one distribution \(s_i\) summarizes the ancestor preferences of all reverse particles. For score estimation, each particle \(j\) uses its own weights \(c_{j,m}\), which include the residual likelihood in \eqref{eq:main_ecsf_estimator}.

\paragraph{Conditional expectation under ancestor correction.}
Condition on the incoming positions, target weights \(\omega_i\), and reverse queries, and let \(\mathcal F_i\) denote this information. Assume \(s_{i,m}>0\) whenever \(\omega_{i,m}>0\), and that the test function \(\varphi\) is integrable under the target empirical measure. If the offspring counts \(N_m\) satisfy \(\mathbb E[N_m\mid\mathcal F_i]=Ms_{i,m}\), then
\begin{equation}
\mathbb E\!\left[\frac1M\sum_{\ell=1}^M
\frac{\omega_{i,A_\ell}}{s_{i,A_\ell}}\varphi(z_{A_\ell}^-)
\,\middle|\,\mathcal F_i\right]
=\sum_m\omega_{i,m}\varphi(z_m^-).
\label{eq:ancestor_conditional_expectation}
\end{equation}
To verify this, group descendants by their ancestor. The left side becomes
\[
\frac1M\sum_m\mathbb E[N_m\mid\mathcal F_i]
\frac{\omega_{i,m}}{s_{i,m}}\varphi(z_m^-),
\]
which gives the stated sum. Stratified resampling has the required expected counts, using an independent uniform draw within each stratum. Taking \(\varphi=1\) gives expected total mass one. Normalizing the realized weights yields a finite-sample ratio estimator of the discrete target integral.

\paragraph{Effect of the move kernel.}
Moving the descendants can change how well their weighted ensemble represents the tempered target. To see where the error comes from, let \(\mu_i=\sum_m\omega_{i,m}\delta_{z_m^-}\) be the measure before resampling. Conditional on \(\mathcal F_i\), the move kernel \(Q_i\) has fixed parameters. In the Gaussian implementation we fix the forecast fit and move parameters before ancestor selection. A descendant of \(z_{A_\ell}^-\) moves according to \(Q_i(z_{A_\ell}^-,\cdot)\) and retains its unnormalized weight \(v_\ell=\omega_{i,A_\ell}/s_{i,A_\ell}\). Write \(\widetilde\nu_i^+(\varphi)=M^{-1}\sum_\ell v_\ell\varphi(Z_\ell^+)\) for the resulting weighted sum. Under the preceding support and integrability conditions, we average first over the moves and then over the offspring counts to obtain
\begin{align}
\E[\widetilde\nu_i^+(\varphi)\mid\mathcal F_i]&=\mu_i(Q_i\varphi),\label{eq:move_weighted_expectation}\\
\E[\widetilde\nu_i^+(\varphi)\mid\mathcal F_i]-\pi_{n,\rho_i}(\varphi)
&=(\mu_i-\pi_{n,\rho_i})(Q_i\varphi)
+\pi_{n,\rho_i}(Q_i\varphi-\varphi).\label{eq:move_error_decomposition}
\end{align}
The first term carries forward any discrepancy already present in \(\mu_i\). The second asks what the move would do even if the proposals started at the tempered target. It vanishes after any finite number of steps when the kernel preserves that target. Approximating the forecast or discretizing the Langevin dynamics can make this term nonzero.

To apply the decomposition to the conditional mean of particle \(j\), take \(\varphi(z)=F_i(x_i^{(j)},z)\) for the denominator. For the numerator, multiply this function by each coordinate of \(z\). This gives their expected errors before normalization. The conditional mean is their ratio, so its error also depends on the random denominator (Appendix~\ref{subsec:weighted_moment_error}).

The reference Langevin steps in \eqref{eq:rho_tempered_move} use the forecast-score approximation and likelihood gradient. If \(\widehat s_n^f=\nabla\log\widetilde\pi_n^f\), their drift direction is the gradient of
\begin{equation}
    \log\left[\widetilde\pi_n^f(z)\,g_n(y_n\given z)^{\rho_i}\right].
    \label{eq:move_drift_density}
\end{equation}

The diagonal approximation supplies coordinatewise prior gradients without estimating a full covariance matrix. Short Langevin moves adjust the proposals within the sampling budget; their contribution to filtering performance is evaluated by the component controls.

\subsection{Importance sampling with a tempered proposal density}
\label{subsec:tempered_representation}

For comparison, the conditional distribution of the system state is
\begin{equation}
    p(z\mid X_\tau=x,Y_n=y_n)
    \propto
    \phi_{\sigma_\tau^2\I_d}(x-a_\tau z)\pi_n^f(z)g_n(y_n\given z).
    \label{eq:target_conditional_mu}
\end{equation}
For normalized diffusion time \(\tau\in[0,1]\), the main training-free benchmarks use
\begin{equation}
    \rho_\tau=1-\tau.
    \label{eq:rho_linear}
\end{equation}
For any fixed \(\rho\in[0,1]\), define the tempered proposal density
\begin{equation}
    \pi_{n,\rho}(z)
    \propto
    \pi_n^f(z)g_n(y_n\given z)^\rho.
    \label{eq:tempered_clean_distribution}
\end{equation}
Factoring out the tempered density gives
\begin{equation}
    p(z\mid X_\tau=x,Y_n=y_n)
    \propto
    \pi_{n,\rho}(z)
    \underbrace{\phi_{\sigma_\tau^2\I_d}(x-a_\tau z)}_{\text{Gaussian noising}}
    \underbrace{g_n(y_n\given z)^{1-\rho}}_{\text{remaining likelihood}}.
    \label{eq:tempered_target_factorization}
\end{equation}
Because \(g_n^\rho g_n^{1-\rho}=g_n\), changing \(\rho\) leaves the conditional distribution unchanged. If \(z^m\) are independent samples from \(\pi_{n,\rho}\), the conditional mean \(D_{n,\tau}(x;y_n)\) can be estimated by
\begin{equation}
    \widehat D_{n,\tau,\rho}^{\rm IS}(x;y_n)
    =
    \sum_{m=1}^M \omega_m^{(\rho)}(\tau,x)z^m,
    \label{eq:tempered_Dhat}
\end{equation}
where
\begin{align}
    b_m^{(\rho)}(\tau,x)
    &=
    -\frac{\norm{x-a_\tau z^m}^2}{2\sigma_\tau^2}
    +(1-\rho)\log g_n(y_n\given z^m),\label{eq:tempered_log_weight}\\
    \omega_m^{(\rho)}(\tau,x)
    &=
    \frac{\exp(b_m^{(\rho)}(\tau,x)-c)}
    {\sum_{\ell=1}^M\exp(b_\ell^{(\rho)}(\tau,x)-c)},
    \qquad
    c=\max_\ell b_\ell^{(\rho)}(\tau,x).
    \label{eq:tempered_normalized_weight}
\end{align}
Here the coefficients are self-normalized importance weights because the samples come from \(\pi_{n,\rho}\). Larger \(\rho\) places more likelihood information in that proposal and leaves less in the residual weight. Such likelihood-tempering paths are also used in sequential Monte Carlo and annealed importance sampling \citep{delmoral2006sequential,beskos2014stability,neal2001annealed}.

\subsection{Using reverse diffusion in the filtering cycle}
\label{subsec:reverse_recursive}

Algorithm~\ref{alg:ecsf_analysis} gives the training-free analysis update. For the differentiable Gaussian noising schedule, define
\begin{equation}
    f_\tau=\partial_\tau\log a_\tau,\qquad
    r_\tau^2=\partial_\tau\sigma_\tau^2-2f_\tau\sigma_\tau^2,\qquad
    \Delta_i=\tau_i-\tau_{i-1}.
    \label{eq:main_reverse_coefficients}
\end{equation}
The schedule is chosen so that \(r_\tau^2\geq0\).
The main training-free recursive benchmarks use
\begin{equation}
    a_\tau=1-(1-\epsilon_a)\tau,\qquad
    \sigma_\tau^2=\epsilon_b+(1-\epsilon_b)\tau,
    \label{eq:main_linear_noise_schedule}
\end{equation}
with \(0<\epsilon_a,\epsilon_b<1\). Thus \(a_0=1\) and
\(\sigma_0^2=\epsilon_b>0\), and the diffusion rate in
\eqref{eq:main_reverse_coefficients} is nonnegative.
The grid is uniform in logSNR:
\begin{equation}
    \lambda(\tau)=\log\frac{a_\tau^2}{\sigma_\tau^2},\qquad
    \lambda(\tau_i)=\lambda(1)+\frac{N_\tau-i}{N_\tau}[\lambda(0)-\lambda(1)].
    \label{eq:logsnr_mesh}
\end{equation}
Inverting this relation gives \(\tau_i\), with
\(\rho_i=1-\tau_i\) and \(\Delta_i=\tau_i-\tau_{i-1}\).
The full logSNR convention differs by a constant factor of two from
\(\log(a_\tau/\sigma_\tau)\) and generates the same uniform grid.

The reverse particles are advanced by \eqref{eq:main_reverse_update}.
At each assimilation time, independent standard-normal entries are drawn and then standardized componentwise across the \(J\) particles to zero mean and unit sample standard deviation. This standardization makes the initialized members dependent; Proposition C.3 quantifies its member-marginal discrepancy from the terminal noised posterior.
The conditional-score estimator can also be evaluated at prescribed query states without running the reverse solver.

\begin{algorithm}[H]
\caption{One AECSF analysis update for the main training-free \(M=J\) configuration}
\label{alg:ecsf_analysis}
\begin{algorithmic}[1]
\Require Forecast ensemble \(\mathcal X_n^f\), observation \(y_n\), likelihood \(g_n\), fitted score \(\widehat s_n^f\), initial reverse particles, diffusion grid, resampling and move schedules
\State Initialize proposals at forecast members; set \(W_m=1/M\), \(\rho^-=0\), and cache likelihoods
\For{\(i=N_\tau,\ldots,1\)}
    \State Set \(\rho_i=1-\tau_i\)
    \If{\(i<N_\tau\)}
        \State Compute \(\omega_i\) by \eqref{eq:incremental_target_weights}; set \(W\gets\omega_i\)
        \If{resampling is scheduled}
            \State Compute the shared resampling probabilities \(s_i\) using \eqref{eq:refresh_resampling}
            \State Draw stratified ancestors from \(s_i\) and correct weights by \eqref{eq:ancestor_correction}
        \EndIf
        \If{movement is scheduled}
            \State Apply \eqref{eq:rho_tempered_move}; retain weights and refresh likelihoods
        \EndIf
        \State Set \(\rho^-\gets\rho_i\)
    \EndIf
    \State Compute conditional means and scores by \eqref{eq:main_ecsf_estimator}--\eqref{eq:main_estimated_score}
    \State Advance reverse particles by \eqref{eq:main_reverse_update}
\EndFor
\State \Return \(\mathcal X_n^a=\{x_0^{(j)}\}_{j=1}^J\)
\end{algorithmic}
\end{algorithm}

\begin{algorithm}[H]
\caption{Recursive filtering with the training-free AECSF update}
\label{alg:recursive_ecsf}
\begin{algorithmic}[1]
\Require Initial analysis ensemble \(\mathcal X_0^a\), observations \(\{y_n\}_{n=1}^{N_a}\), forecast model \(\mathcal M_n\), differentiable likelihoods \(g_n\), reverse-particle initialization rule, and the settings in Algorithm~\ref{alg:ecsf_analysis}
\For{\(n=1,\ldots,N_a\)}
    \State Forecast: propagate \(\mathcal X_{n-1}^a\) through \(\mathcal M_n\) to obtain \(\mathcal X_n^f\)
    \State Analysis: initialize the reverse particles at \(\tau=1\) and apply Algorithm~\ref{alg:ecsf_analysis} to \(\mathcal X_n^f\) and \(y_n\)
\EndFor
\State \Return analysis ensembles \(\{\mathcal X_n^a\}_{n=1}^{N_a}\)
\end{algorithmic}
\end{algorithm}

\subsection{Numerical stability and computational cost}
\label{subsec:active_stabilization_cost}

The implementation computes \eqref{eq:active_logweight} in the log domain and normalizes the conditional kernel coefficients after subtracting their largest log value:
\begin{align}
    c &= \max_m b_m,\\
    \widetilde c_m &= \exp(b_m-c),\\
    c_m &= \widetilde c_m/\sum_\ell \widetilde c_\ell.
\end{align}
The same log-sum-exp operation normalizes the incremental target weights, ancestor selection probabilities, and corrected descendant weights separately.
For finite log coefficients, subtracting their maximum preserves the normalized coefficients and prevents exponential overflow.
At least one unnormalized coefficient equals one.

The reverse score and the combined Langevin move score are clipped
componentwise to the configured score limits. After each proposal move,
non-finite state components are replaced by zero and proposal coordinates
are clipped to the configured state bounds. The reverse integrator also
replaces non-finite state components by zero.

With \(J\) reverse particles, \(M\) proposal members, \(N_\tau\) reverse steps, and state dimension \(d\), evaluating the Gaussian factors and weighted conditional means requires
\begin{equation}
    O(JN_\tau Md)
\end{equation}
arithmetic work. At fixed \(J\), this work grows linearly with \(M\);
when \(M=J\), it grows quadratically with ensemble size.
For the Gaussian kernel, we use
\[
\|x_j-a_\tau z_m\|^2
=\|x_j\|^2+a_\tau^2\|z_m\|^2-2a_\tau x_j^Tz_m.
\]
The implementation applies this identity after centering the states and
computes the pairwise inner products and weighted conditional means by
matrix multiplication on the GPU. This avoids storing a
\(J\times M\times d\) difference tensor and requires
\(O(JM+(J+M)d)\) working storage for these calculations.
If each proposal receives \(L\) Langevin steps and one gradient evaluation
costs \(C_{\nabla}(d)\), their gradient work is
\(O(LM C_{\nabla}(d))\); likelihood evaluation contributes its
observation-model cost separately.
At $T=20$, the measured analysis time for $J=M\in\{10,20,40,80,160\}$ remains within 0.099--0.104 seconds on the RTX 4090 (Figure~\ref{fig:ks_cost_scale}). The arithmetic and storage costs still grow with ensemble size.

\section{Exact identities and approximation error}
\label{sec:theory}

\subsection{Assumptions for the conditional Tweedie identity}
\label{subsec:theory_assumptions}

For the fixed observation, assume that its marginal density is positive and finite:
\begin{equation}
    C_n(y_n)=\int g_n(y_n\given z)\pi_n^f(z)\,\dd z,
    \qquad 0<C_n(y_n)<\infty.
    \label{eq:observation_normalizer}
\end{equation}
This condition ensures that \(\pi_n^a(z)=g_n(y_n\given z)\pi_n^f(z)/C_n(y_n)\) is a probability density.
For fixed \(\tau\) with \(a_\tau>0\) and \(\sigma_\tau>0\), the Gaussian kernel and its first derivative in \(x\) are bounded as functions of \(z\).
For fixed \(x\), the function \(\norm{z}\phi_{\sigma_\tau^2\I_d}(x-a_\tau z)\) is also bounded.
These bounds justify differentiation under the integral and ensure that the conditional first moment is finite.
The derivation in Appendix~\ref{sec:ecsf} therefore yields \eqref{eq:conditional_tweedie_identity}.
The likelihood factorization in \eqref{eq:tempered_target_factorization} leaves this conditional distribution unchanged for every \(\rho\in[0,1]\).

\subsection{Consistency under independent sampling}
\label{subsec:importance_sampling_consistency}

Fix \((\tau,x,y_n,\rho)\), and let \(z^1,\ldots,z^M\) be independent samples from the tempered density \(\pi_{n,\rho}\).
Define the unnormalized importance weight for this fixed query by
\begin{equation}
    W(z)=\phi_{\sigma_\tau^2\I_d}(x-a_\tau z)g_n(y_n\given z)^{1-\rho}.
    \label{eq:ideal_importance_weight}
\end{equation}
Assume \(0<\E_{\pi_{n,\rho}}[W(Z)]<\infty\) and \(\E_{\pi_{n,\rho}}[W(Z)\norm{Z}]<\infty\).
The importance-sampling estimator in \eqref{eq:tempered_Dhat} then converges to the population conditional mean:
\begin{equation}
    \widehat D_{n,\tau,\rho}^{\rm IS}(x;y_n)
    \xrightarrow[M\to\infty]{p}
    D_{n,\tau}(x;y_n).
    \label{eq:tempered_snis_consistency}
\end{equation}
To establish this result, write the estimator as the ratio of sample averages:
\begin{equation}
    \begin{aligned}
    \widehat D_{n,\tau,\rho}^{\rm IS}(x;y_n)
    &=\frac{M^{-1}\sum_{m=1}^M W(z^m)z^m}{M^{-1}\sum_{m=1}^M W(z^m)}\\
    &\xrightarrow[M\to\infty]{p}
    \frac{\E_{\pi_{n,\rho}}[W(Z)Z]}{\E_{\pi_{n,\rho}}[W(Z)]}
    =D_{n,\tau}(x;y_n).
    \end{aligned}
    \label{eq:ideal_snis_proof}
\end{equation}
The law of large numbers applies separately to the numerator and denominator.
Their ratio converges because the limiting denominator is positive and finite.
The final equality follows from the conditional-density factorization in \eqref{eq:tempered_target_factorization}.
Importance sampling becomes more demanding as the target and proposal separate, particularly in high intrinsic dimension \citep{agapiou2017importance,chatterjee2018sample}.

For a different proposal density \(q\) whose support covers the conditional target, the corresponding unnormalized importance weight is instead
\begin{equation}
    W_q(z)
    =
    \frac{
    \phi_{\sigma_\tau^2\I_d}(x-a_\tau z)\pi_n^f(z)g_n(y_n\given z)
    }{q(z)}.
    \label{eq:general_proposal_importance_weight}
\end{equation}
Setting \(q=\pi_{n,\rho}\) reduces \eqref{eq:general_proposal_importance_weight} to \eqref{eq:ideal_importance_weight}, up to a constant that cancels under normalization.

\paragraph{Target condition for a fixed proposal measure.}
\label{subsec:surrogate_target_condition}

Write \(p=\pi_n^f\), \(g(z)=g_n(y_n\given z)\), and \(\pi_\rho(\dd z)\propto p(z)g(z)^\rho\dd z\). Let \(\nu_\tau\) be a probability measure held fixed during score-field evaluation. Assume \(g\) is positive and finite on the supports of \(p\) and \(\nu_\tau\), and all normalizing constants below and for \(\pi_\rho,\pi_1\) are positive and finite. Define
\begin{equation}
\eta_\tau(\dd z)=
\frac{g(z)^{1-\rho}\nu_\tau(\dd z)}{\nu_\tau(g^{1-\rho})},
\qquad
q_\tau^\nu(x)=\int\phi_{\sigma_\tau^2\I_d}(x-a_\tau z)\eta_\tau(\dd z).
\label{eq:frozen_candidate_target}
\end{equation}

\paragraph{Proposition.}
For \(a_\tau\ne0\) and \(\sigma_\tau^2>0\), \(q_\tau^\nu\) equals the noised posterior everywhere if and only if \(\nu_\tau=\pi_\rho\). The same equivalence holds for their score fields.

\paragraph{Proof.}
Gaussian noising maps a measure \(\eta\) to a density with Fourier transform
\[
\widehat q(\xi)=e^{-\sigma_\tau^2\|\xi\|^2/2}\widehat\eta(a_\tau\xi).
\]
The Gaussian multiplier is nonzero and the scaling invertible, so equality of noised densities is equivalent to \(\eta_\tau=\pi_1\). Inverting the likelihood tilt gives
\[
\nu_\tau(\dd z)=
\frac{g(z)^{\rho-1}\pi_1(\dd z)}{\pi_1(g^{\rho-1})}
=\pi_\rho(\dd z).
\]
Conversely, this substitution gives \(\eta_\tau=\pi_1\). Positive normalized noised densities have identical scores exactly when they are equal.

At initialization, \(\rho=0\) and the population choice \(\nu_1=p\) gives \(\eta_1=\pi_1\). Subsequent stages advance the empirical target weights even when resampling is skipped. The weighted-moment analysis below characterizes the discrepancy between the updated proposal measure and the tempered target.

\subsection{Using weighted moments to bound the error in the conditional mean}
\label{subsec:weighted_moment_error}

Fix \((\tau,x,y_n,\rho)\), write \(\pi=\pi_{n,\rho}\), and set
\begin{equation}
    F(z)=\phi_{\sigma_\tau^2\I_d}(x-a_\tau z)g_n(y_n\given z)^{1-\rho}.
\end{equation}
For any probability measure \(\nu\) with \(0<\nu(F)<\infty\) and \(\nu(F\norm{z})<\infty\), define \(D_\nu=\nu(Fz)/\nu(F)\). Since \(D_{n,\tau}(x;y_n)=\pi(Fz)/\pi(F)\), the conditional-mean error obeys the exact identity
\begin{equation}
    D_\nu-D_{n,\tau}(x;y_n)
    =
    \frac{(\nu-\pi)\!\left(F[\,z-D_{n,\tau}(x;y_n)\,]\right)}{\nu(F)}.
    \label{eq:weighted_moment_error_identity}
\end{equation}
Let
\begin{equation}
    \delta_0=\left|(\nu-\pi)(F)\right|,
    \qquad
    \delta_1=\left\|(\nu-\pi)(Fz)\right\|.
\end{equation}
If \(\delta_0<\pi(F)\), then
\begin{equation}
    \left\|D_\nu-D_{n,\tau}(x;y_n)\right\|
    \le
    \frac{\delta_1+\norm{D_{n,\tau}(x;y_n)}\delta_0}
    {\pi(F)-\delta_0}.
    \label{eq:weighted_moment_error_bound}
\end{equation}
Under the stated moment conditions and \(\delta_0<\pi(F)\), this bound also applies to the empirical measure \(\nu_{n,i}^M\) of the proposal ensemble. Thus, convergence of both weighted moments is sufficient for convergence of the conditional-mean estimate at the fixed query.

Let \(\widehat s_{\nu,n,\tau}^{a}\) denote the corresponding field. At a fixed query state, its error follows directly from \eqref{eq:conditional_tweedie_identity}:
\begin{equation}
    \widehat s_{\nu,n,\tau}^{a}(x;y_n)-s_{n,\tau}^{a}(x;y_n)
    =\frac{a_\tau}{\sigma_\tau^2}
    \left[D_\nu-D_{n,\tau}(x;y_n)\right].
    \label{eq:score_mean_error_relation}
\end{equation}
Resampling and Langevin steps affect the weighted moments in \eqref{eq:weighted_moment_error_identity}. The recursive filter also introduces time discretization, terminal initialization, and a retained final noise variance, as specified in Appendix~\ref{subsec:reverse_recursive}.

\subsection{How proposal errors propagate through reverse sampling}
\label{app:adaptive_candidate_theory}
We follow one assimilation cycle in the order the algorithm executes it. First we track the error in the weighted proposals, then its effect on the conditional means, and finally its effect on the reverse particles.
Let \(k=1,\ldots,K\) count the \(K=N_\tau\) reverse steps. Step \(k\) starts at
\(\theta_k=\tau_{N_\tau-k+1}\) and has length \(\delta_k=\tau_{N_\tau-k+1}-\tau_{N_\tau-k}\).
In Algorithm~\ref{alg:ecsf_analysis}, particle \(j\) moves from \(x_{N_\tau-k+1}^{(j)}\) to \(x_{N_\tau-k}^{(j)}\). We denote these two states by \(X_{k-1}^j\) and \(X_k^j\), respectively, so \(X_K^j\) is the numerical endpoint. As before, \(Z\) denotes the system state before noising. Write \(\varrho_k=1-\theta_k\) and \(\varrho_0=0\). At the first score evaluation, the shared proposals keep their initial locations and weights, so \(G_1=1\) and \(Q_1=I\). Each reverse particle still computes its own conditional weights using \eqref{eq:main_ecsf_estimator}.

We condition on the forecast ensemble, observation, and initial proposal
ensemble. Let $p=\widetilde\pi_n^f$ be the fitted forecast density, $g$ the likelihood, and
$\pi_k=pg^{\varrho_k}/Z_k$, where $Z_k=\int pg^{\varrho_k}$. Let $Q_k$
be the transition kernel for one proposal member at stage $k$, including the identity when
movement is omitted. Its parameters are fixed within each reverse run.
Across proposal counts, $Q_k$ abbreviates the corresponding deterministic
kernel $Q_k^M$; each reference uses that run's forecast fit and initial proposals.
Define
\[
G_k=g^{\varrho_k-\varrho_{k-1}},\qquad T_kf=G_kQ_kf,\qquad
\Gamma_k=\eta_0T_1\cdots T_k,\qquad
\eta_k=\Gamma_k/\Gamma_k(1).
\]
The initial measure $\eta_0$ is the empirical proposal measure. Thus
$\eta_k$ is the numerical reference generated by the initial proposal ensemble,
incremental weighting, and proposal transitions.

\paragraph{Assumptions.}
The number of stages and the dimension are fixed. Conditional on the inputs,
$G_k$ and $Q_k$ are deterministic, $0<G_k\le b_k<\infty$, and all displayed
integrals exist. At stage $k$, the current queries and incoming proposal ensemble determine
the ancestor probabilities before new sampling noise is drawn. Selection
covers every positive target weight and uses the correction $v=\omega/s$.
Stratified offsets and descendant transition noises are conditionally independent.
Ratios and variance sums are taken on the positive-weight support, with
zero-weight terms contributing zero.

\paragraph{Proposition C.1 (corrected adaptive sampling).}
Let $\nu_{k-1}$ be the normalized incoming proposal measure,
$c_k=\nu_{k-1}(G_k)$, and
\[
Y_k(f)=\frac1M\sum_{\ell=1}^M
\frac{\omega_{k,A_\ell}}{s_{k,A_\ell}}f(Z_\ell),\quad
R_k=Y_k(1),\quad m_k=m_{k-1}c_kR_k,\quad
\widehat\Gamma_k=m_k\nu_k,\quad m_0=1,
\]
where $Z_\ell$ is the moved descendant and $\nu_k=Y_k/Y_k(1)$.
Without resampling, let $Z'_m\sim Q_k(z_m,\cdot)$ independently conditional
on the incoming ensemble and define
\[
Y_k(f)=\sum_m\omega_{k,m}f(Z'_m),\qquad R_k=Y_k(1)=1.
\]
The identity kernel covers stages with no movement. The same definition also
covers a deterministic one-member transition. If
$\mathcal F_k^-$ contains the incoming proposal ensemble, queries, and past randomness,
then
\[
\mathbb E[\widehat\Gamma_k(f)\mid\mathcal F_k^-]
=\widehat\Gamma_{k-1}(T_kf),\qquad
\mathbb E\widehat\Gamma_k(f)=\Gamma_k(f).
\]
With resampling, the first equality follows by grouping descendants by
ancestor and using the expected count $Ms_{k,m}$, which cancels the selection
probability in $\omega_{k,m}/s_{k,m}$. Without resampling,
$\mathbb E[Y_k(f)\mid\mathcal F_k^-]=\sum_m\omega_{k,m}Q_kf(z_m)$
gives the same equality. These statements concern unnormalized weighted sums
and allow dependence between the queries and incoming proposals.

For stratified resampling define
$\Lambda_k=\sum_m\omega_{k,m}^2/s_{k,m}$. The query-dependent selection rule
admits a direct bound on this factor. Let
\[
\zeta_{k,j}=\sum_m\omega_{k,m}K_k(X_{k-1}^j,z_m),\qquad
H_{k,m}=\frac1J\sum_{j=1}^J
\frac{K_k(X_{k-1}^j,z_m)}{\zeta_{k,j}}.
\]
Then $s_{k,m}=\omega_{k,m}H_{k,m}$ and
\[
\Lambda_k=\sum_m\frac{\omega_{k,m}}{H_{k,m}}
=1+\chi^2(\omega_k\Vert s_k).
\]
If $\|z_m\|\le B$, the Gaussian noising kernel satisfies
\[
\frac{\max_mK_k(X_{k-1}^j,z_m)}{\min_mK_k(X_{k-1}^j,z_m)}
\le e^{D_{k,j}},\qquad
D_{k,j}=\frac{2|a_{\theta_k}|B\|X_{k-1}^j\|
+a_{\theta_k}^2B^2/2}{\sigma_{\theta_k}^2}.
\]
Consequently,
\[
H_{k,m}\ge\frac1J\sum_{j=1}^Je^{-D_{k,j}},\qquad
\Lambda_k\le
\left(\frac1J\sum_{j=1}^Je^{-D_{k,j}}\right)^{-1}
\le \exp\!\left(\frac1J\sum_{j=1}^JD_{k,j}\right).
\]
This connects query--proposal mismatch to the correction variance in the
resampling branch. In particular,
$\mathbb E[m_{k-1}^2\exp(J^{-1}\sum_jD_{k,j})]<\infty$
controls its weighted local-variance factor.

\paragraph{Local variance with and without resampling.}
For a bounded test function $f$, independent stratified offsets and independent
transition noises give
\[
\operatorname{Var}(Y_k(f)\mid\mathcal F_k^-)
\le\frac1M\sum_m\frac{\omega_{k,m}^2}{s_{k,m}}Q_k(f^2)(z_m)
\le\frac{\Lambda_k}{M}\|f\|_\infty^2
\]
when resampling is used, by summing second moments across strata.
Without resampling, conditional independence of the one-member moves gives
\begin{equation}
\begin{aligned}
\operatorname{Var}(Y_k(f)\mid\mathcal F_k^-)
&=\sum_m\omega_{k,m}^2
 \left[Q_k(f^2)(z_m)-(Q_kf(z_m))^2\right]\\
&\le\|f\|_\infty^2\sum_m\omega_{k,m}^2.
\end{aligned}
\label{eq:adaptive_no_resampling_variance}
\end{equation}
This variance is zero for a deterministic transition. Define
\begin{equation}
\Lambda_k^{\mathrm{loc}}=
\begin{cases}
\displaystyle\sum_m\omega_{k,m}^2/s_{k,m},
    &\text{with resampling},\\[2pt]
\displaystyle M\sum_m\omega_{k,m}^2,
    &\text{without resampling, with a random move},\\[2pt]
0,  &\text{without resampling, with a deterministic transition}.
\end{cases}
\label{eq:adaptive_branch_variance_factor}
\end{equation}
Then all branches obey
$\operatorname{Var}(Y_k(f)\mid\mathcal F_k^-)
\le\Lambda_k^{\mathrm{loc}}\|f\|_\infty^2/M$.
The following quantitative bounds assume
$C_k:=\mathbb E[m_{k-1}^2\Lambda_k^{\mathrm{loc}}]<\infty$.
When proposals move randomly without resampling, the relevant factor is
$M\sum_m\omega_{k,m}^2=M/\operatorname{ESS}(\omega_k)$,
where $\operatorname{ESS}(\omega_k)=(\sum_m\omega_{k,m}^2)^{-1}$.

Set
\[
V_k=\frac1M\sum_{r=1}^k C_r\prod_{\ell=r}^kb_\ell^2,\qquad
E_k=32V_k/\Gamma_k(1)^2.
\]
For every fixed bounded $f$,
\[
\mathbb E|\widehat\Gamma_k(f)-\Gamma_k(f)|^2
\le V_k\|f\|_\infty^2,
\qquad
\mathbb E|\nu_k(f)-\eta_k(f)|^2
\le\min(4,E_k)\|f\|_\infty^2.
\]
To see the first bound, apply the local-variance inequality to the deterministic
backward functions $T_{r+1}\cdots T_kf$. Their suprema are at most
$\|f\|_\infty\prod_{\ell=r+1}^kb_\ell$, while
$c_r=\nu_{r-1}(G_r)\le b_r$. The martingale increment
$\widehat\Gamma_r(T_{r+1}\cdots T_kf)
 -\widehat\Gamma_{r-1}(T_r\cdots T_kf)$
therefore has second moment at most
$C_r\|f\|_\infty^2\prod_{\ell=r}^kb_\ell^2/M$.
Summing the orthogonal increments proves the bound.
For the normalized bound, on $m_k\ge\Gamma_k(1)/2$ use
\[
\nu_k(f)-\eta_k(f)=
\frac{\widehat\Gamma_k(f)-\Gamma_k(f)
      -\eta_k(f)[m_k-\Gamma_k(1)]}{m_k}.
\]
The two unnormalized error bounds contribute at most
$16V_k\|f\|_\infty^2/\Gamma_k(1)^2$ on this event.
On its complement, use $|\nu_k(f)-\eta_k(f)|\le2\|f\|_\infty$
and the mass-error bound to obtain the same contribution. The trivial
$4\|f\|_\infty^2$ bound yields the displayed minimum.

Retaining the test function in the local variance gives a more specific bound.
For $h_r=T_{r+1}\cdots T_kf$, define
\[
B_r(h)=\begin{cases}
M^{-1}\sum_m\omega_{r,m}^2Q_r(h^2)(z_m)/s_{r,m},&\text{with resampling},\\
\sum_m\omega_{r,m}^2[Q_r(h^2)(z_m)-(Q_rh(z_m))^2],&\text{without resampling}.
\end{cases}
\]
The same martingale argument yields
\[
\mathbb E|\widehat\Gamma_k(f)-\Gamma_k(f)|^2
\le\sum_{r=1}^k\mathbb E[(m_{r-1}c_r)^2 B_r(h_r)].
\]

\paragraph{Proposition C.1a (full-chain consistency).}
Suppose the initial proposal measures have uniformly bounded $r$th moments
for some $r>0$, and their largest initial weights tend to zero. Let each
$G_k$ be continuous and let the one-member kernels satisfy
\[
Q_k(\|\cdot\|^r)(z)\le A_k(1+\|z\|^r)
\]
with $A_k$ uniform in $M$. Assume average tightness of the queries:
\[
q_{k,R}:=\sup_M\frac1{J_M}\sum_{j=1}^{J_M}
\Pr(\|X_{k-1}^{j,M}\|>R)\longrightarrow0
\quad\text{as }R\to\infty.
\]
For the Gaussian affinity with fixed $a_{\theta_k}\ne0$ and
$\sigma_{\theta_k}>0$, the corrected proposal recursion satisfies
\[
\sup_{\|f\|_\infty\le1}
\mathbb E|\nu_k(f)-\eta_k^M(f)|^2\longrightarrow0
\qquad(k\le K).
\]
The supremum is over deterministic bounded measurable functions and is
outside the expectation. This includes $J_M=M$ and queries dependent on
the proposal ensemble.

\emph{Proof.}
The reference moment bound follows by induction. An incoming moment bound
places at least half the reference mass in a fixed ball, where $G_k$ has a
positive minimum. Hence $\eta_{k-1}^M(G_k)$ is uniformly positive and
\[
\eta_k^M(\|z\|^r)
\le\frac{b_k A_k[1+\eta_{k-1}^M(\|z\|^r)]}
{\eta_{k-1}^M(G_k)}.
\]
Thus $\gamma_k^M:=\Gamma_k^M(1)$ has a uniform positive lower bound,
$\gamma_k^M\le\bar\gamma_k:=\prod_{r=1}^kb_r$, and
$t_{k,B}:=\sup_M\Gamma_k^M(1_{\{\|z\|>B\}})\to0$ as $B\to\infty$.
Proposition C.1 gives $\mathbb E m_k=\gamma_k^M$, so
$\Pr(m_k>L)\le\bar\gamma_k/L$.

Let $E_{k,R}$ be the event that at least half the queries lie in the ball
of radius $R$. Its complement has probability at most $2q_{k,R}$.
Cancel the common Gaussian prefactor in $H_{k,m}$, so $\zeta_{k,j}\le1$; on this event every parent with $\|z_m\|\le B$
satisfies
\[
H_{k,m}\ge c_{k,R,B}:=\tfrac12
\exp\!\left[-\frac{(R+|a_{\theta_k}|B)^2}{2\sigma_{\theta_k}^2}\right]>0.
\]
Write $u_{k,m}$ for the unnormalized member weights, so
$\widehat\Gamma_k=\sum_m u_{k,m}\delta_{z_{k,m}}$.
On $E_{k,R}\cap\{m_{k-1}\le L\}$, descendants of these parents have
$u_{k,m}\le Lb_k/(Mc_{k,R,B})$. The expected total weight of descendants
whose parents lie outside the ball is at most $b_kt_{k-1,B}$, by the
conditional expectation identity in Proposition C.1. Markov's inequality
therefore gives $\max_m u_{k,m}\to0$ in probability: fix $L,R,B$, let
$M\to\infty$, then enlarge the three bounds. At a stage without
resampling, $\max_m u_{k,m}\le b_k\max_m u_{k-1,m}$, which completes
the induction from the initial weights.

For deterministic $\|h\|_\infty\le1$, define the innovation
$\mathcal E_k(h)=\widehat\Gamma_k(h)-\widehat\Gamma_{k-1}(T_kh)$.
At a resampling stage, split it according to whether the parent lies
inside or outside the ball of radius $B$. On the same event as above,
the conditional variance of the inside contribution is at most
$L^2b_k^2/(Mc_{k,R,B})$. The outside contribution has expected absolute
value at most $2b_kt_{k-1,B}$. Conditional Chebyshev and Markov inequalities give
\[
\begin{aligned}
\sup_{\|h\|_\infty\le1}\Pr(|\mathcal E_k(h)|>\epsilon)
\le{}&\bar\gamma_{k-1}/L+2q_{k,R}\\
&+\frac{4L^2b_k^2}{Mc_{k,R,B}\epsilon^2}
+\frac{4b_kt_{k-1,B}}\epsilon.
\end{aligned}
\]
The same order of limits makes this probability vanish. Without
resampling, the conditional variance is bounded by
$b_k^2(\max_m u_{k-1,m})m_{k-1}$, which tends to zero in probability.
Taking expectations of the conditional probability bound truncated at
one gives the same conclusion. Deterministic transitions have zero
innovation at these stages.

Finally, telescope with the deterministic backward functions
$h_r=T_{r+1}\cdots T_kf$:
\[
\widehat\Gamma_k(f)-\Gamma_k^M(f)
=\sum_{r=1}^k\mathcal E_r(h_r),\qquad
\|h_r\|_\infty\le\|f\|_\infty\prod_{\ell=r+1}^kb_\ell.
\]
The uniform innovation bounds give convergence in probability uniformly
over $\|f\|_\infty\le1$. Taking $f=1$ controls $m_k-\gamma_k^M$.
The positive reference mass bound then permits normalization.
Since $|\nu_k(f)-\eta_k^M(f)|\le2$, this uniform probability convergence
also gives the stated mean-square convergence.

\paragraph{Moment growth for ordinary Langevin moves.}
Suppose the drift satisfies $\|b_k^M(z)\|\le L_k\|z\|+B_k$, with constants
uniform in $M$. For $Z'=z+(h/2)b_k^M(z)+\sqrt h\,\xi$, where
$\xi\sim\mathcal N(0,I_d)$,
\[
\mathbb E[\|Z'\|^2\mid z]
\le 2(1+hL_k/2)^2\|z\|^2+h^2B_k^2/2+hd.
\]
Iteration over a fixed number of moves establishes the moment-growth
condition in Proposition C.1a with $r=2$. For the diagonal Gaussian
forecast, a uniform variance lower bound and bounded forecast means give
this linear-growth bound for the prior gradient. With fixed observation
$y$, the arctangent observation map and its derivative are bounded, so
the likelihood gradient is bounded. A linear Gaussian observation gives
an affine likelihood gradient and also satisfies the same growth condition.

\paragraph{Proposition C.2 (adaptive conditional means).}
For this proposition, suppose proposal measures and numerical references
are supported in a common ball of radius $B$, uniformly in $M$.
The residual likelihood is continuous and positive there, the noising variance
is positive, and the adaptive queries $X=X_M$ are uniformly tight.
Cancel the noising factor independent of $z$ and write
\[
F_x(z)=\exp\!\left(
\frac{a_{\theta_k}x^Tz}{\sigma_{\theta_k}^2}
-\frac{a_{\theta_k}^2\|z\|^2}{2\sigma_{\theta_k}^2}
\right)g(z)^{1-\varrho_k},\qquad
D_\mu(x)=\frac{\mu(zF_x)}{\mu(F_x)}.
\]
On $\|x\|\le R$, let $U_R=\sup F_x(z)$ and
$l_R=\inf F_x(z)>0$ over the proposal ball. The weighted-moment ratio
identity and Proposition C.1 give the fixed-query squared-error bound
$A_R=4dB^2(U_R/l_R)^2E_k$. Moreover,
$\nabla_xD_\mu=(a_{\theta_k}/\sigma_{\theta_k}^2)
\operatorname{Cov}_{\mu,x}(Z)$, so the difference between two such conditional
means is $2|a_{\theta_k}|B^2/\sigma_{\theta_k}^2$-Lipschitz. An
$\epsilon$-net argument therefore gives, for an adaptive query $X$,
\[
\mathbb E\|D_{\nu_k}(X)-D_{\eta_k}(X)\|^2
\le\min\!\left\{4B^2,
2(1+2R/\epsilon)^dA_R
+\frac{8a_{\theta_k}^2B^4\epsilon^2}{\sigma_{\theta_k}^4}
+4B^2\Pr(\|X\|>R)\right\}.
\]
The bound allows dependence between $X$ and $\nu_k$. If the
$C_k$ are uniformly bounded, $\Gamma_k(1)$ is uniformly positive, and the
query tightness condition holds, taking $M\to\infty$, then $\epsilon\to0$ and
$R\to\infty$, yields mean-square convergence of the conditional means.

\paragraph{Proposition C.2a (conditional means on an unbounded state space).}
Under Proposition C.1a, suppose $\ell(z)=g(z)^{1-\varrho_k}$ is positive,
continuous and bounded by $L_\ell$. Then, for an independent uniform member
index $I_M$,
\[
D_{\nu_k}(X_{k-1}^{I_M})-D_{\eta_k^M}(X_{k-1}^{I_M})
\longrightarrow0\quad\text{in probability}.
\]
If the squared errors are uniformly integrable, convergence also holds
in mean square.

\emph{Proof.}
Write $a=a_{\theta_k}$, $\sigma=\sigma_{\theta_k}$ and retain the full
Gaussian factor,
$\widetilde F_x(z)=\exp[-\|x-az\|^2/(2\sigma^2)]\ell(z)$.
It gives the same ratio $D_\mu=\mu(z\widetilde F_x)/\mu(\widetilde F_x)$.
For $\|x\|\le R$, uniformly over $z\in\R^d$,
\[
\begin{aligned}
|\widetilde F_x|&\le L_\ell,&
\|z\widetilde F_x\|&\le\frac{L_\ell}{|a|}(R+\sigma/\sqrt e),\\
\|\nabla_x\widetilde F_x\|&\le\frac{L_\ell}{\sigma\sqrt e},&
\|\nabla_x(z\widetilde F_x)\|_{\mathrm{op}}
&\le\frac{L_\ell}{|a|}\left(\frac R{\sigma\sqrt e}+\frac2e\right).
\end{aligned}
\]
These bounds follow from $\|z\|\le(R+\|x-az\|)/|a|$ and the first two
polynomial Gaussian maxima. Reference tightness places a fixed positive
mass in a ball where $\ell$ has a positive minimum, giving
$\inf_M\inf_{\|x\|\le R}\eta_k^M(\widetilde F_x)>0$.
Apply Proposition C.1a to the numerator coordinates and denominator on
a finite query net. The derivative bounds extend convergence to the
whole query ball, and the positive denominator bound gives uniform
convergence in probability of the ratio. Average query tightness handles
the probability outside that ball. Uniform integrability gives the
additional mean-square conclusion.

\paragraph{Proposition C.3 (standardized terminal initialization).}
For one coordinate, let $G_1,\ldots,G_J$ be independent standard Gaussians,
$\bar G=J^{-1}\sum_jG_j$,
$S^2=(J-1)^{-1}\sum_j(G_j-\bar G)^2$, and
$\widetilde S=\max(S,\varepsilon)$. The recursive implementation initializes
$X_0^j=(G_j-\bar G)/\widetilde S$ componentwise. If $I$ is uniform on
$\{1,\ldots,J\}$ and independent of the ensemble, then
\[
\mathbb E(X_0^I-G_I)^2
=\kappa_{J,\varepsilon}
:=\frac1J+\frac{J-1}{J}
\mathbb E\!\left[S^2(\widetilde S^{-1}-1)^2\right].
\]
For $\varepsilon=0$ this becomes
\[
\kappa_{J,0}=\frac1J+\frac{2(J-1)}J(1-c_J),\qquad
c_J=\sqrt{\frac2{J-1}}\,
\frac{\Gamma(J/2)}{\Gamma((J-1)/2)}.
\]
Across $d$ independent coordinates,
$\mathbb E\|X_0^I-G_I\|^2=d\kappa_{J,\varepsilon}$. Since the exact terminal
noised posterior has the form $\epsilon_aZ+G$ for the reported schedule,
where $Z$ has the analysis distribution and $G\sim\mathcal N(0,I_d)$ is independent of $Z$,
\[
W_2(\mathcal L(X_0^I),\mathcal L(\epsilon_aZ+G))
\le\sqrt{d\kappa_{J,\varepsilon}}
+\epsilon_a(\mathbb E\|Z\|^2)^{1/2}.
\]
This is a member-marginal bound; standardization makes the $J$ initialized
members dependent.

\paragraph{Theorem C.4 (reverse endpoint control).}
Let $s_k^R$ be the score obtained from the numerical proposal reference and
let
$\Psi_k(x)=(1-\delta_k f_{\theta_k})x+\alpha_ks_k^R(x)$, where
$\alpha_k=\delta_kr_{\theta_k}^2$. Assume $\Psi_k$ is
$\beta_k$-Lipschitz. Couple the actual reverse particles $X_k^j$ and reference
particles $Y_k^j$ with the same Euler noise, and define
\[
q_k^2=\mathbb E\frac1J\sum_j
\|\widehat s_k^j(X_{k-1}^j)-s_k^R(X_{k-1}^j)\|^2,\qquad
w_k=\alpha_k\prod_{\ell=k+1}^K\beta_\ell.
\]
For a uniform member index \(I\) independent of the ensembles, write
\(e_k=(\mathbb E\|X_k^I-Y_k^I\|^2)^{1/2}\).
The shared Euler noise cancels in each pair. Adding and subtracting the
reference drift map at the actual particle and using its Lipschitz bound gives
\(e_k\leq\beta_k e_{k-1}+\alpha_k q_k\).
Iterating this inequality and applying the triangle inequality gives, for any
target \(q_\star\) with a finite second moment,
\[
W_2(\mathcal L(X_K^I),q_\star)
\le
\left(\prod_{k=1}^K\beta_k\right)
\left(\mathbb E\|X_0^I-Y_0^I\|^2\right)^{1/2}
+\sum_{k=1}^Kw_kq_k
+W_2(\mathcal L(Y_K^I),q_\star).
\]
The first term can be bounded by Proposition C.3 in the recursive setting; it
is zero in the controlled Gaussian experiment, which samples the analytic
terminal reference. Under Proposition C.2, the conditional Tweedie identity gives
$q_k\to0$ for raw scores and for scores transformed by the same nonexpansive
map in both chains.

\paragraph{Corollary C.4a (numerical endpoint consistency).}
Use the proposal system of Proposition C.2a. Let $P_k$ be a common
nonexpansive score map with $\sup_v\|P_k(v)\|\le C_k^S$, and define
\[
\widehat S_k(x)=P_k\!\left[-\frac{x}{\sigma_{\theta_k}^2}
+\frac{a_{\theta_k}}{\sigma_{\theta_k}^2}D_{\nu_k}(x)\right],\qquad
S_k^R(x)=P_k\!\left[-\frac{x}{\sigma_{\theta_k}^2}
+\frac{a_{\theta_k}}{\sigma_{\theta_k}^2}D_{\eta_k^M}(x)\right].
\]
Couple the Euler updates using the same noise for each paired member.
Assume uniformly bounded initial average second moments and
\[
\sup_M\frac1{J_M}\sum_j\mathbb E\|\xi_k^{j,M}\|^2<\infty
\qquad(k\le K).
\]
If $\mathbb E\|X_0^{I_M}-Y_0^{I_M}\|^2\to0$, then
\[
\mathbb E\|X_K^{I_M}-Y_K^{I_M}\|^2\to0,\qquad
W_2(\mathcal L(X_K^{I_M}),\mathcal L(Y_K^{I_M}))\to0.
\]
Here the independent uniform index $I_M$ is kept along the entire trajectory.

\emph{Proof.}
Set $A_k^E=1-\delta_k f_{\theta_k}$ and
$\alpha_k=\delta_k r_{\theta_k}^2$. The bounded scores and uniform noise
moments give a uniform average second-moment bound at each stage, using
\[
\mathbb E\|X_k^{I_M}\|^2
\le3|A_k^E|^2\mathbb E\|X_{k-1}^{I_M}\|^2
+3\alpha_k^2(C_k^S)^2+3\alpha_k\mathbb E\|\xi_k^{I_M}\|^2,
\]
and the same inequality for $Y$. This establishes average query tightness
directly from the reverse recursion. Proposition C.2a then gives
$\widehat S_k(X_{k-1}^{I_M})-S_k^R(X_{k-1}^{I_M})\to0$ in probability
and in mean square, since its norm is bounded by $2C_k^S$.
The Gaussian derivative and denominator bounds in its proof make
$S_k^R$ uniformly locally Lipschitz. Tightness and a previous-stage
coupling error tending to zero in probability therefore imply
$S_k^R(X_{k-1}^{I_M})-S_k^R(Y_{k-1}^{I_M})\to0$ in probability.
The full score difference is bounded by $2C_k^S$, so it converges in mean
square. The common noise cancels, leaving
\[
\|X_k^{I_M}-Y_k^{I_M}\|_{L^2}
\le |A_k^E|\|X_{k-1}^{I_M}-Y_{k-1}^{I_M}\|_{L^2}
+|\alpha_k|\|\widehat S_k(X_{k-1}^{I_M})-S_k^R(Y_{k-1}^{I_M})\|_{L^2}.
\]
Finite-stage induction proves the endpoint result; this coupling also
bounds $W_2$. A common nonexpansive output map preserves the bound.

\paragraph{Proposition C.4b (endpoint consistency with raw scores).}
Start with the corrected proposal recursion in Proposition C.1a. Retain its assumptions on the initial measure, potentials, and individual move kernels, together with the residual-likelihood conditions of Proposition C.2a. Assume uniform average tightness of the initial reverse particles and the Euler noise at each stage. Couple the actual and reference particles using the same Euler noise, and use the raw conditional score in each chain. Choose a uniformly random member index \(I_M\) independently and follow that member throughout the run. If \(X_0^{I_M}-Y_0^{I_M}\to0\) in probability, then
\[
X_k^{I_M}-Y_k^{I_M}\longrightarrow0\quad\text{in probability}
\qquad(k\le K).
\]
The reverse queries remain average-tight at every stage, and the endpoint
laws approach each other in bounded-Lipschitz distance.

\emph{Proof.}
The reference moment recursion in Proposition C.1a depends only on the
initial proposals, potentials and move kernels. It gives tight reference
measures before considering the reverse queries. The Gaussian envelopes
and positive denominator bound in Proposition C.2a then make the reference
scores $s_k^{R,M}$ uniformly bounded and Lipschitz on every fixed query
ball, with constants uniform in $M$.

Proceed jointly through the proposal and reverse stages. Tight initial
queries and the initial coupling give tight reference queries. Suppose
both chains are tight through stage $k-1$ and their coupled difference
tends to zero in probability. The localization proof of Proposition C.1a
applies to the proposal prefix through stage $k$, whose input queries
are all tight by induction. Proposition C.2a therefore gives
\[
e_k^M:=\widehat s_k(X_{k-1}^{I_M})
       -s_k^{R,M}(X_{k-1}^{I_M})\longrightarrow0\quad\text{in probability}.
\]
Local boundedness of the reference score and tightness of the query imply
tightness of this reference score evaluation. Adding $e_k^M$ gives
tightness of the actual score evaluation. The Euler update and tight
noise thus give tight queries at stage $k$ in both chains.

Again let $A_k^E=1-\delta_k f_{\theta_k}$. On the event that both previous queries
lie in a ball of radius $R$, the common noise cancels and
\[
\begin{aligned}
\|X_k^{I_M}-Y_k^{I_M}\|
\le{}& (|A_k^E|+|\alpha_k|L_{k,R})
          \|X_{k-1}^{I_M}-Y_{k-1}^{I_M}\|\\
&+|\alpha_k|\|e_k^M\|,
\end{aligned}
\]
where $L_{k,R}$ is a uniform local Lipschitz constant of $s_k^{R,M}$.
The right side tends to zero in probability for fixed $R$. The probability
outside this event tends uniformly to zero as $R\to\infty$, completing
the finite-stage induction. For any function with sup norm and Lipschitz
constant at most one,
\[
|\mathbb E f(X_K^{I_M})-\mathbb E f(Y_K^{I_M})|
\le\mathbb E\min\{2,\|X_K^{I_M}-Y_K^{I_M}\|\}\to0.
\]
This proves the endpoint law comparison.

The main training-free recursive benchmarks use $M=J$, with one
proposal initialized at each forecast member. Along $M=J\to\infty$, Proposition C.3 controls the
terminal-member marginal and Propositions C.1a--C.2a control adaptive proposal integration.

\paragraph{Reference bias relative to the target.}
Let $f_{r:k}=T_{r+1}\cdots T_kf$. The numerical proposal reference obeys the
exact identity
\[
\Gamma_k(f)-Z_k\pi_k(f)
=(\eta_0-p)(T_1\cdots T_kf)
+\sum_{r=1}^kZ_r\pi_r[(Q_r-I)f_{r:k}].
\]
It follows by iterating
$\Gamma_k-\gamma_k^*=(\Gamma_{k-1}-\gamma_{k-1}^*)T_k
+\gamma_k^*(Q_k-I)$, where $\gamma_k^*=Z_k\pi_k$. Put
$D_k^*(x)=\pi_k(zF_x)/\pi_k(F_x)$ and
$\varphi_x=(z-D_k^*(x))F_x$. Then
\[
D_{\eta_k}(x)-D_k^*(x)=
\frac{(\eta_0-p)T_1\cdots T_k\varphi_x
+\sum_{r=1}^kZ_r\pi_r[(Q_r-I)T_{r+1}\cdots T_k\varphi_x]}
{\Gamma_k(F_x)}.
\]
If the displayed backward test functions are Lipschitz, define
$L_0=\operatorname{Lip}(T_1\cdots T_k\varphi_x)$,
$L_r=\operatorname{Lip}(T_{r+1}\cdots T_k\varphi_x)$, and
$\Delta_r=W_1(\pi_rQ_r,\pi_r)$. The preceding identity gives
\[
\|D_{\eta_k}(x)-D_k^*(x)\|
\le\frac{L_0W_1(\eta_0,p)+\sum_{r=1}^kZ_rL_r\Delta_r}
{\Gamma_k(F_x)}.
\]
Thus the target discrepancy separates the initial forecast representation from
the invariance defect of each proposal transition.

For an unprocessed ULA step, write
$\widetilde P_h(x,\cdot)=\mathcal L(x+h\widetilde b(x)+\sqrt h\,\xi)$,
where $\widetilde b$ is its drift approximation, and let $P_h$ use the ideal
drift $b=\tfrac12\nabla\log\pi$. A synchronous coupling gives
\[
W_1(\pi\widetilde P_h,\pi)
\le h\,\mathbb E_\pi\|\widetilde b-b\|+W_1(\pi P_h,\pi).
\]
If $b$ is globally $L_b$-Lipschitz, $\pi$ is stationary for the continuous
Langevin diffusion, and
$B_b=\mathbb E_\pi\|b\|<\infty$, then
\[
W_1(\pi P_h,\pi)
\le L_b\left(\tfrac12B_bh^2+\tfrac23\sqrt d\,h^{3/2}\right).
\]
The first term measures drift approximation and the second ULA discretization.
For several steps, the corresponding one-step defects propagate through the
Lipschitz constants of the intervening kernels. These target-bias terms, the
reverse-time discretization, and the retained endpoint noise do not generally
vanish when only the proposal count grows.

\subsection{Computing the proposal reference exactly in the Gaussian case}
\label{app:gaussian_reference_recursion}
For a Gaussian fitted forecast and linear Gaussian observations, the
numerical proposal reference has a finite Gaussian-mixture representation.
This gives a computable example of the preceding reference-bias decomposition.
Here we use the ordinary linear ULA transition; this example is separate
from the compact-support assumptions of Proposition C.2.

Let $p=\mathcal N(m_f,C_f)$, $C_f\succ0$, and let
$y=Lz+\varepsilon$, where $\varepsilon\sim\mathcal N(0,R)$ and
$R\succ0$. Using the execution index $k$ defined above, start from
$\eta_0=\sum_{m=1}^M u_{0,m}\delta_{z_m}$ and write
\[
\eta_{k-1}=\sum_{m=1}^M
u_{k-1,m}\mathcal N(b_{k-1,m},C_{k-1,m}),
\]
where zero covariance denotes a point mass. For
$\Delta\varrho_k=\varrho_k-\varrho_{k-1}>0$, set
$V_k^{\rm obs}=R/\Delta\varrho_k$ and define
\[
S_{k,m}=LC_{k-1,m}L^T+V_k^{\rm obs},\qquad
K_{k,m}=C_{k-1,m}L^TS_{k,m}^{-1}.
\]
Incremental likelihood weighting gives
\[
\bar b_{k,m}=b_{k-1,m}+K_{k,m}(y-Lb_{k-1,m}),\qquad
\bar C_{k,m}=C_{k-1,m}-K_{k,m}LC_{k-1,m},
\]
\[
\bar u_{k,m}\propto u_{k-1,m}
\phi_{S_{k,m}}(y-Lb_{k-1,m}).
\]
The omitted likelihood normalization factor is common to all components
and cancels when their masses are normalized. When
$\Delta\varrho_k=0$, weighting is the identity.

For a ULA step of size $h>0$, define
\[
\mathcal H_k=C_f^{-1}+\varrho_kL^TR^{-1}L,\qquad
v_k=C_f^{-1}m_f+\varrho_kL^TR^{-1}y,\qquad
A_k=I-\tfrac h2\mathcal H_k.
\]
Each step maps a component's moments as
\[
b\longmapsto A_kb+\tfrac h2v_k,\qquad
C\longmapsto A_kCA_k^T+hI,
\]
with unchanged component mass. Applying this map for the scheduled number
of Langevin steps computes $\eta_k$ with $M$ components at every stage;
if movement is omitted, the weighted mixture is retained. At the initial
score evaluation, both weighting and movement are identities.
These formulas follow by completing the square for likelihood weighting
and applying an affine Gaussian transition.

For the tempered target
$\pi_k=\mathcal N(\mathcal H_k^{-1}v_k,\mathcal H_k^{-1})$,
one ULA step preserves its mean and increases its covariance by
$h^2\mathcal H_k/4$. After $\ell$ steps from $\pi_k$, the covariance
excess is
\[
\frac{h^2}{4}\sum_{r=0}^{\ell-1}
A_k^r\mathcal H_k(A_k^r)^T.
\]
If $0<h\lambda_{\max}(\mathcal H_k)<4$, the stationary covariance is
$[\mathcal H_k(I-h\mathcal H_k/4)]^{-1}$.
These identities quantify the transition defect at a fixed step size.

To compute $D_{\eta_k}(x)$, update each component first with the
remaining likelihood $g^{1-\varrho_k}$ and then with the Gaussian
observation $x=a_{\theta_k}z+\sigma_{\theta_k}\xi$.
The first update uses observation covariance $R/(1-\varrho_k)$
when $\varrho_k<1$ and is omitted when $\varrho_k=1$.
The second uses observation matrix $a_{\theta_k}I$ and covariance
$\sigma_{\theta_k}^2I$. Normalize the component masses after multiplying
by both marginal likelihoods and average their conditional means.
The corresponding reference score is
\[
s_k^R(x)=\frac{a_{\theta_k}D_{\eta_k}(x)-x}
{\sigma_{\theta_k}^2}.
\]
Thus the proposal reference and its conditional score can be evaluated
without sampling additional proposals.

\section{Experimental protocols and supplementary analyses}
\label{app:experimental_protocols}

\subsection{Controlled Gaussian evaluation}
\label{app:controlled_highdim_details}

We draw the system state from \(Z\sim\N(0,\I_{128})\) and observe \(Y=Z+\varepsilon\), where \(\varepsilon\sim\N(0,0.25\I_{128})\). For each of ten observation problems (seeds 10--19), we draw an independent 20-member forecast ensemble and fit the diagonal Gaussian \(\widetilde\pi^f=\N(m_f,\operatorname{diag}(v_f))\). We regularize the variances with jitter \(10^{-5}\) and a floor of \(10^{-3}\). Within each problem, all methods and all four repetitions share the observation, forecast ensemble, fitted prior, and reference samples. Only the sampler's random streams vary across repetitions. We therefore summarize results across the ten observation problems after averaging repetitions within each one.

The fitted posterior has componentwise variance \(v_a=(v_f^{-1}+4)^{-1}\) and mean \(m_a=v_a(m_f/v_f+4y)\). The data-generating posterior has mean \(0.8y\) and variance \(0.2\I\). Each endpoint reference includes the retained noising variance \(\epsilon_b=0.025\). We draw 4096 independent samples from each reference per problem. All methods start reverse diffusion from the same fitted noisy-posterior distribution at \(\tau=1\). This common analytic initialization isolates the subsequent sampling procedure; the Lorenz--96 and original Kuramoto--Sivashinsky recursive benchmarks instead use the standardized Gaussian initialization in Appendix~\ref{subsec:reverse_recursive}. The diffusion schedule uses \(\epsilon_a=0.001\), 100 steps uniform in logSNR, and the actual nonuniform time increments.

AECSF and AECSF without resampling use the same fitted prior gradient, 294 Langevin steps with \(h=0.02\), and the same separate streams for reverse diffusion and proposal movement. AECSF without resampling retains incremental proposal weights but omits resampling. The other controls use the same fitted Gaussian information. Writing \(b_\tau=\sigma_\tau^2\) and interpreting vector products componentwise, define
\begin{align}
s_\tau^f(x)&=\frac{a_\tau m_f-x}{a_\tau^2v_f+b_\tau},\nonumber\\
D_\tau^f(x)&=m_f+\frac{a_\tau v_f}{a_\tau^2v_f+b_\tau}(x-a_\tau m_f),\nonumber\\
s_\tau^{\rm point}(x;y)&=s_\tau^f(x)+(1-\tau)\nabla\log g(y\given x),\nonumber\\
s_\tau^{\rm denoise}(x;y)&=s_\tau^f(x)+J_{D_\tau^f}(x)^\top\nabla\log g(y\given D_\tau^f(x)).
\label{eq:controlled_guidance_definitions}
\end{align}
Here \(D_\tau^f(x)\) is the conditional mean of the system state given \(X_\tau=x\) under the fitted forecast, without conditioning on the current observation. Its Jacobian \(J_{D_\tau^f}\) is diagonal with entries \(a_\tau v_f/(a_\tau^2v_f+b_\tau)\). The analytic posterior-score control uses the exact score of the noised posterior under the fitted prior \((a_\tau m_a-x)/(a_\tau^2v_a+b_\tau)\), integrated with the same numerical reverse solver and finite output ensemble.

\paragraph{Endpoint energy statistics.}
Let \(X_{rm}\) be output member \(m\) of repetition \(r\), where \(R=4\), \(M=20\), and \(n=RM=80\); write \(X_p\) for the pooled samples. In this subsection, \(M=J\) counts output samples and \(n\) is the pooled sample count, not assimilation time. Let \(Y_k\), \(k=1,\ldots,K\), be the reference samples, with \(K=4096\), and set \(d(u,v)=\|u-v\|/\sqrt{128}\). The two reported discrepancies are
\begin{align}
E_V&=\frac{2}{nK}\sum_{p,k}d(X_p,Y_k)
-\frac{1}{n^2}\sum_{p,q}d(X_p,X_q)
-\frac{1}{K^2}\sum_{k,l}d(Y_k,Y_l),\label{eq:controlled_gaussian_energy}\\
E_{\rm cross}&=\frac{2}{nK}\sum_{p,k}d(X_p,Y_k)
-\frac{\sum_{r\ne s}\sum_{m,l}d(X_{rm},X_{sl})}{R(R-1)M^2}
-\frac{\sum_{k\ne l}d(Y_k,Y_l)}{K(K-1)}.\label{eq:controlled_gaussian_cross_energy}
\end{align}
The V-statistic evaluates the pooled empirical distribution of 80 particles. The cross-repetition statistic pairs particles from different runs and excludes reference self-pairs. Conditional on each problem, independent repetitions and reference samples make it an estimator of energy discrepancy for the marginal output distribution averaged over member indices. Negative estimates are retained. Figures show means and sample standard deviations across ten problems, with individual problem values overlaid.

\begin{figure}[t]
\centering
\includegraphics[trim=0 4 0 0,clip,width=\textwidth]{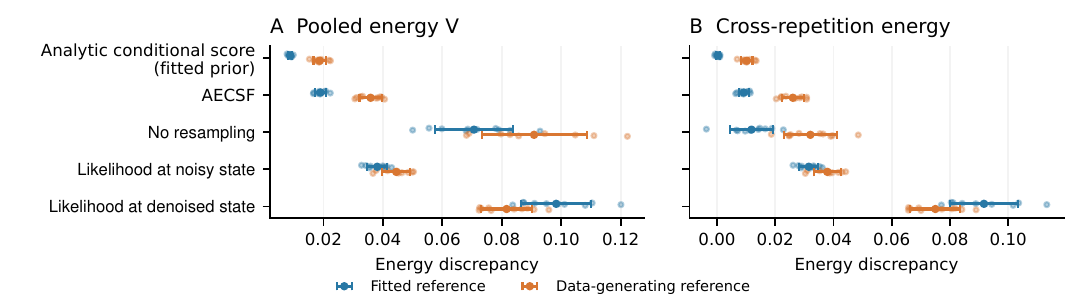}
\caption{Gaussian endpoint energy discrepancies. (A) Pooled-particle statistic;
(B) cross-repetition statistic. Colors distinguish fitted and data-generating
posterior references at the retained noise level. Large points/bars show
means/sample SDs over ten problems; small points show individual problems.}
\label{fig:controlled_score_endpoint_dimension}
\end{figure}

The analytic-score control in Figure~\ref{fig:controlled_score_endpoint_dimension}
uses the exact noised score of the fitted posterior.

\begin{table}[htbp]
\centering\small
\caption{Gaussian endpoint summaries across ten problems. Each problem uses four repetitions of 20 particles. The two energy statistics are defined in Eqs.~\eqref{eq:controlled_gaussian_energy}--\eqref{eq:controlled_gaussian_cross_energy}.}
\label{tab:controlled_gaussian_primary}
\resizebox{\textwidth}{!}{%
\begin{tabular}{lccccc}
\toprule
Method & Fitted $E_V$ & True $E_V$ & Fitted $E_{\rm cross}$ & True $E_{\rm cross}$ & Seconds/run \\
\midrule
Analytic posterior score (fitted prior) & 0.008850 & 0.018799 & 0.000284 & 0.010230 & 0.069935 \\
AECSF & 0.019040 & 0.035886 & 0.009168 & 0.026011 & 0.300461 \\
AECSF without resampling & 0.070636 & 0.090871 & 0.011818 & 0.032051 & 0.216578 \\
Noisy-state likelihood guidance & 0.038129 & 0.044543 & 0.031485 & 0.037897 & 0.062395 \\
Denoised-state likelihood guidance & 0.098303 & 0.081641 & 0.091629 & 0.074965 & 0.066121 \\
\bottomrule
\end{tabular}}
\end{table}

To evaluate individual 20-member outputs, let \(\bar X_r=M^{-1}\sum_m X_{rm}\) and \(\widehat v_{r,q}=(M-1)^{-1}\sum_m(X_{rm,q}-\bar X_{r,q})^2\). For each repetition, we compute
\[
e_{\mu,r}=d^{-1}\|\bar X_r-\mu_\star\|^2,\qquad e_{v,r}=d^{-1}\|\widehat v_r-v_\star\|^2,
\]
using the analytic mean and diagonal variance of the corresponding retained-noise reference. We first average the four repetitions within each problem, then summarize the ten problems. The diagonal-variance error includes finite-ensemble sampling variability.

\begin{table}[htbp]
\centering\small
\caption{Mean and diagonal-variance errors of individual 20-member Gaussian outputs. Each entry averages four repetitions within each of ten problems and then averages across problems. Both references include the retained endpoint noise.}
\label{tab:gaussian_individual_moments}
\resizebox{\textwidth}{!}{%
\begin{tabular}{lcccc}
\toprule
Method & Fitted mean MSE & Fitted variance MSE & True mean MSE & True variance MSE\\
\midrule
Analytic posterior score (fitted prior) & 0.011516 & 0.005614 & 0.018309 & 0.005799\\
AECSF & 0.015528 & 0.008179 & 0.025495 & 0.009191\\
AECSF without resampling & 0.093729 & 0.028066 & 0.105807 & 0.029772\\
Noisy-state likelihood guidance & 0.020602 & 0.009073 & 0.024058 & 0.010208\\
Denoised-state likelihood guidance & 0.058519 & 0.009065 & 0.047680 & 0.009776\\
\bottomrule
\end{tabular}}
\end{table}

\paragraph{Direct proposal and score diagnostics.}
\label{app:gaussian_direct_diagnostics}
We perform two comparisons using ten Gaussian observation problems, with four
repetitions per problem and five saved reverse stages. First, we compare full
AECSF with AECSF without resampling, which retains incremental weights,
proposal movement, and reverse sampling. This comparison assesses how
resampling changes the proposal approximation of the fitted Gaussian
reference at the corresponding temperature. For normalized weights \(W_m\),
we use
\[
\widehat\mu=\sum_m W_mz_m,\qquad
E_\mu=d^{-1}\|\widehat\mu-\mu^{\rm fit}_{\rho}\|_2^2.
\]
The diagonal-variance error is defined analogously using weighted
population variances. Table~\ref{tab:gaussian_proposal_moments} reports
the resulting errors.

\begin{table}[htbp]
\centering\small
\caption{Proposal-moment errors relative to the fitted Gaussian reference at
the current temperature. Errors are averaged over five stages and four
repetitions within each problem, then over ten problems.}
\label{tab:gaussian_proposal_moments}
\begin{tabular}{lcc}
\toprule
Method & Mean MSE & Diagonal-variance MSE\\
\midrule
AECSF & 0.107085 & 0.078392\\
AECSF without resampling & 0.193791 & 0.120003\\
\bottomrule
\end{tabular}
\end{table}

Second, we compare reverse-particle-guided resampling with standard SMC
resampling within AECSF. The latter selects ancestors from the tempered
weights and assigns equal weights to descendants, retaining the same
proposal moves and reverse sampler. To compare the two score estimators
at the same locations, we combine each method's 20 current reverse particles
into a shared set \(Q\) of 40 evaluation points. Both proposal ensembles
estimate the score at all 40 points, using the same analytic noised-posterior
score \(s^{\rm fit}\) based on the fitted prior and current observation:
\[
\operatorname{NMSE}(Q)=
\frac{\sum_{x\in Q}\|\widehat s(x)-s^{\rm fit}(x)\|_2^2}
{\max\{\sum_{x\in Q}\|s^{\rm fit}(x)\|_2^2,10^{-12}\}}.
\]
We evaluate each set of proposals after movement at stages 8, 28, 48, 68, and 88. Within each observation problem, we give the five stages and four repetitions equal weight when averaging NMSE. We then average over the ten problems.

\begin{table}[htbp]
\centering\small
\caption{Score NMSE at the same 40 reverse-particle locations for both
estimators. These locations are drawn from the two methods' current
sampling trajectories.}
\label{tab:gaussian_score_diagnostics}
\begin{tabular}{lc}
\toprule
Ancestor selection & Score NMSE\\
\midrule
Reverse-particle-guided resampling & 0.531663\\
Standard SMC resampling & 0.557599\\
\bottomrule
\end{tabular}
\end{table}

Guided resampling reduces score NMSE by 4.65\%, with lower error in all ten
problems. This diagnostic measures accuracy where the two samplers currently
require conditional estimates. We also evaluate the resulting output
distributions against the fitted-posterior reference. Relative to standard
SMC resampling, guided resampling reduces pooled and cross-repetition
endpoint energy discrepancies by 31.7\% and 27.6\%, respectively.

\subsection{Evaluation metrics and aggregation}

State RMSE is the primary accuracy measure and mean componentwise CRPS measures probabilistic accuracy. Normalized observation misfit is reported for Lorenz--96. SSR compares ensemble spread with the error of the ensemble mean. Computational cost is reported for the timing interval specified for each experiment.

The per-cycle RMSE, CRPS, and normalized observation misfit are
\begin{align}
    \mathrm{RMSE}_n &= \left(\frac{1}{d}\norm{\bar x_n^a-x_n^{\rm true}}^2\right)^{1/2},\label{eq:rmse_metric}\\
    \mathrm{CRPS}_n
    &= \frac{1}{Jd}\sum_{j=1}^{J}\norm{x_n^{a,j}-x_n^{\rm true}}_1
    -\frac{1}{2J^2d}\sum_{j=1}^{J}\sum_{k=1}^{J}\norm{x_n^{a,j}-x_n^{a,k}}_1,\label{eq:crps_metric}\\
    \mathrm{NOM}_n
    &= \frac{1}{\sigma_y}\left(\frac{1}{m_n}\norm{\mathcal H_n(\bar x_n^a)-y_n}^2\right)^{1/2}.\label{eq:normalized_obs_misfit_metric}
\end{align}

The CRPS expression in \eqref{eq:crps_metric} applies ensemble CRPS to each state component and then averages over components \citep{gneiting2007strictly}.
Here, \(\sigma_y\) is the observation-noise standard deviation in the reported homoscedastic protocols.
Normalized observation misfit measures agreement between the analysis mean and the assimilated observation in units of observation noise. A value of one corresponds to an RMS observation residual equal to the observation-noise standard deviation.

For each run, RMSE and CRPS are averaged over the specified evaluation
window: cycles 50--199 for the main Lorenz--96 comparisons,
950--999 for the original Kuramoto--Sivashinsky analysis comparisons,
and cycles 41--50 for the sparse Navier--Stokes table; its accuracy curves cover all 50 cycles.
The free-forecast comparison first averages the five forecast origins
within each run at each physical lead.

In the sparse Navier--Stokes experiment, RMSE is evaluated on velocity,
whereas CRPS and SSR are evaluated on vorticity.
In the KS forecast table, coverage is the empirical coverage of the
central 90\% ensemble interval.

Let \(\mathcal W\) denote the corresponding evaluation cycles or forecast origins. Define the ensemble variance \(v_{n,k}=J^{-1}\sum_j(x_{n,k}^{a,j}-\bar x_{n,k}^a)^2\) and ensemble-mean error \(e_{n,k}=\bar x_{n,k}^a-x_{n,k}^{\rm true}\). The spread--skill ratio pools their second moments over the evaluation window:
\[
\mathrm{SSR}=\sqrt{\frac{\sum_{n\in\mathcal W}\sum_k v_{n,k}}{\sum_{n\in\mathcal W}\sum_k e_{n,k}^2}}.
\]
The tables report the mean and sample standard deviation of the seed-level summaries.
All prescribed seeds are included in each reported comparison.

Recursive comparisons share the physical inputs and initial ensemble within each paired evaluation; subsequent forecasts are propagated from each method's own analyses. AECSF uses independent deterministic random streams for resampling, Langevin movement, and reverse diffusion.

\subsection{Shared implementation settings and timing}

\begin{table}[htbp]
\centering\small
\caption{AECSF configurations. \(M\) counts proposal members and \(J\) counts reverse particles; the forecast ensemble has \(J\) members. The training-free configurations shown here use \(M=J\); Table~\ref{tab:l96_proposal_capacity} varies \(M\) at fixed \(J\). All configurations use stratified resampling.}
\label{tab:ecsf_settings_summary}
\begin{tabular}{lccc}
\toprule
Setting & Gaussian & Lorenz--96 & Kuramoto--Sivashinsky\\
\midrule
\(M/J\) & 20/20 & 20/20 & \(J/J\)\\
Reverse steps & 100 & 100 & \(T\)\\
Weight updates / resampling / move calls & 99/50/98 & 99/50/98 & \((T-1)/(T/2)/(T-2)\)\\
Langevin steps per move & 3 & 3 & 3\\
Langevin step size \(h\) & 0.02 & 0.006 & 0.002\\
\(\epsilon_a\) & 0.001 & 0.0005 & 0.001\\
\(\epsilon_b\) & 0.025 & 0.0125 & 0.015\\
\bottomrule
\end{tabular}
\end{table}

For the configurations in Table~\ref{tab:ecsf_settings_summary}, the forecast ensemble has \(J=M\) members. For coordinate \(q\), the diagonal forecast approximation uses \(m_q=M^{-1}\sum_m X^f_{mq}\) and
\[
v_q=\max\!\left\{\max\!\left[\frac1M\sum_m(X^f_{mq}-m_q)^2,10^{-12}\right]+10^{-5},10^{-3}\right\}.
\]
The fitted mean and variance remain fixed throughout the analysis update.

In these configurations, the grid is uniform in logSNR. For \(T\) score evaluations, the first evaluation at \(\tau=1\) uses the initial proposal ensemble. Numbering the remaining \(T-1\) stages from zero, resampling occurs at even-numbered stages and movement at stages \(0,\ldots,T-3\). For the even values of \(T\) used here, each analysis makes \(T/2\) resampling calls and \(T-2\) move calls, with \(3(T-2)\) Langevin steps. Weight updates continue when resampling is omitted; the final score stage resamples but does not move. Gaussian and Lorenz--96 use \(T=100\); the Kuramoto--Sivashinsky values of \(T\) and \(J\) are given in Appendix~\ref{app:ks_nonlinear_scaling}.

Analysis-update timings use CUDA-synchronized wall-clock measurements
on an RTX 4090, excluding forecast propagation, metric evaluation,
and file output. The original Kuramoto--Sivashinsky analysis timings
include per-cycle preparation and sampling; controlled Gaussian timings
also include per-run model construction.

For Lorenz--96, LETKF timings cover the complete analysis update in
float64. EnSF and AECSF report reverse-sampling time, including AECSF
proposal updates; Binder's timing interval is specified in
Appendix~\ref{app:binder_transfer}.
The separate KS free-forecast comparison uses an RTX 3090 and reports
the cost of one assimilation cycle plus the complete ensemble forecast
to physical lead 8.

\subsection{Lorenz--96 nonlinear filtering}
\label{app:l96_protocol}

The Lorenz--96 system provides a scalable chaotic benchmark for recursive data assimilation \citep{lorenz1996predictability}:
\begin{equation}
    \frac{\dd x_i}{\dd t}=(x_{i+1}-x_{i-2})x_{i-1}-x_i+F,
    \qquad i=1,\ldots,d,
    \label{eq:l96}
\end{equation}
with periodic indexing and forcing \(F=8\). The observation operator is nonlinear and saturating:
\begin{equation}
    y_n = \arctan(Px_n)+\varepsilon_n,
    \qquad \varepsilon_n\sim\N(0,R),
    \label{eq:l96_obs}
\end{equation}
where \(P\) selects observed components.
This setting tests whether the method can assimilate observations that become weakly informative when state magnitudes are large.
The integration time step is \(0.01\), with observations every five steps, corresponding to \(0.05\) physical time units.
AECSF propagates its forecast ensemble with the classical fourth-order Runge--Kutta method in double precision.

\paragraph{Methods and configuration selection.}
AECSF, EnSF, LETKF, and the method of Binder et al. share physical inputs and the initial ensemble on seeds 10--19. Subsequent forecasts depend on each method's recursive analyses. The EnSF and LETKF configurations were first selected from complete \(10\times10\) grids at \(d=1000\) on seeds 0--9, ranked by mean post-burn-in RMSE. We additionally checked five nearby EnSF configurations at \(d=10{,}000\) on development seeds 0--9. The original configuration, \((\epsilon_a,\epsilon_b)=(0.6,0.025)\), retained the lowest mean RMSE.

\phantomsection\label{app:l96_letkf_implementation}%
LETKF uses cyclic localization radius 2, neighbor size 5, and inflation 1.1. LETKF distributes independent local analyses across eight CUDA streams. The implementation reuses cached workspaces and CUDA graphs containing up to eight local updates, with solver error codes checked collectively after each full-field update. The development results appear below.

AECSF uses \(h=0.006\) and three Langevin steps per scheduled move, with the weighted update and logSNR grid specified in Table~\ref{tab:ecsf_settings_summary}. The \(d=10{,}000\) experiment has 200 cycles, full arctangent observations, noise standard deviation 0.05, and evaluation cycles 50--199.

\begin{figure}[H]
\centering
\includegraphics[trim=0 1 0 2,clip,width=\textwidth]{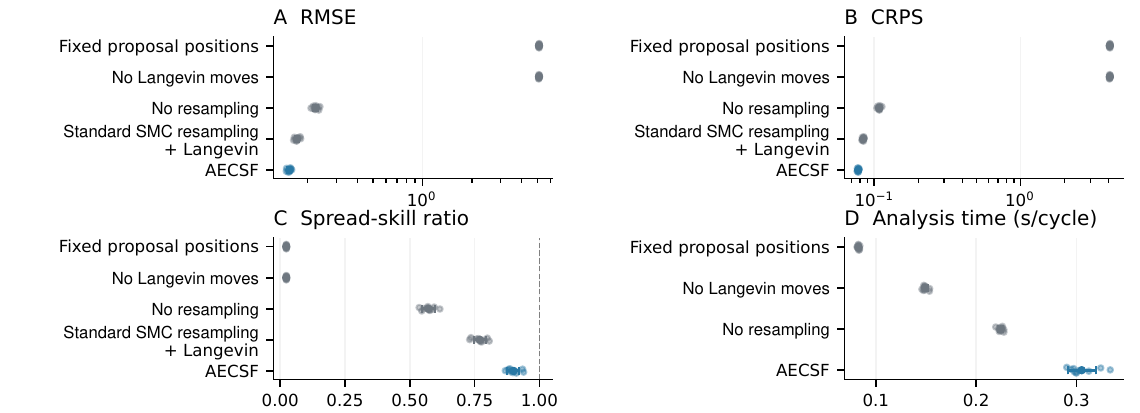}
\caption{Lorenz--96 proposal-update ablations. (A--C) Accuracy and calibration;
(D) analysis time. Large points and bars show means and sample standard deviations
over ten paired runs. Small points show individual runs.}
\label{fig:l96_candidate_components}
\end{figure}

\paragraph{Component controls.}
In the paired ablation in Figure~\ref{fig:l96_candidate_components}, guided selection
reduces RMSE by 10.2\% and CRPS by 7.5\% relative to standard SMC resampling.

All five variants use the same prior approximation, diffusion grid,
observation model, and incremental weight updates. They differ only in
the proposal-location updates and the associated ancestor weights:
\begin{center}
\resizebox{\textwidth}{!}{%
\begin{tabular}{llll}
\toprule
Variant & Ancestor law & Weights after resampling & Movement\\
\midrule
AECSF with fixed proposal locations & None & Not applicable & No\\
AECSF without Langevin moves & \(s_i\) & \(W_m^+\propto\omega_{i,A_m}/s_{i,A_m}\) & No\\
AECSF without resampling & None & Not applicable & Yes\\
AECSF with standard SMC resampling & \(\omega_i\) & \(1/M\) & Yes\\
AECSF & \(s_i\) & \(W_m^+\propto\omega_{i,A_m}/s_{i,A_m}\) & Yes\\
\bottomrule
\end{tabular}}
\end{center}
Without resampling, the updated weights \(\omega_i\) are retained.
Movement preserves the carried weights and refreshes the likelihoods.
The variant with standard SMC resampling uses
\(s_{i,m}=\omega_{i,m}\), producing weights \(1/M\) after resampling;
its proposal moves and reverse sampler are retained. Its comparison with AECSF tests the
joint effect of query-guided selection and its associated importance
correction within the reverse-sampling algorithm.\footnote{For the Lorenz--96 standard SMC resampling control, the recorded
analysis time includes the retained bridge calculation. Ancestor
selection uses only the target weights.}

For all ten seeds, AECSF's SSR is closer to one than that of the standard SMC resampling control. The control has a normalized observation misfit closer to one on average (0.9497 for the control and 0.9209 for AECSF).

\begin{table}[t]
\centering
\footnotesize
\setlength{\tabcolsep}{1.5pt}
\caption{Lorenz--96 results (\(d=10{,}000\), 20 forecast and output members).
Entries are means \(\pm\) sample SDs of run-level summaries over cycles
50--199. The original comparison and the new SMC comparison each contain
ten runs; the Binder conditioning comparison contains three paired runs.
SMC uses 320 internal particles; the displayed AECSF configurations use
20 shared proposals. Timed update intervals are specified in
Appendix~\ref{app:experimental_protocols}. Standard SMC resampling uses the tempered weights and equal-weight descendants. LETKF uses parallel local
updates with CUDA graphs. Dashes denote unreported metrics.}
\label{tab:l96_full_comparisons}
\begin{tabular}{lccccc}
\toprule
Method & RMSE \(\downarrow\) & CRPS \(\downarrow\) & SSR \((\to1)\) &
\shortstack{Norm. obs.\\misfit} & \shortstack{Timed update/\\cycle (s)}\\
\midrule
\multicolumn{6}{l}{\textit{Original benchmark comparison}}\\
EnSF & \(0.2418\pm0.0006\) & \(0.1432\pm0.0002\) & \(1.7366\pm0.0042\) & \shortstack{\(1.8180\)\\\(\pm0.0028\)} & \(0.0485\pm0.0010\)\\
LETKF & \(0.1263\pm0.0116\) & \(0.0578\pm0.0015\) & \(0.6119\pm0.0604\) & \shortstack{\(1.0393\)\\\(\pm0.0349\)} & \(0.5093\pm0.0014\)\\
Binder et al. & \(5.1323\pm0.0082\) & \(4.1428\pm0.0076\) & \shortstack{\((24.66\pm13.29)\)\\\(\times10^{-5}\)} & \shortstack{\(28.4641\)\\\(\pm0.0390\)} & \(0.1507\pm0.0015\)\\
AECSF & \(0.1545\pm0.0039\) & \(0.0776\pm0.0007\) & \(0.8998\pm0.0232\) & \shortstack{\(0.9209\)\\\(\pm0.0042\)} & \(0.3052\pm0.0139\)\\
\midrule
\multicolumn{6}{l}{\textit{Tempered-SMC and ancestor-selection comparison}}\\
\shortstack[l]{Factorized tempered\\SMC} & \(2.7330\pm0.0377\) & \(1.7974\pm0.0340\) & \(0.0125\pm0.0002\) & \shortstack{\(14.4999\)\\\(\pm0.1912\)} & \(1.2507\pm0.0032\)\\
\shortstack[l]{SMC + transport\\and marginal\\moment matching} & \(2.6054\pm0.0265\) & \(1.6862\pm0.0240\) & \(0.0133\pm0.0001\) & \shortstack{\(13.7825\)\\\(\pm0.1407\)} & \(1.2602\pm0.0031\)\\
\shortstack[l]{AECSF with standard\\SMC resampling} & \(0.1708\pm0.0054\) & \(0.0840\pm0.0010\) & \(0.7768\pm0.0252\) & \shortstack{\(0.9475\)\\\(\pm0.0083\)} & \(0.2739\pm0.0027\)\\
\shortstack[l]{AECSF with reverse-\\particle-guided\\resampling} & \(0.1531\pm0.0044\) & \(0.0776\pm0.0008\) & \(0.9079\pm0.0266\) & \shortstack{\(0.9194\)\\\(\pm0.0052\)} & \(0.2851\pm0.0037\)\\
\midrule
\multicolumn{6}{l}{\textit{Binder conditioning comparison}}\\
Binder et al. & \(5.1277\pm0.0060\) & \(4.1372\pm0.0054\) & \shortstack{\((1.9293\pm0.6666)\)\\\(\times10^{-4}\)} & --- & ---\\
Enhanced Binder & \(1.4312\pm0.0143\) & \(0.7589\pm0.0028\) & \(0.0422\pm0.0007\) & --- & ---\\
AECSF & \(0.1538\pm0.0018\) & \(0.0777\pm0.0004\) & \(0.9040\pm0.0104\) & --- & ---\\
\bottomrule
\end{tabular}
\end{table}

\paragraph{Connection to SMC when proposal locations are fixed.}
\label{app:fixed_proposal_smc_connection}
Consider the fixed-location control with resampling and movement disabled.
Let \(W_m^{\mathrm{init}}\) denote its initial weights at tempering level
zero, and write \(L_m=g_n(y_n\given z_m)\).
The incremental updates incorporate \(L_m^{\rho_i}\), while conditional-mean
estimation supplies the remaining \(L_m^{1-\rho_i}\):
\[
\begin{aligned}
W_{i,m}&\propto W_m^{\mathrm{init}}L_m^{\rho_i},\\
W_{i,m}L_m^{1-\rho_i}
&\propto W_m^{\mathrm{init}}L_m^{\rho_i+(1-\rho_i)}
=W_m^{\mathrm{init}}L_m.
\end{aligned}
\]
Equation~\eqref{eq:main_frozen_bank_score} therefore gives the score of
\[
\widehat q_{n,i}^M(x;y_n)
\propto\sum_m W_m^{\mathrm{init}}L_m K_i(x,z_m),
\]
where \(K_i\) is the Gaussian compatibility kernel from
Section~\ref{subsec:main_proposal_updates}.
For the uniform initial weights used here, this is Gaussian noising
of the bootstrap likelihood-weighted empirical posterior.
Full AECSF adapts the shared integration proposals using the evolving
reverse particles; reverse sampling produces the analysis ensemble.

\paragraph{Comparison with tempered SMC.}
The additional ten-run comparison uses the same L96 dimension,
observation model, 200-cycle horizon, and evaluation window as the
main benchmark. All methods use 20 forecast and output members.
With a diagonal Gaussian forecast and a coordinatewise likelihood, the tempered
target factorizes across coordinates. The independent SMC filter uses this
factorization with 320 internal particles, a conditional-ESS threshold of 0.5,
and three Metropolis-adjusted Langevin moves per temperature. We fixed these
settings on development runs. The transport variant uses the same internal
particles and matches their weighted marginal means and variances when it
constructs the 20 output members.

Both AECSF variants use 20 shared proposals, the original 100-step reverse
schedule, and 294 Langevin steps with \(h=0.006\). They differ only in how they
select ancestors and assign the corresponding descendant weights. After the
common initialization, each method propagates its own analysis ensemble.
We report the measured cost of each configuration.

\paragraph{Direct posterior sampling control.}
At the same Langevin step count and step size, AECSF reduces mean RMSE
by 16.2\% and CRPS by 1.1\%, at 2.23 times the sampling cost
(Table~\ref{tab:l96_direct_posterior}).

Direct posterior ULA moves the 20 output members under the full posterior drift at every step. AECSF without resampling still estimates conditional scores and runs the reverse sampler. The direct ULA control uses neither resampling nor incremental weights. We match its 294 steps and step size of 0.006 to the AECSF configuration and use paired physical inputs on seeds 10--19. AECSF has lower RMSE on all ten seeds and lower CRPS on eight.

\begin{table}[htbp]
\centering\small
\caption{Matched-step Lorenz--96 comparison with 20 members at dimension 10,000. Both methods use 294 Langevin steps per analysis with step size 0.006. Entries are means and sample standard deviations across seeds 10--19, using cycles 50--199 within each seed. Timing brackets the sampling update and excludes model construction and final analysis processing.}
\label{tab:l96_direct_posterior}
\resizebox{\textwidth}{!}{%
\begin{tabular}{lccccc}
\toprule
Method & RMSE \(\downarrow\) & CRPS \(\downarrow\) & SSR \(\to1\) & Norm. obs. misfit & Sampling-update time (s/cycle)\\
\midrule
AECSF & \(0.154482\pm0.003892\) & \(0.077631\pm0.000740\) & \(0.899823\pm0.023152\) & \(0.920890\pm0.004196\) & \(0.305238\pm0.013899\)\\
Direct posterior ULA & \(0.184311 \pm 0.007514\) & \(0.078474 \pm 0.001215\) & \(0.596951 \pm 0.023914\) & \(1.042469 \pm 0.013634\) & \(0.136624 \pm 0.003211\)\\
\bottomrule
\end{tabular}}
\end{table}

We calibrated direct ULA runtime at 294, 588, and 882 steps, then set a budget of 1,350 steps to approximately match AECSF's analysis time. Development runs on seeds 0--9 tested step sizes 0.003, 0.006, and 0.012 at both 294 and 1,350 steps. At 1,350 steps, the lowest mean development RMSE occurred at step size 0.012. The evaluation compares this configuration and AECSF using common truth, observations, and initial ensembles. Each method then propagated its own analysis ensemble. On the same RTX 4090, timing covered one complete reverse run for AECSF or all 1,350 ULA steps per cycle. CUDA synchronization bracketed each call; forecast propagation and fitting of the forecast density were outside the timed interval.

\begin{table}[htbp]
\centering\small
\caption{Approximately time-matched posterior sampling on Lorenz--96 with 20 members in 10,000 dimensions. AECSF uses 294 Langevin steps; direct ULA uses 1,350 steps with step size 0.012. Entries are means and sample standard deviations across ten seeds, using cycles 50--199 within each seed.}
\label{tab:l96_direct_cost}
\resizebox{\textwidth}{!}{%
\begin{tabular}{lccccc}
\toprule
Method & RMSE \(\downarrow\) & CRPS \(\downarrow\) & SSR \(\to1\) & Norm. obs. misfit & Update time (s/cycle)\\
\midrule
AECSF & \(0.154968\pm0.004083\) & \(0.077842\pm0.000623\) & \(0.896359\pm0.024766\) & \(0.921380\pm0.005849\) & \(0.289720\pm0.001397\)\\
Direct posterior ULA & \(0.171986 \pm 0.001801\) & \(0.094169 \pm 0.000456\) & \(1.116732 \pm 0.010192\) & \(1.149319 \pm 0.002589\) & \(0.296331 \pm 0.000446\)\\
\bottomrule
\end{tabular}}
\end{table}

\paragraph{Sensitivity to proposal movement.}
After selecting the main configuration, we evaluated sensitivity to the
allocation and step size of Langevin moves using five configurations in the
10,000-dimensional Lorenz--96 benchmark with 20 members on seeds 10--19,
keeping the remaining settings fixed. Every configuration uses 294 Langevin steps
across 98 movement stages. Besides three steps at every stage, we assign
four steps to the first 49 stages and two to the last 49, or reverse
this allocation. We also vary the step size by $25\%$ around 0.006
while retaining three steps per stage.
The main configuration has the lowest mean RMSE among these five settings;
allocating more steps later lowers CRPS in all ten paired seeds, while its
mean RMSE increases by $0.20\%$ and its SSR moves further below one.
Relative to the main configuration, the largest changes in mean RMSE and
CRPS are approximately $1.44\%$ and $2.56\%$,
respectively (Table~\ref{tab:l96_movement_sensitivity}).

\begin{table}[htbp]
\centering\small
\caption{Local movement sensitivity on Lorenz--96. Entries are means
across seeds 10--19 over cycles 50--199. Every setting uses 294 Langevin
steps; varying the step size also changes the total Langevin integration time.}
\label{tab:l96_movement_sensitivity}
\begin{tabular}{lrrrr}
\toprule
Steps in first/last 49 stages & $h$ & RMSE $\downarrow$ & CRPS $\downarrow$ & SSR $\to1$\\
\midrule
3/3 & 0.0060 & 0.154482 & 0.077631 & 0.899823\\
4/2 & 0.0060 & 0.156707 & 0.079617 & 0.939406\\
2/4 & 0.0060 & 0.154794 & 0.077192 & 0.872296\\
3/3 & 0.0045 & 0.154682 & 0.078449 & 0.931029\\
3/3 & 0.0075 & 0.155021 & 0.077493 & 0.879071\\
\bottomrule
\end{tabular}
\end{table}

\paragraph{Proposal capacity at fixed output size.}
We vary the shared proposal ensemble size in a 1,000-dimensional
Lorenz--96 experiment, keeping the forecast ensemble, reverse particles,
and output ensemble at 20 members. We use the observation model and
remaining settings of the Lorenz--96 benchmark. The proposals are
initialized by evenly replicating the forecast members with uniform
weights and then updated using the same resampling and movement schedule.
Each proposal receives 294 Langevin steps across 98 stages.
Table~\ref{tab:l96_proposal_capacity} reports ten-seed results.
All larger proposal ensembles lower mean RMSE and CRPS relative to
\(M=20\), although the improvement is not monotonic in \(M\).
At \(M=400\), mean RMSE and CRPS decrease by 4.08\% and 2.82\%,
respectively, with a 1.19\% increase in measured sampling time;
SSR changes from 0.8935 to 0.9598. Increasing proposal capacity thus
improves the estimates while retaining a 20-member output ensemble.

\begin{table}[htbp]
\centering\small
\caption{Proposal capacity in Lorenz--96 with \(d=1000\) and
\(J=20\). Entries are means over seeds 10--19, each summarized over
cycles 50--199. All configurations run on one RTX 4090.
The per-proposal Langevin budget is fixed, so total gradient work
increases with \(M\). Time measures the sampling update.}
\label{tab:l96_proposal_capacity}
\begin{tabular}{rrrrrr}
\toprule
\(M\) & \(M/J\) & RMSE $\downarrow$ & CRPS $\downarrow$ & SSR $\to1$ & s/update\\
\midrule
20  & 1  & 0.153809 & 0.079206 & 0.8935 & 0.2995\\
40  & 2  & 0.148537 & 0.077347 & 0.9335 & 0.3018\\
80  & 4  & 0.148571 & 0.077442 & 0.9395 & 0.2998\\
160 & 8  & 0.148918 & 0.077408 & 0.9428 & 0.2969\\
240 & 12 & 0.150529 & 0.077595 & 0.9352 & 0.2985\\
320 & 16 & 0.147748 & 0.077230 & 0.9571 & 0.2986\\
400 & 20 & 0.147535 & 0.076975 & 0.9598 & 0.3031\\
\bottomrule
\end{tabular}
\end{table}

\subsubsection{Tuning grids}
\label{app:l96_tuning_grids}

\begin{figure}[p]
\centering
\includegraphics[width=0.86\textwidth]{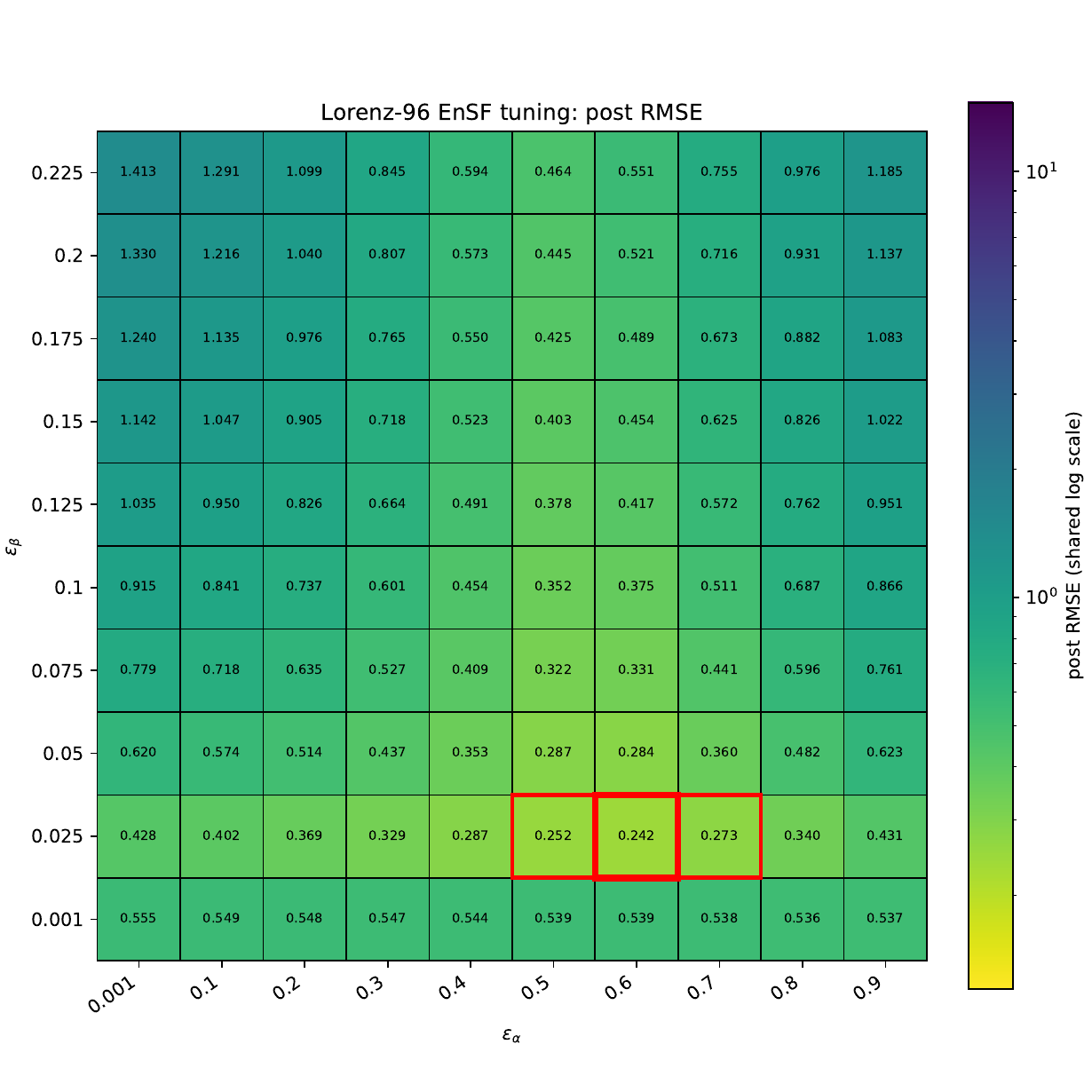}
\caption{Lorenz--96 EnSF \(d=1000\) tuning grid over seeds \(0,\ldots,9\). The rank-1 configuration transferred to Table~\ref{tab:l96_representative_results} is \(\epsilon_a=0.6,\epsilon_b=0.025\).}
\label{fig:l96_ensf_tuning_grid}
\end{figure}

\begin{figure}[p]
\centering
\includegraphics[width=0.86\textwidth]{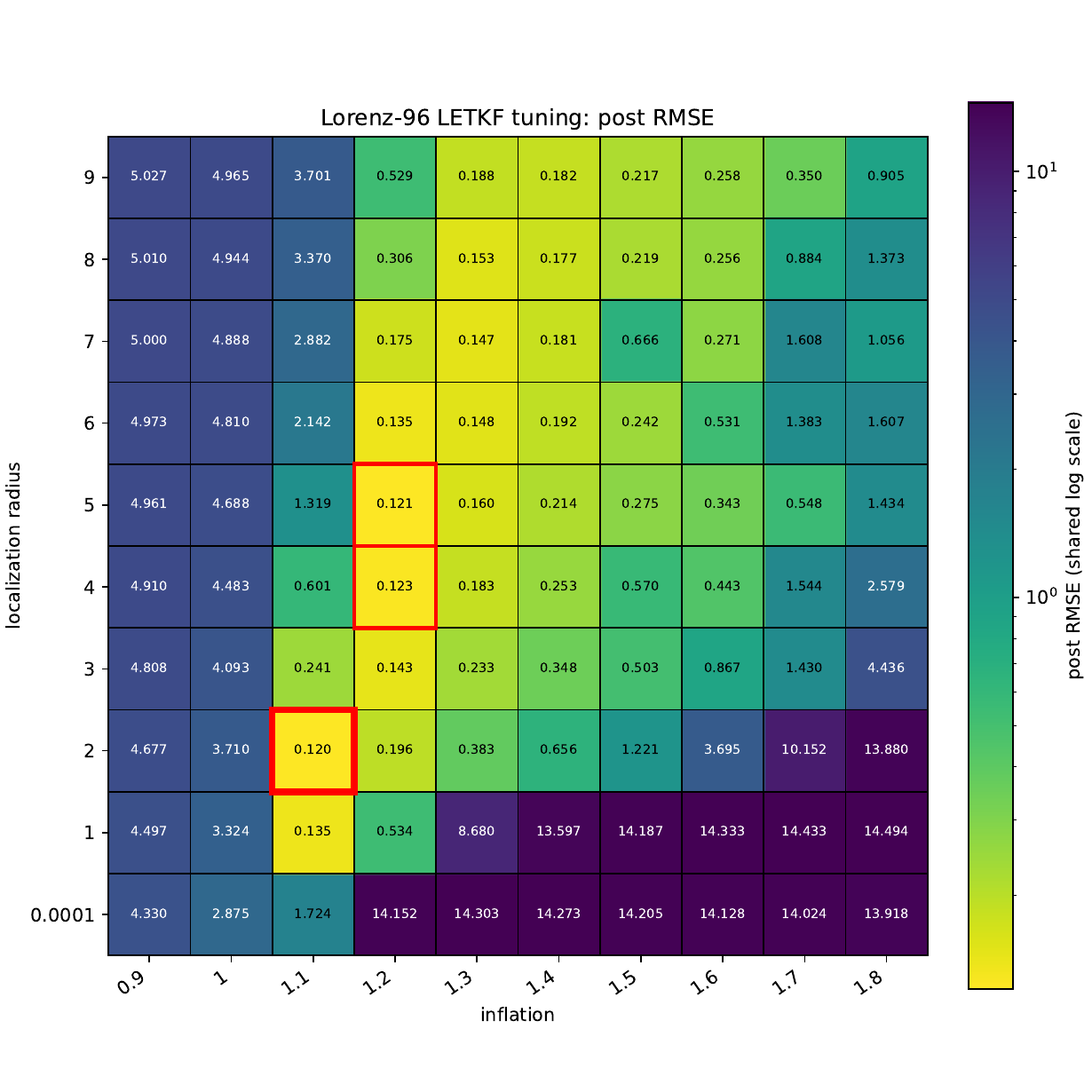}
\caption{Lorenz--96 LETKF \(d=1000\) tuning grid over seeds \(0,\ldots,9\). The rank-1 configuration transferred to Table~\ref{tab:l96_representative_results} uses localization radius \(2\), neighbor size \(5\), and inflation \(1.1\).}
\label{fig:l96_letkf_tuning_grid}
\end{figure}

Figures~\ref{fig:l96_ensf_tuning_grid} and~\ref{fig:l96_letkf_tuning_grid} show the baseline Lorenz--96 tuning grids at \(d=1000\). Cells aggregate post-burn-in RMSE over seeds 0--9. Red rectangles mark the three lowest-RMSE configurations, with a thicker rectangle for rank 1.

\clearpage
\paragraph{Binder reproduction and transfer protocol.}
\label{app:binder_transfer}
We implemented the conditional score from the joint KDE of \citet{binder2026closedform} and checked its numerical evaluation against analytic references. Under the paper's protocol with \(d=10\), \(M=20\), and \((h_x,h_y)=(0.2,1)\), one 500-cycle run yielded RMSE 3.0198 without ensemble collapse.

For the common benchmark in 10,000 dimensions, bandwidths \((h_x,h_y)=(0.2,33.5)\) were selected during development and fixed before evaluation on seeds \(10,\ldots,19\). Forecast states and simulated observations were centered coordinatewise and divided by the maximum absolute centered deviation. Coordinates with zero deviation used unit scale. The implementation uses float64 GPU state updates with the Dormand--Prince RK45 coefficients and adaptive step-size rules of SciPy, at relative and absolute tolerances \(10^{-6}\) and \(10^{-8}\). Analysis timing includes simulated observations, normalization, integration, and required device transfers. The common benchmark differs from the original protocol in observation noise, initialization, assimilation interval, and process noise. Earlier runs of the common protocol in low dimensions also showed ensemble contraction.

\paragraph{How Binder's method and AECSF estimate the conditional score.}
\label{app:binder_kde_comparison}
In Binder's KDE coordinates, let \((z_m,\widetilde y_m)\) denote the
forecast--observation pairs, with state and observation kernel
covariances \(H\) and \(G\).
Conditioning on \(y_n\), then adding Gaussian noise of variance
\(\sigma_{B,i}^2\), gives
\[
\begin{aligned}
p_B(z\mid y_n)
&\propto\sum_m\phi_H(z-z_m)\phi_G(y_n-\widetilde y_m),\\
s_{B,i}(x;y_n)
&=\nabla_x\log\!\left[
\sum_m\phi_G(y_n-\widetilde y_m)
\phi_{H+\sigma_{B,i}^2\I_d}(x-z_m)\right].
\end{aligned}
\]
AECSF's conditional Tweedie estimator has the unclipped mixture-score form in
\eqref{eq:main_frozen_bank_score}.
Holding the current proposals and integration weights fixed during
differentiation gives
\[
\widehat s_{n,\tau_i}^a(x;y_n)
=\nabla_x\log\!\left[
\sum_m W_{i,m}g_n(y_n\given z_i^m)^{1-\rho_i}
K_i(x,z_i^m)\right].
\]
Binder conditions through an observation kernel and keeps forecast
centers fixed within an analysis; component score contributions vary
with \(x\).
AECSF uses the original likelihood and updates proposal locations and
integration weights during reverse sampling.

\paragraph{Enhanced Binder.}
To assess observation conditioning, we condition Binder's forecast-state KDE directly on the original arctangent likelihood:
\[
p_{\mathrm{enh},h}(z\mid y)\propto
L(y\mid z)\,\frac{1}{M}\sum_m\phi_{H_h}(z-z_m).
\]
Numerical integration updates both the within-kernel conditional distributions
and mixture weights, followed by direct sampling. The state bandwidth
\(h=0.5\) was fixed before the high-dimensional evaluation and used at both
\(d=10\) and \(d=10{,}000\). With 20 members and the same observation-noise
SD of 0.05, Enhanced Binder's RMSE increased from 0.495 to 1.431,
whereas AECSF's changed from 0.140 to 0.154.
The median component ESS decreased from 7.3--7.7 to approximately one.
Each component represents an entire state vector, and its posterior weight
multiplies coordinate-wise evidence; accumulated evidence differences can
therefore concentrate the global mixture weights and remove
between-component variability.

\begin{table}[htbp]
\centering\small
\caption{Binder conditioning comparison at two dimensions. Each comparison
uses three paired runs, 20 members, and observation-noise SD 0.05.
Enhanced Binder uses state bandwidth \(h=0.5\) in both dimensions.}
\label{tab:binder_conditioning_dimensions}
\begin{tabular}{rlccc}
\toprule
Dimension & Method & RMSE & CRPS & SSR\\
\midrule
10 & Binder et al. & 7.801 & 6.057 & 2.303\\
10 & Enhanced Binder & 0.495 & 0.228 & 1.244\\
10 & AECSF & 0.140 & 0.083 & 1.178\\
10,000 & Binder et al. & 5.128 & 4.137 & 0.000193\\
10,000 & Enhanced Binder & 1.431 & 0.759 & 0.042\\
10,000 & AECSF & 0.154 & 0.078 & 0.904\\
\bottomrule
\end{tabular}
\end{table}

Enhanced conditioning recovers 74.3\% of the high-dimensional RMSE gap
between Binder and AECSF in this paired comparison. This cross-method comparison
reflects differences in prior representation, likelihood conditioning, and
proposal adaptation. The within-AECSF control in
Table~\ref{tab:l96_full_comparisons} isolates guided ancestor selection and
its importance correction.

\subsection{Kuramoto--Sivashinsky filtering with nonlinear observations}
\label{app:ks_nonlinear_scaling}

We use the periodic Kuramoto--Sivashinsky equation \citep{sivashinsky1977nonlinear}
\begin{equation}
u_t=-u u_x-u_{xx}-u_{xxxx},\qquad x\in[0,128\pi),
\label{eq:kse}
\end{equation}
discretized on 1,024 spatial coordinates with a Fourier pseudospectral ETDRK4 solver and time step 0.25. The nonlinear term is evaluated by transforming the squared physical-space field back to Fourier space, as in the reference implementation. Observations are
\begin{equation}
y_{n,k}=\arctan(u_{n,k})+\varepsilon_{n,k},\qquad
\varepsilon_n\sim\mathcal N(0,0.1^2 I_{1024}).
\label{eq:ks_arctan_observation}
\end{equation}
Four model steps separate successive observations, giving an assimilation interval of one time unit. The first observation is assimilated at the initial time. The model propagation is deterministic.

\paragraph{Inputs and ensemble initialization.}
The reference initial profile is evolved for 600 steps with the DAPPER initialization routine. For each seed, independent standard Gaussian noise is added to this state, followed by 2,000 model steps before assimilation begins. Initial ensemble members are obtained by adding independent standard Gaussian perturbations to the truth at this point. The first \(J\) members of a saved 160-member ensemble are used for each ensemble-size setting. Methods share the saved truth, observations, and initial ensemble at seeds 30--34.

\paragraph{Compared methods.}
The Kuramoto--Sivashinsky experiment protocol, together with the EnSF, EnFF-OT, and EnFF-F2P implementations and their parameter settings, follows the public EnFF repository \citep{transue2025flowmatching}. We extend this setup with the sampling-step and ensemble-size comparisons specified below. The EnSF path parameters are \((\epsilon_a,\epsilon_b)=(0.5,0.025)\). EnFF-OT uses \(\sigma_{\min}=0.01\) and guidance strength 1; EnFF-F2P uses \(\sigma_{\min}=0.01\) and guidance strength 0.2. EnFF-F2P receives the preceding analysis ensemble when constructing its probability path. The original four methods operate in float32 on an RTX 4090. Computation is divided into blocks of at most 20 queries in the public implementations.

AECSF uses $(\epsilon_a,\epsilon_b)=(0.001,0.015)$, Langevin step size $h=0.002$, three steps per move, Stratified ancestor selection, and a uniform-logSNR grid. At each analysis, AECSF initializes reverse particles from Gaussian noise at $\tau=1$ and standardizes each coordinate to zero sample mean and unit sample SD. Baselines retain their own terminal-noise initialization.

\paragraph{IEnSF-L1 transfer conventions.}
\label{app:ks_iensf_transfer}
We implement IEnSF based on the formulation and implementation details provided by its authors.
For each output coordinate, we use an
overlapping periodic patch of 13 coordinates and retain its centre
score. The forecast covariance, normalized by $J$, is multiplied
entrywise by a Gaspari--Cohn taper with scale 3 and support 6.
Writing this covariance as $C_{\rm GC}$ and the local forecast mean
as $m$, mixture components use
\[
 m_j=m+0.8\sqrt{1-\gamma^2}(x_j-m),\qquad
 \Sigma=\gamma^2 C_{\rm GC},\qquad \gamma=0.25.
\]
The guidance matrix is
$\alpha_t\Sigma(\alpha_t^2\Sigma+\beta_t^2 I)^{-1}$.
The Gaussian reference uses mean $m$ and covariance $C_{\rm GC}$;
it is reinitialized from each cycle's forecast. We use one reverse pass ($L=1$) and invert the covariance matrices within each local patch. The patch construction, reference localization, centre factor, and nonlinear covariance blending specify our adaptation to this benchmark.

For mixture likelihood weights, the arctangent observation is
linearized at each component's conditional clean-state mean. If
$C_{0|t}$ is its conditional covariance and $H_j$ the diagonal
Jacobian, we use
$R+0.7H_j C_{0|t}H_j^\top+0.3\operatorname{diag}(H_j C_{0|t}H_j^\top)$.
Reverse sampling uses Euler--Maruyama with
$\alpha_t=1-0.999t$ and $\beta_t^2=0.001+0.999t$.
There are $N_t$ updates evaluated at $t_k=1-k/N_t$ for
$k=0,\ldots,N_t-1$; the score is not evaluated at zero.
The total score is clipped componentwise at 1,000. The terminal samples and diffusion
increments use separate deterministic streams. IEnSF-L1 starts
from independent standard Gaussian samples, without the
coordinatewise sample standardization used by AECSF. All configurations
retain the same physical input bundles for seeds 30--34, with the
first $J$ members of the saved initial ensemble.

\paragraph{Budgets and evaluation.}
We vary two budgets separately. At \(J=20\), all five methods use step counts \(\{20,50,100,200,500\}\). At 20 steps, we vary ensemble size over \(\{10,20,40,80,160\}\) for AECSF, EnSF and EnFF, and over \(\{10,20,40,80\}\) for IEnSF-L1. The IEnSF-L1 comparison stops at 80 members because its particle--mixture evaluations scale quadratically with ensemble size. Its measured analysis time rises from 1.219 to 4.218 seconds between 40 and 80 members. We use \(M=J\) for AECSF and \(L=1\) for IEnSF-L1, and evaluate the common 20-member, 20-step configuration once per method.

A nominal step count corresponds to different operations across samplers. EnSF and EnFF perform \(T-1\) sampling updates. AECSF evaluates the score \(T\) times and takes \(3(T-2)\) Langevin steps. IEnSF-L1 takes \(N_t\) updates.

Each run uses 1,000 assimilation cycles. Quality metrics use cycles 950--999; analysis time uses all cycles. Tables~\ref{tab:ks_representative}--\ref{tab:ks_ensemble_full} report means and sample SDs across five seeds. Analysis timing follows Appendix~\ref{app:experimental_protocols} and includes IEnSF-L1's terminal preparation, local covariance construction and reverse sampling.

\begin{table}[htbp]
\centering\scriptsize
\caption{Representative Kuramoto--Sivashinsky results with 20 members. Entries are means $\pm$ sample standard deviations across seeds 30--34. Quality metrics use cycles 950--999; analysis time uses all 1,000 cycles. AECSF uses $\epsilon_b=0.015$. Steps denote $T$ for AECSF, EnSF and EnFF, and $N_t$ for IEnSF-L1.}
\label{tab:ks_representative}
\setlength{\tabcolsep}{2.2pt}
\begin{tabular}{lrrccccc}
\toprule
Method & $J$ & Steps & RMSE & CRPS & Energy & SSR & s/cycle\\
\midrule
AECSF & 20 & 20 & $0.0889\pm0.0018$ & $0.0625\pm0.0005$ & $0.0792\pm0.0007$ & $2.1560\pm0.0417$ & $0.0988\pm0.0040$\\
AECSF & 20 & 100 & $0.0601\pm0.0040$ & $0.0431\pm0.0012$ & $0.0545\pm0.0017$ & $2.2261\pm0.1376$ & $0.4681\pm0.0188$\\
EnSF & 20 & 20 & $0.4466\pm0.0047$ & $0.2722\pm0.0027$ & $0.3673\pm0.0038$ & $1.8207\pm0.0106$ & $0.0297\pm0.0006$\\
EnSF & 20 & 100 & $0.1206\pm0.0006$ & $0.0684\pm0.0003$ & $0.0876\pm0.0004$ & $1.0333\pm0.0075$ & $0.1480\pm0.0024$\\
EnFF-F2P & 20 & 20 & $0.2096\pm0.0014$ & $0.1569\pm0.0013$ & $0.2076\pm0.0014$ & $0.0135\pm0.0006$ & $0.0237\pm0.0004$\\
EnFF-F2P & 20 & 100 & $0.2077\pm0.0014$ & $0.1557\pm0.0013$ & $0.2058\pm0.0014$ & $0.0138\pm0.0006$ & $0.1147\pm0.0034$\\
IEnSF-L1 & 20 & 20 & $0.0740\pm0.0006$ & $0.0665\pm0.0001$ & $0.0833\pm0.0002$ & $3.1173\pm0.0258$ & $0.4519\pm0.0115$\\
IEnSF-L1 & 20 & 100 & $0.0498\pm0.0026$ & $0.0345\pm0.0009$ & $0.0433\pm0.0011$ & $2.0622\pm0.1020$ & $2.2554\pm0.0393$\\
EnFF-OT & 20 & 20 & $0.1759\pm0.0012$ & $0.1346\pm0.0009$ & $0.1732\pm0.0012$ & $0.0228\pm0.0008$ & $0.0232\pm0.0005$\\
EnFF-OT & 20 & 100 & $0.1237\pm0.0009$ & $0.0955\pm0.0008$ & $0.1205\pm0.0009$ & $0.0388\pm0.0011$ & $0.1150\pm0.0033$\\
\bottomrule
\end{tabular}
\end{table}

\begin{table}[htbp]
\centering\scriptsize
\caption{Complete Kuramoto--Sivashinsky sampling-budget comparison at $J=20$. Parameter settings, statistical summaries and step notation follow Table~\ref{tab:ks_representative}.}
\label{tab:ks_schedule_full}
\setlength{\tabcolsep}{2.2pt}
\begin{tabular}{lrrccccc}
\toprule
Method & $J$ & Steps & RMSE & CRPS & Energy & SSR & s/cycle\\
\midrule
AECSF & 20 & 20 & $0.0889\pm0.0018$ & $0.0625\pm0.0005$ & $0.0792\pm0.0007$ & $2.1560\pm0.0417$ & $0.0988\pm0.0040$\\
AECSF & 20 & 50 & $0.0638\pm0.0016$ & $0.0467\pm0.0005$ & $0.0591\pm0.0006$ & $2.3116\pm0.0612$ & $0.2402\pm0.0086$\\
AECSF & 20 & 100 & $0.0601\pm0.0040$ & $0.0431\pm0.0012$ & $0.0545\pm0.0017$ & $2.2261\pm0.1376$ & $0.4681\pm0.0188$\\
AECSF & 20 & 200 & $0.0598\pm0.0035$ & $0.0418\pm0.0010$ & $0.0530\pm0.0015$ & $2.1292\pm0.1276$ & $0.9723\pm0.0179$\\
AECSF & 20 & 500 & $0.0579\pm0.0027$ & $0.0405\pm0.0008$ & $0.0514\pm0.0011$ & $2.1330\pm0.0982$ & $2.3375\pm0.0765$\\
EnSF & 20 & 20 & $0.4466\pm0.0047$ & $0.2722\pm0.0027$ & $0.3673\pm0.0038$ & $1.8207\pm0.0106$ & $0.0297\pm0.0006$\\
EnSF & 20 & 50 & $0.1377\pm0.0010$ & $0.0869\pm0.0003$ & $0.1057\pm0.0004$ & $1.4840\pm0.0200$ & $0.0738\pm0.0022$\\
EnSF & 20 & 100 & $0.1206\pm0.0006$ & $0.0684\pm0.0003$ & $0.0876\pm0.0004$ & $1.0333\pm0.0075$ & $0.1480\pm0.0024$\\
EnSF & 20 & 200 & $0.1165\pm0.0014$ & $0.0660\pm0.0008$ & $0.0845\pm0.0010$ & $0.9588\pm0.0079$ & $0.2964\pm0.0054$\\
EnSF & 20 & 500 & $0.1155\pm0.0006$ & $0.0655\pm0.0003$ & $0.0837\pm0.0005$ & $0.9188\pm0.0139$ & $0.7333\pm0.0134$\\
EnFF-F2P & 20 & 20 & $0.2096\pm0.0014$ & $0.1569\pm0.0013$ & $0.2076\pm0.0014$ & $0.0135\pm0.0006$ & $0.0237\pm0.0004$\\
EnFF-F2P & 20 & 50 & $0.2082\pm0.0014$ & $0.1560\pm0.0013$ & $0.2062\pm0.0014$ & $0.0137\pm0.0006$ & $0.0578\pm0.0012$\\
EnFF-F2P & 20 & 100 & $0.2077\pm0.0014$ & $0.1557\pm0.0013$ & $0.2058\pm0.0014$ & $0.0138\pm0.0006$ & $0.1147\pm0.0034$\\
EnFF-F2P & 20 & 200 & $0.2075\pm0.0014$ & $0.1556\pm0.0013$ & $0.2056\pm0.0014$ & $0.0139\pm0.0006$ & $0.2299\pm0.0060$\\
EnFF-F2P & 20 & 500 & $0.2074\pm0.0014$ & $0.1555\pm0.0013$ & $0.2054\pm0.0014$ & $0.0139\pm0.0006$ & $0.5785\pm0.0123$\\
IEnSF-L1 & 20 & 20 & $0.0740\pm0.0006$ & $0.0665\pm0.0001$ & $0.0833\pm0.0002$ & $3.1173\pm0.0258$ & $0.4519\pm0.0115$\\
IEnSF-L1 & 20 & 50 & $0.0604\pm0.0014$ & $0.0453\pm0.0005$ & $0.0567\pm0.0005$ & $2.3726\pm0.0541$ & $1.1257\pm0.0239$\\
IEnSF-L1 & 20 & 100 & $0.0498\pm0.0026$ & $0.0345\pm0.0009$ & $0.0433\pm0.0011$ & $2.0622\pm0.1020$ & $2.2554\pm0.0393$\\
IEnSF-L1 & 20 & 200 & $0.0362\pm0.0023$ & $0.0250\pm0.0008$ & $0.0317\pm0.0010$ & $2.0777\pm0.1289$ & $4.4251\pm0.1107$\\
IEnSF-L1 & 20 & 500 & $0.0336\pm0.0037$ & $0.0205\pm0.0017$ & $0.0264\pm0.0020$ & $1.6052\pm0.1626$ & $11.1443\pm0.2650$\\
EnFF-OT & 20 & 20 & $0.1759\pm0.0012$ & $0.1346\pm0.0009$ & $0.1732\pm0.0012$ & $0.0228\pm0.0008$ & $0.0232\pm0.0005$\\
EnFF-OT & 20 & 50 & $0.1338\pm0.0009$ & $0.1034\pm0.0008$ & $0.1307\pm0.0010$ & $0.0340\pm0.0011$ & $0.0571\pm0.0004$\\
EnFF-OT & 20 & 100 & $0.1237\pm0.0009$ & $0.0955\pm0.0008$ & $0.1205\pm0.0009$ & $0.0388\pm0.0011$ & $0.1150\pm0.0033$\\
EnFF-OT & 20 & 200 & $0.1195\pm0.0009$ & $0.0922\pm0.0008$ & $0.1162\pm0.0009$ & $0.0411\pm0.0012$ & $0.2225\pm0.0058$\\
EnFF-OT & 20 & 500 & $0.1173\pm0.0009$ & $0.0904\pm0.0008$ & $0.1140\pm0.0009$ & $0.0425\pm0.0012$ & $0.5704\pm0.0182$\\
\bottomrule
\end{tabular}
\end{table}

\begin{table}[htbp]
\centering\scriptsize
\caption{Kuramoto--Sivashinsky ensemble-size comparison at 20 sampling steps. The shared $J=20$ settings are reported in Table~\ref{tab:ks_schedule_full}. Parameter settings, statistical summaries and step notation follow Table~\ref{tab:ks_representative}.}
\label{tab:ks_ensemble_full}
\setlength{\tabcolsep}{2.2pt}
\begin{tabular}{lrrccccc}
\toprule
Method & $J$ & Steps & RMSE & CRPS & Energy & SSR & s/cycle\\
\midrule
AECSF & 10 & 20 & $0.1155\pm0.0047$ & $0.0738\pm0.0017$ & $0.0942\pm0.0025$ & $1.6032\pm0.0687$ & $0.0992\pm0.0032$\\
AECSF & 40 & 20 & $0.0732\pm0.0012$ & $0.0571\pm0.0003$ & $0.0722\pm0.0004$ & $2.6609\pm0.0429$ & $0.0998\pm0.0033$\\
AECSF & 80 & 20 & $0.0650\pm0.0012$ & $0.0546\pm0.0003$ & $0.0691\pm0.0004$ & $3.0259\pm0.0558$ & $0.1034\pm0.0033$\\
AECSF & 160 & 20 & $0.0593\pm0.0010$ & $0.0531\pm0.0002$ & $0.0671\pm0.0003$ & $3.3327\pm0.0530$ & $0.1038\pm0.0031$\\
EnSF & 10 & 20 & $0.4867\pm0.0056$ & $0.2943\pm0.0035$ & $0.3982\pm0.0036$ & $1.6259\pm0.0182$ & $0.0297\pm0.0006$\\
EnSF & 40 & 20 & $0.4259\pm0.0042$ & $0.2625\pm0.0030$ & $0.3525\pm0.0032$ & $1.9366\pm0.0116$ & $0.0384\pm0.0007$\\
EnSF & 80 & 20 & $0.4138\pm0.0031$ & $0.2559\pm0.0016$ & $0.3443\pm0.0023$ & $2.0062\pm0.0240$ & $0.0543\pm0.0010$\\
EnSF & 160 & 20 & $0.4059\pm0.0039$ & $0.2518\pm0.0020$ & $0.3397\pm0.0030$ & $2.0536\pm0.0177$ & $0.0867\pm0.0020$\\
EnFF-F2P & 10 & 20 & $0.2096\pm0.0014$ & $0.1569\pm0.0013$ & $0.2077\pm0.0014$ & $0.0131\pm0.0006$ & $0.0238\pm0.0007$\\
EnFF-F2P & 40 & 20 & $0.2095\pm0.0014$ & $0.1569\pm0.0013$ & $0.2076\pm0.0014$ & $0.0137\pm0.0007$ & $0.0326\pm0.0003$\\
EnFF-F2P & 80 & 20 & $0.2095\pm0.0014$ & $0.1568\pm0.0013$ & $0.2075\pm0.0014$ & $0.0139\pm0.0007$ & $0.0508\pm0.0007$\\
EnFF-F2P & 160 & 20 & $0.2095\pm0.0014$ & $0.1568\pm0.0013$ & $0.2075\pm0.0014$ & $0.0139\pm0.0007$ & $0.0870\pm0.0015$\\
IEnSF-L1 & 10 & 20 & $0.0965\pm0.0008$ & $0.0749\pm0.0002$ & $0.0938\pm0.0003$ & $2.3223\pm0.0183$ & $0.2926\pm0.0065$\\
IEnSF-L1 & 40 & 20 & $0.0585\pm0.0002$ & $0.0619\pm0.0000$ & $0.0777\pm0.0001$ & $3.9905\pm0.0171$ & $1.2187\pm0.0101$\\
IEnSF-L1 & 80 & 20 & $0.0504\pm0.0004$ & $0.0600\pm0.0001$ & $0.0752\pm0.0001$ & $4.6569\pm0.0339$ & $4.2181\pm0.0229$\\
EnFF-OT & 10 & 20 & $0.1759\pm0.0012$ & $0.1347\pm0.0009$ & $0.1733\pm0.0012$ & $0.0222\pm0.0007$ & $0.0232\pm0.0005$\\
EnFF-OT & 40 & 20 & $0.1759\pm0.0012$ & $0.1345\pm0.0009$ & $0.1731\pm0.0012$ & $0.0231\pm0.0008$ & $0.0314\pm0.0005$\\
EnFF-OT & 80 & 20 & $0.1759\pm0.0012$ & $0.1345\pm0.0009$ & $0.1731\pm0.0012$ & $0.0233\pm0.0008$ & $0.0479\pm0.0017$\\
EnFF-OT & 160 & 20 & $0.1759\pm0.0012$ & $0.1345\pm0.0009$ & $0.1730\pm0.0012$ & $0.0234\pm0.0008$ & $0.0805\pm0.0013$\\
\bottomrule
\end{tabular}
\end{table}

Energy is reported as a supplementary metric, using
\[
 d^{-1/2}\left[\frac{1}{J}\sum_j\|x_j-x^\star\|_2
 -\frac{1}{2J^2}\sum_{j,k}\|x_j-x_k\|_2\right],\qquad d=1024.
\]
The pair sum includes its zero diagonal entries.

\begin{table}[t]
\centering
\caption{KS forecasts at a horizon of 8: means $\pm$ sample SDs across ten paired
runs. Parentheses: ensemble size/$T$ ($N_t$ for IEnSF-L1); AECSF uses $M=J$.
Coverage: central 90\% interval.}
\label{tab:ks_forecast_comparable_cost}
\small
\begin{tabular}{lccccc}
\toprule
Configuration & RMSE $\downarrow$ & CRPS $\downarrow$ & Cost (s) & SSR & Coverage \\
\midrule
AECSF (160,100) & $\mathbf{0.1180}\pm0.0078$ & $\mathbf{0.0640}\pm0.0016$ & 0.5561 & 2.24 & 99.83\% \\
IEnSF-L1 (20,20) & $0.2536\pm0.0177$ & $0.1283\pm0.0058$ & 0.6648 & 1.73 & 97.76\% \\
EnSF (160,100) & $0.2816\pm0.0253$ & $0.1265\pm0.0096$ & 0.5135 & 1.00 & 89.34\% \\
EnFF-OT (160,100) & $0.2665\pm0.0231$ & $0.1627\pm0.0134$ & 0.4771 & 0.04 & 5.26\% \\
EnFF-F2P (160,100) & $0.4323\pm0.0395$ & $0.2729\pm0.0237$ & 0.5097 & 0.02 & 1.78\% \\
\bottomrule
\end{tabular}
\end{table}

\paragraph{Free forecasts after KS assimilation.}
Forecast lead is measured in physical time. We start free forecasts from the complete
analysis ensembles at cycle indices 950, 955, 960, 965, and 970. We average
forecast metrics over these five origins within each run and report means and
sample SDs across ten paired runs.
Forecast accuracy is evaluated through lead 24; the table summarizes
lead 8. The reported cost combines one assimilation cycle and the
complete ensemble forecast to lead 8, measured on an RTX 3090 with
initialization amortized.
In Table~\ref{tab:ks_forecast_comparable_cost}, the two configuration
numbers specify ensemble size and sampler setting: \(T\) score evaluations
for AECSF and \(T-1\) updates for EnSF/EnFF, versus \(N_t\) updates for
IEnSF-L1. AECSF uses \(M=J\). The dashed IEnSF-L1 curve in
Figure~\ref{fig:ks_forecast_comparable_cost} uses 100 updates and is a
higher-cost reference, distinct from the 20-update configuration in the
table.

\begin{figure}[t]
\centering
\includegraphics[width=\textwidth]{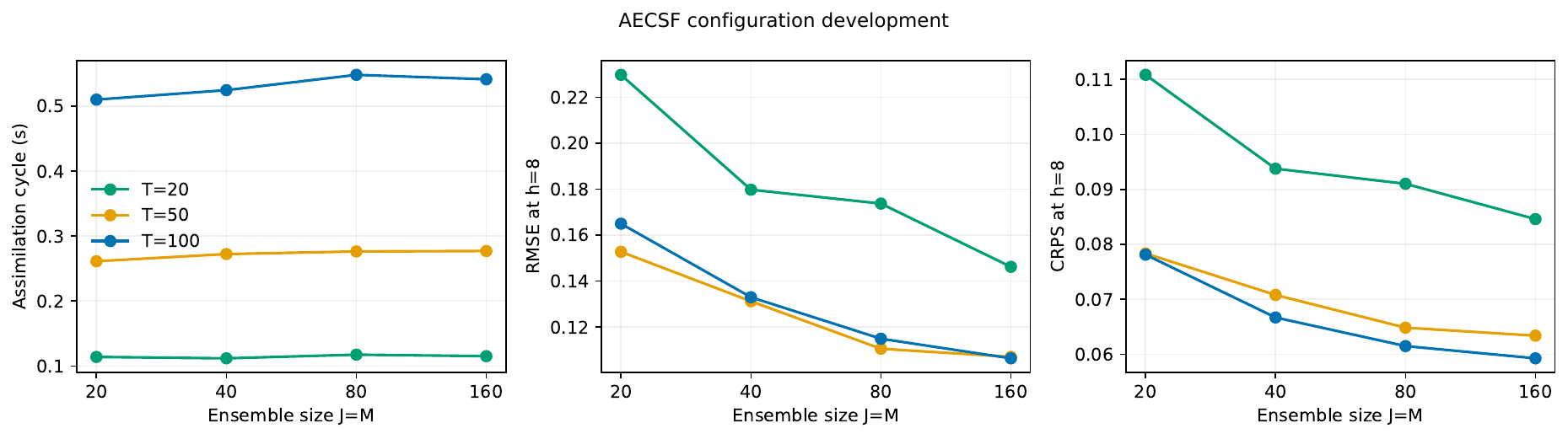}
\caption{AECSF configuration development: cost per assimilation cycle,
and RMSE and CRPS at a forecast horizon of 8, for different ensemble
sizes and three sampler settings.}
\label{fig:ks_forecast_development}
\end{figure}

\clearpage
\subsection{Learned forecast prior for sparse Navier--Stokes assimilation}
\label{app:ns_learned_thirty}

\subsubsection{From the conditional flow to the forecast prior}
\label{app:ns_flow_training}

We observe velocity at only a few locations. To estimate vorticity away from the sensors,
we use a forecast prior that captures how vorticity at different locations varies together. We learn those
dependencies with a conditional normalizing flow. Given a forecast ensemble, the flow models
a joint distribution over the full vorticity field. The training data pair simulated forecast
ensembles with full fields and are generated independently of the assimilation runs. During
assimilation, each new forecast ensemble supplies the context for the flow, whose parameters
are held fixed. The current sensor measurements enter through the observation likelihood.

We represent the zero-mean vorticity field on the \(128\times128\) periodic grid by 16,383
orthonormal Haar coefficients. Let \(c\) denote the features of a given forecast
ensemble. At cycle \(n\), \(c=c_n\) summarizes the ensemble \(\mathcal X_n^f\).
It includes the means and scales of its Haar coefficients, products of anomalies
at neighbouring locations, and features averaged over the encoded ensemble members.
With \(c\) held fixed, the decoder \(z=\mathcal D_\theta(v;c)\) maps a standard
Gaussian input \(v\) to a field in Haar coordinates using the forecast mean and scale.
For the same \(c\), the inverse \(v=\mathcal D_\theta^{-1}(z;c)\) recovers the
input that generated a given field. The change of variables expresses the flow density in terms of the
standard Gaussian density \(\varphi\) and the Jacobian of the inverse map.
We obtain the score of the flow component by differentiating \(\log p_\theta(z\mid c)\)
with respect to the state coordinates \(z\).

\begin{equation}
\begin{aligned}
p_\theta(z\mid c)
  &=\varphi(v)\left|\det D_z\mathcal D_\theta^{-1}(z;c)\right|,\\
\nabla_z\log p_\theta(z\mid c)
  &=-\big[D_z\mathcal D_\theta^{-1}(z;c)\big]^\top v
    +\nabla_z\log\left|\det D_z\mathcal D_\theta^{-1}(z;c)\right|.
\end{aligned}
\label{eq:ns_flow_density_score}
\end{equation}

We fit the flow density in \eqref{eq:ns_flow_density_score} offline by minimizing
the per-coordinate negative log likelihood. During assimilation the decoder stays
fixed. The current ensemble also determines how much
the decoder inputs vary and how they vary together. At training time the
decoder input \(v\) has a standard Gaussian base distribution. At analysis
time we draw \(u\sim\N(0,\I)\) and set
\(v=(D_n^{\rm sc})^{-1}B_n^{\rm OAS}u\), using the ensemble's oracle
approximating shrinkage (OAS) covariance factor and coordinate scales.
The same decoder maps this transformed input to Haar coefficients, which
we convert to the vorticity field:

\[
g_\theta(u;c_n)=
\mathcal W^{-1}\mathcal D_\theta\bigl((D_n^{\rm sc})^{-1}B_n^{\rm OAS}u;c_n\bigr).
\]

Here \(D_n^{\rm sc}\) is diagonal, and the shrinkage covariance includes a fixed
calibration factor of \(33/29\). We allow for additional uncertainty at large spatial
scales by adding an independent low-frequency Gaussian field. The forecast prior
is the distribution of the resulting vorticity field,

\begin{equation}
\begin{aligned}
\omega(u,b)&=g_\theta(u;c_n)+FQ_n^{1/2}b,\\
u&\sim\N(0,\I_{16383}),\qquad b\sim\N(0,\I_{192}).
\end{aligned}
\label{eq:ns_forecast_prior}
\end{equation}

The inputs \(u\) and \(b\) are independent. The columns of \(F\) span the real Fourier modes
with \(0<|k|<8\), each normalized to unit spatial mean square. The low-frequency covariance
\(Q_n\) is \hyperref[app:ns_low_frequency_covariance]{estimated from past innovations}.
It stays fixed during the analysis, as do \(c_n\) and \(B_n^{\rm OAS}\).

\subsubsection{AECSF analysis with the learned prior}
\label{app:ns_analysis}

Equation~\eqref{eq:ns_forecast_prior} represents each proposed vorticity field as
\(\omega(u,b)\). We update the field by changing its two inputs. The input \(u\)
determines the decoded field, while \(b\) controls an additional low-frequency
perturbation (see \hyperref[app:ns_low_frequency_covariance]{Appendix~\ref*{app:ns_low_frequency_covariance}}
for how we estimate the covariance \(Q_n\) of its Fourier coefficients). In these coordinates, the prior
is simply the product of two standard Gaussian densities.

To incorporate the observations, we apply \(H\) to \(\omega(u,b)\) to predict
the velocities at the sensors. The likelihood \(\ell_n(u,b)\) compares these
predictions with the measured velocities \(y_n\) using the noise covariance \(R\).
Combining the tempered likelihood with the two Gaussian priors gives the target
in \eqref{eq:main_tempered_proposal} expressed in \(u\) and \(b\):

\begin{equation}
\begin{aligned}
\ell_n(u,b)&:=g_n(y_n\given\omega(u,b))
\propto \exp\!\left[-\tfrac12\|y_n-H\omega(u,b)\|_{R^{-1}}^2\right],\\
\pi_{n,\rho}^{\rm latent}(u,b)&\propto\varphi(u)\varphi(b)\ell_n(u,b)^\rho,
\end{aligned}
\label{eq:ns_latent_target}
\end{equation}

where \(\varphi\) is the standard Gaussian density in the corresponding dimension.
We move proposals both while tempering the likelihood and later during denoising,
when they supply the conditional mean for each reverse particle
(Algorithm~\ref{alg:ns_learned_analysis}).
To update \(u\), we hold \(b\) fixed and differentiate the log target in
\eqref{eq:ns_latent_target} with respect to \(u\). The Gaussian prior contributes
\(-u\). The likelihood \(\ell_n(u,b)=g_n(y_n\mid\omega(u,b))\) contributes the
gradient through the decoded field. Adding these two terms gives the Langevin
step shown after \eqref{eq:rho_tempered_move}:

\begin{equation}
u^+=u+\tfrac h2\bigl[-u+\rho\,\nabla_u\log g_n(y_n\mid\omega(u,b))\bigr]
       +\sqrt h\,\xi,\qquad \xi\sim\N(0,\I_{\dim u}).
\label{eq:ns_latent_move}
\end{equation}

Each movement block uses 16 isotropic unadjusted Langevin steps with \(h=0.002\),
so the move in \(u\) has finite-step discretization error. Once \(u\) is fixed,
the decoded field \(g_\theta(u;c_n)\) is known. The predicted sensor velocities
then depend linearly on \(b\) through \(B=HFQ_n^{1/2}\). With a Gaussian prior
on \(b\) and Gaussian observation noise, we can draw \(b\) exactly from its
conditional distribution at temperature \(\rho\). Its covariance and mean are

\begin{equation}
\Sigma_\rho=(I+\rho B^\top R^{-1}B)^{-1},\qquad
\mu_\rho(u)=\Sigma_\rho\,\rho B^\top R^{-1}\{y_n-Hg_\theta(u;c_n)\}.
\label{eq:ns_low_frequency_draw}
\end{equation}

Before reverse sampling begins, we temper the likelihood for the \(M=128\) proposals.
A Gaussian reference based on the shrinkage covariance determines the temperature
schedule. We allow at most 256 levels and choose each temperature increment to
target a conditional effective sample size (ESS) of \(0.8M\) under this reference.
At each level we update the proposal weights with the incremental likelihood.
If the reference ESS falls below \(0.5M\), we draw stratified ancestors using the
updated proposal weights and reset the descendant weights to \(1/M\). We move the
proposals under their current tempered target at every level.

During denoising, the likelihood is fully incorporated (\(\rho=1\)). On reverse
steps \(k=2,4,\ldots,100\), the current reverse particles guide ancestor selection
before the proposals move.
Each particle weights the decoded proposals in Haar coordinates using the Gaussian
kernel at its noise level. Equation~\eqref{eq:refresh_resampling} combines these
kernel values with the incoming shared weights and averages the normalized
probabilities across particles. We draw ancestors from this distribution, copy
each selected pair \((u,b)\), and correct its weight by \eqref{eq:ancestor_correction}.

Each selected proposal carries its corrected weight through the move. We express the
resulting fields in Haar coordinates to obtain the proposals \(z_i^m\) in
\eqref{eq:main_ecsf_estimator}. The remaining likelihood factor is now one, so each
reverse particle uses its Gaussian kernel to reweight the proposals and estimate its
conditional mean. Its conditional score \(\widehat s_t(x_t)\) from
\eqref{eq:main_estimated_score} then drives the probability-flow ODE.

The \(J=32\) reverse particles start from independent standard Gaussian samples in Haar
coordinates. We use the noising schedule with \(\varepsilon=10^{-5}\) given by

\[
a_\tau=1-(1-\varepsilon)\tau,\qquad
\sigma_\tau^2=\varepsilon+(1-\varepsilon)\tau.
\]

We integrate the probability-flow ODE from \(\tau=1\) to zero over 100 uniform-\(\tau\)
intervals. Here we use half the logSNR in \eqref{eq:logsnr_mesh},
\(\lambda_\tau^{\rm half}=\tfrac12\lambda(\tau)=\log(a_\tau/\sigma_\tau)\).
The update between adjacent times \(t>s\) is

\begin{equation}
x_s=\frac{a_s}{a_t}x_t+
\sigma_s\sigma_t\bigl[e^{\lambda_s^{\rm half}-\lambda_t^{\rm half}}-1\bigr]\widehat s_t(x_t).
\label{eq:ns_reverse_ode_step}
\end{equation}

\begin{algorithm}[H]
\caption{AECSF analysis with a learned forecast prior for sparse Navier--Stokes observations}
\label{alg:ns_learned_analysis}
\small
\begin{algorithmic}[1]
\Require Forecast ensemble \(\mathcal X_n^f\), observation \(y_n\), observation map \(H\) and noise covariance \(R\), trained flow, and low-frequency covariance \(Q_n\)
\State Condition the flow on \(\mathcal X_n^f\) and set its input covariance from the ensemble.
\State Draw \(M=128\) Gaussian pairs \((u_m,b_m)\) for the shared proposals in \eqref{eq:ns_forecast_prior}. Set \(W_m=1/M\) and \(\rho=0\).
\For{each successive temperature \(\rho'\) up to \(1\) on the Gaussian reference schedule}
    \State Set \(W_m\propto W_m\,\ell_n(u_m,b_m)^{\rho'-\rho}\), normalize over \(m\), and set \(\rho=\rho'\).
    \If{the reference effective sample size falls below \(M/2\)}
        \State Resample the pairs from \(W_m\) by stratified sampling and reset the weights to \(1/M\).
    \EndIf
    \State For each pair, take 16 Langevin steps \eqref{eq:ns_latent_move} in \(u_m\) with \(b_m\) fixed (\(h=0.002\)), then draw \(b_m\) by \eqref{eq:ns_low_frequency_draw}.
\EndFor
\State Draw \(J=32\) reverse particles from standard Gaussian noise in Haar coordinates.
\For{reverse steps \(k=1,\ldots,100\), with \(\rho=1\)}
    \If{\(k\) is even}
        \State Use the current reverse particles, decoded proposals in Haar coordinates, and \(W_m\) to compute \(s_i\) by \eqref{eq:refresh_resampling}.
        \State Draw stratified ancestors from \(s_i\), copy their pairs \((u,b)\), and set \(W_m\) to the normalized corrected weights in \eqref{eq:ancestor_correction}.
        \State Repeat the 16 Langevin steps and the conditional draw for each pair.
    \EndIf
    \State Use the decoded proposals in Haar coordinates and \(W_m\) to compute each particle's conditional mean \eqref{eq:main_ecsf_estimator} and score \eqref{eq:main_estimated_score}.
    \State Advance each reverse particle with its score using the probability-flow ODE step \eqref{eq:ns_reverse_ode_step}.
\EndFor
\State \Return the \(J\) final reverse particles as vorticity fields after the inverse Haar transform.
\end{algorithmic}
\end{algorithm}

\subsubsection{Low-frequency covariance and flow training}
\label{app:ns_low_frequency_covariance}

In a diagnostic run with isotropic Langevin moves, the proposal ensemble gained variance
mainly at middle and high spatial frequencies, with little increase in the low-frequency
modes. To represent additional uncertainty in large-scale vorticity patterns, we add a
low-frequency Gaussian field to each decoded proposal in \eqref{eq:ns_forecast_prior}.
Its Fourier coefficients have covariance \(Q_n\), which determines the variance added in
each mode and how perturbations in different modes vary together. We estimate \(Q_n\)
from the velocity innovations of the preceding four cycles.

To estimate the uncertainty to add in the current cycle, we first infer the
low-frequency error in each of the four preceding cycles from its sensor residuals.
At a completed cycle \(s\), the innovation \(d_s=y_s-H\bar\omega_s^f\) is the difference between
the observed velocities and their forecast mean. We write \(S_s=\widehat
C_s^{\rm obs}+R\), with \(\widehat C_s^{\rm obs}\) a shrinkage estimate of forecast
covariance in observation space. The model below uses \(D=HF\) to relate the low-frequency
coefficients \(\eta_s\) to the innovation:

\[
d_s=D\eta_s+\epsilon_s,\qquad
\eta_s\sim\mathcal N(0,\tau_{\rm LF}I),\qquad
\epsilon_s\sim\mathcal N(0,S_s).
\]

We fit the common variance \(\tau_{\rm LF}\) by marginal negative log likelihood over
\(s=n-4,\ldots,n-1\). For each cycle, \(m_s\) estimates its low-frequency
error and \(P_s\) gives the uncertainty in that estimate. Averaging the
resulting second moments \(P_s+m_sm_s^\top\) gives

\begin{equation}
\begin{aligned}
P_s &=\tau_{\rm LF}(I+\tau_{\rm LF}D^\top S_s^{-1}D)^{-1},
&m_s &=P_sD^\top S_s^{-1}d_s,\\
\widetilde Q_n&=\frac14\sum_{s=n-4}^{n-1}(P_s+m_sm_s^\top).
\end{aligned}
\label{eq:ns_low_frequency_second_moment}
\end{equation}

We limit the added variance using the low-frequency energy of the
initial ensemble, \(c_0=J^{-1}\sum_j\|F^\top\omega_1^{f,j}/128^2\|^2\). We fit
\(\tau_{\rm LF}\) within \([0,c_0/192]\) and set
\(Q_n=\min(1,c_0/\operatorname{tr}\widetilde Q_n)\widetilde Q_n\), with \(Q_n=0\)
when the trace vanishes. For the first four cycles we set \(Q_n=0\). Thereafter,
the four previous cycles determine \(Q_n\) before we assimilate \(y_n\).
We use the true field only for evaluation.

\paragraph{Flow architecture and offline training}

The decoder works from coarse to fine over seven Haar scales. Each scale uses
two affine coupling layers of convolutional width 32, conditioned on the coarser
fields already decoded. The forecast features \(c_n\) include Haar coefficient
means and scales, second moments of standardized anomalies, and products of
horizontally and vertically neighbouring anomalies. They also include the mean
of the features obtained by applying a shared encoder to each ensemble member.

Each of the 128 training groups and 32 development groups pairs a simulated
32-member forecast ensemble with eight independent target fields. We use AdamW
with a batch size of 32 for 1,000 updates and select the checkpoint by development
negative log likelihood.

\subsubsection{Navier--Stokes experiment and results}
\label{app:ns_experiment_results}

We test how well the filters recover the vorticity field from sparse velocity
measurements. The velocity field on the periodic domain \([0,2\pi)^2\) satisfies

\[
\partial_t\mathbf v+(\mathbf v\cdot\nabla)\mathbf v
=-\nabla p+0.02\,\Delta\mathbf v+(\sin 4y,0)-0.1\mathbf v,
\qquad \nabla\cdot\mathbf v=0.
\]

Analyses are separated by 0.2 units of physical time. The observation map \(H\) first
converts vorticity to staggered-grid velocity through the compatible discrete inverse curl.
It then uses periodic bilinear interpolation to evaluate both velocity components at 64 fixed
sensor locations within each trajectory. We use Gaussian observation noise with covariance
\(R=0.01I\).

Within each of four development trajectories, all methods share the truth, observations, and
32-member initial forecast ensemble. Each filter then propagates its own 32-member analysis
ensemble over 50 cycles. The forecast-only baseline propagates the initial ensemble over the
same period with the same dynamics. LETKF uses a localization-radius fraction of 0.25 and
anomaly inflation of 1.3, selected by single-analysis CRPS on four separate tuning groups
from 24 radius--inflation pairs.

We also test how the settings from the Kuramoto--Sivashinsky experiments transfer to sparse
Navier--Stokes assimilation. EnSF, EnFF-OT, and EnFF-F2P use the same implementations and
parameters as in that comparison, with the sparse velocity likelihood and fixed orthonormal
Haar coordinates. EnSF uses \(\epsilon_\alpha=0.5\) and \(\epsilon_\beta=0.025\). EnFF-OT and
EnFF-F2P both use \(\sigma_{\min}=0.01\), with guidance strengths of 1 and 0.2, respectively.
All three methods use 32 members and 100 path points. From the second cycle onward, EnFF-F2P
uses its own preceding analysis ensemble.

\paragraph{Accuracy over 50 cycles}

Figure~\ref{fig:ns_thirty_accuracy} traces velocity RMSE and vorticity CRPS over all 50
cycles. Relative to LETKF, AECSF reduced the full-cycle averages by 32.5\% and 36.8\%,
respectively. In cycles 41--50, AECSF had lower values for both metrics in every cycle of all
four paired trajectories.

\paragraph{Low-frequency error and spread}
The prior adds uncertainty in the Fourier modes \(0<|k|<8\) through \eqref{eq:ns_forecast_prior}.
We therefore project the vorticity fields onto these modes with \(\mathcal P_{<8}\)
and compare the ensemble spread with the error of the ensemble mean at the same scales.
We measure ensemble-mean error by the spatial RMS of \(\mathcal
P_{<8}(\bar\omega_n^a-\omega_n^{\rm true})\). In each window, we average the error over
cycles and define spread as the square root of the pooled ensemble variance in the same
subspace.

\begin{figure}[!htbp]
\centering
\includegraphics[width=\textwidth]{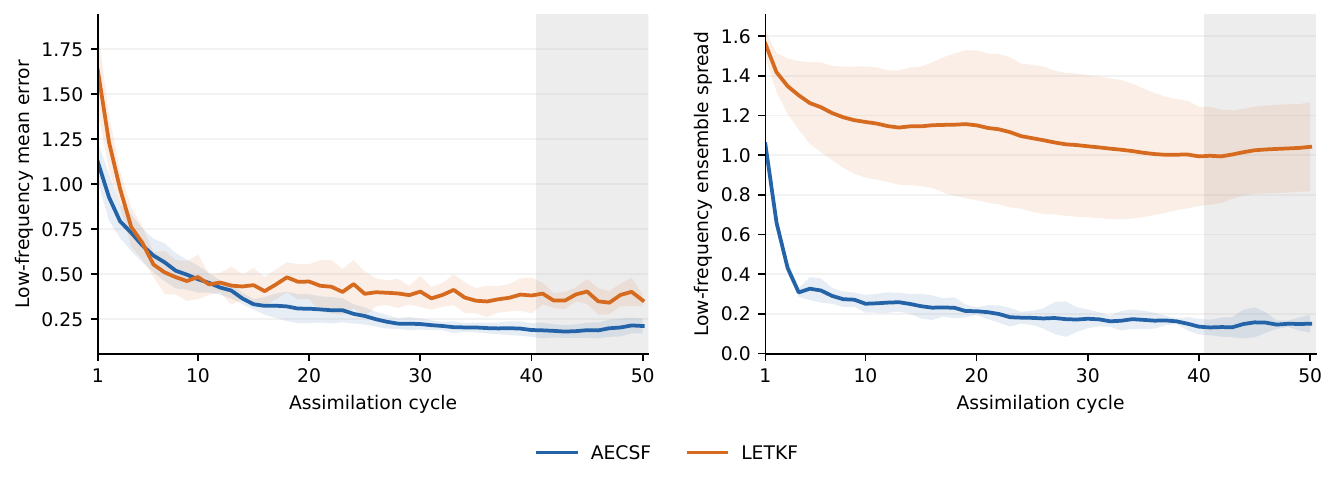}
\caption{Low-frequency ensemble-mean error (left) and spread (right) over 50 cycles for \(0<|k|<8\). Lines and bands show means and sample standard deviations across four trajectories.}
\label{fig:ns_thirty_low}
\end{figure}

Over cycles 41--50, AECSF's low-frequency error was \(0.1936\pm0.0374\), compared with
\(0.3713\pm0.0503\) for LETKF, a 47.9\% reduction. The trajectory-averaged error was lower in
all four pairs. AECSF's low-frequency spread was \(0.1493\pm0.0280\). Its full-state
spread--skill ratios were 1.014, 0.679, 1.004, and 0.744 across the four trajectories, with
two close to one and two underdispersed.
Figure~\ref{fig:ns_thirty_low} shows the low-frequency error and spread over all 50 cycles.

\begin{figure}[!htbp]
\centering
\includegraphics[width=\textwidth]{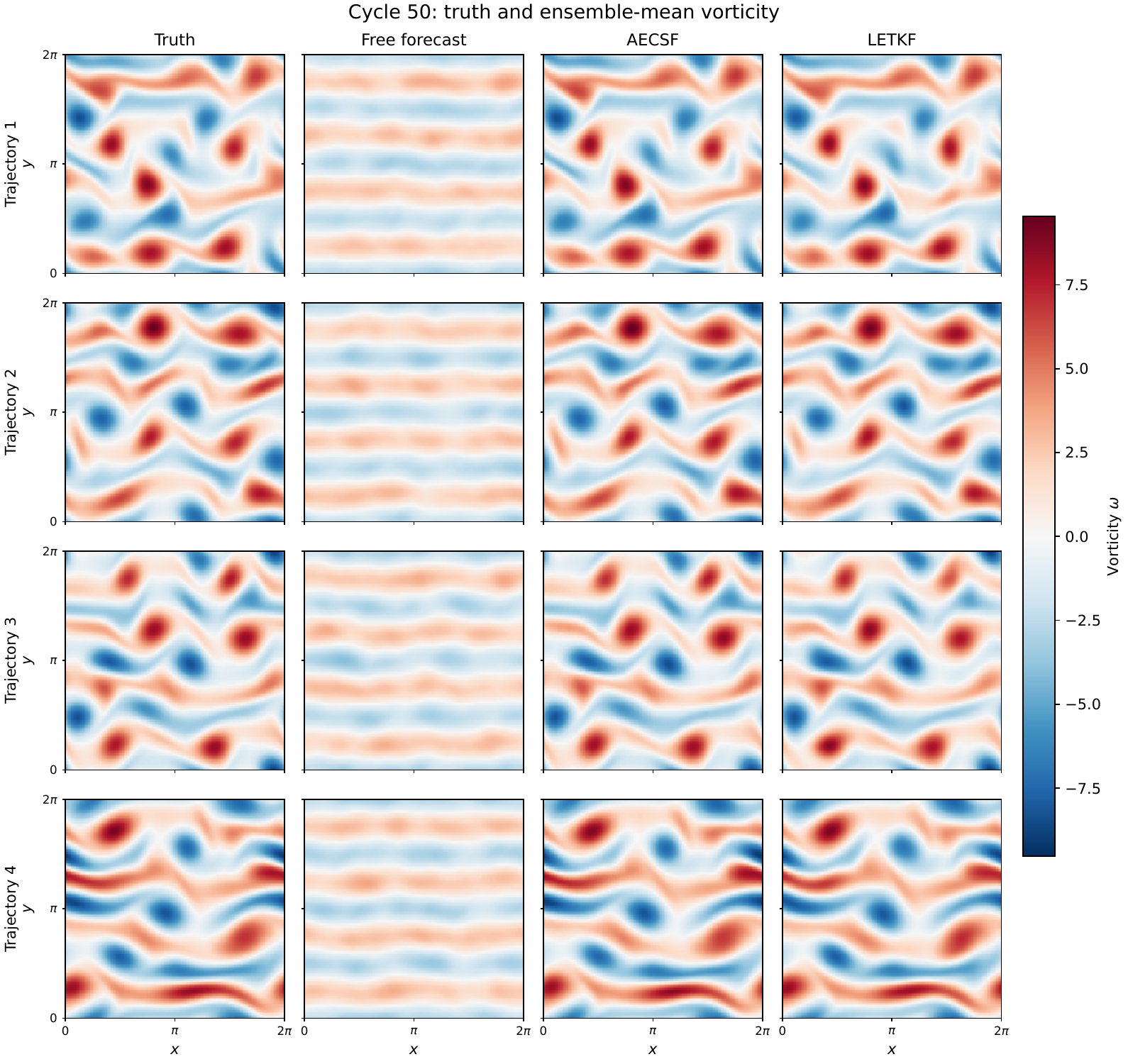}
\caption{Vorticity at cycle 50 for four trajectories (rows). Columns show the truth and the forecast-only, AECSF, and LETKF ensemble means. All panels share a colour scale.}
\label{fig:ns50_vorticity_fields}
\end{figure}

\begin{figure}[!htbp]
\centering
\includegraphics[width=\textwidth]{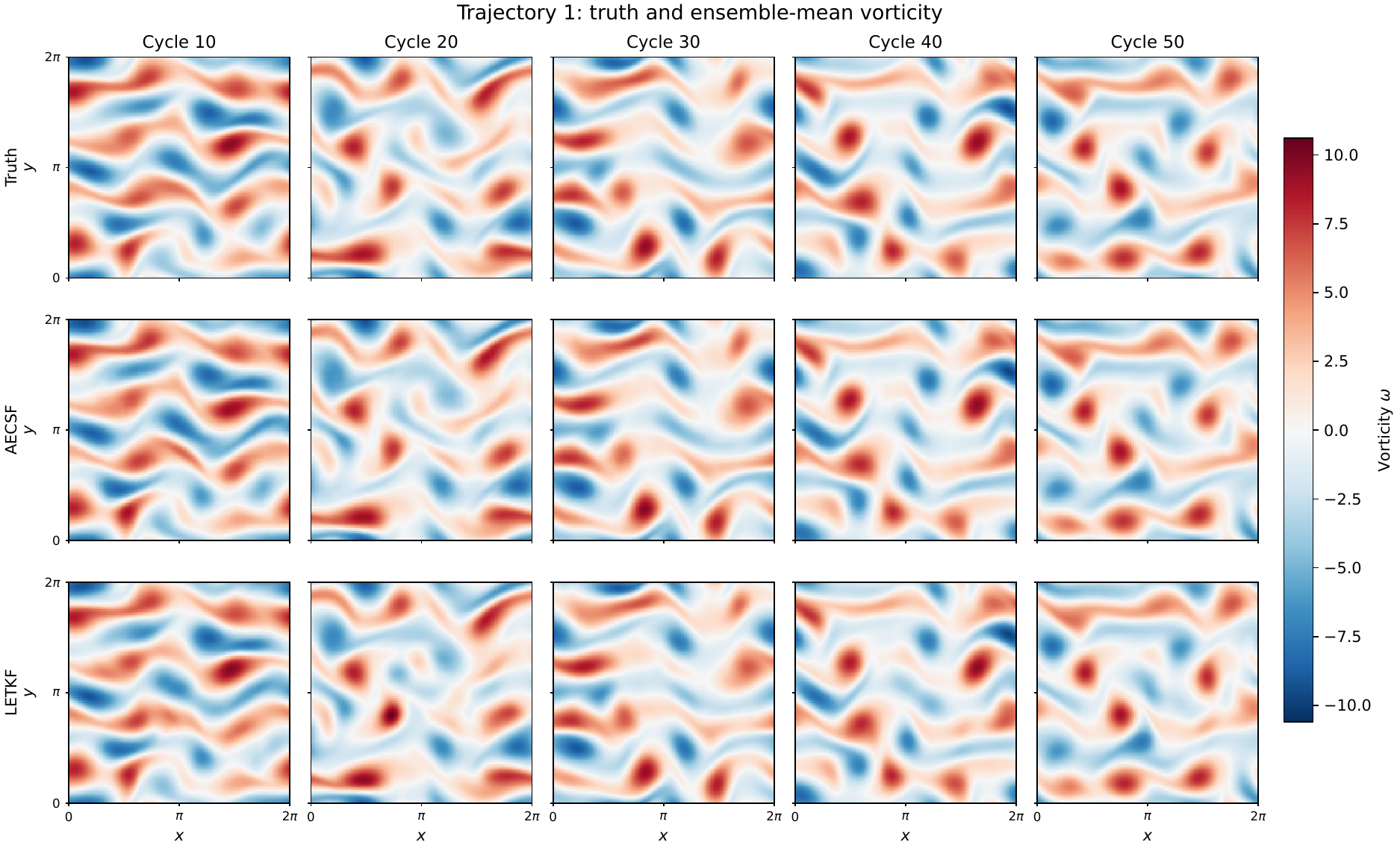}
\caption{Vorticity evolution along the first trajectory. Columns show cycles 10, 20, 30, 40, and 50. Rows show the truth and the AECSF and LETKF ensemble means. All panels share a colour scale.}
\label{fig:ns50_vorticity_evolution}
\end{figure}

\begin{figure}[!htbp]
\centering
\includegraphics[width=\textwidth]{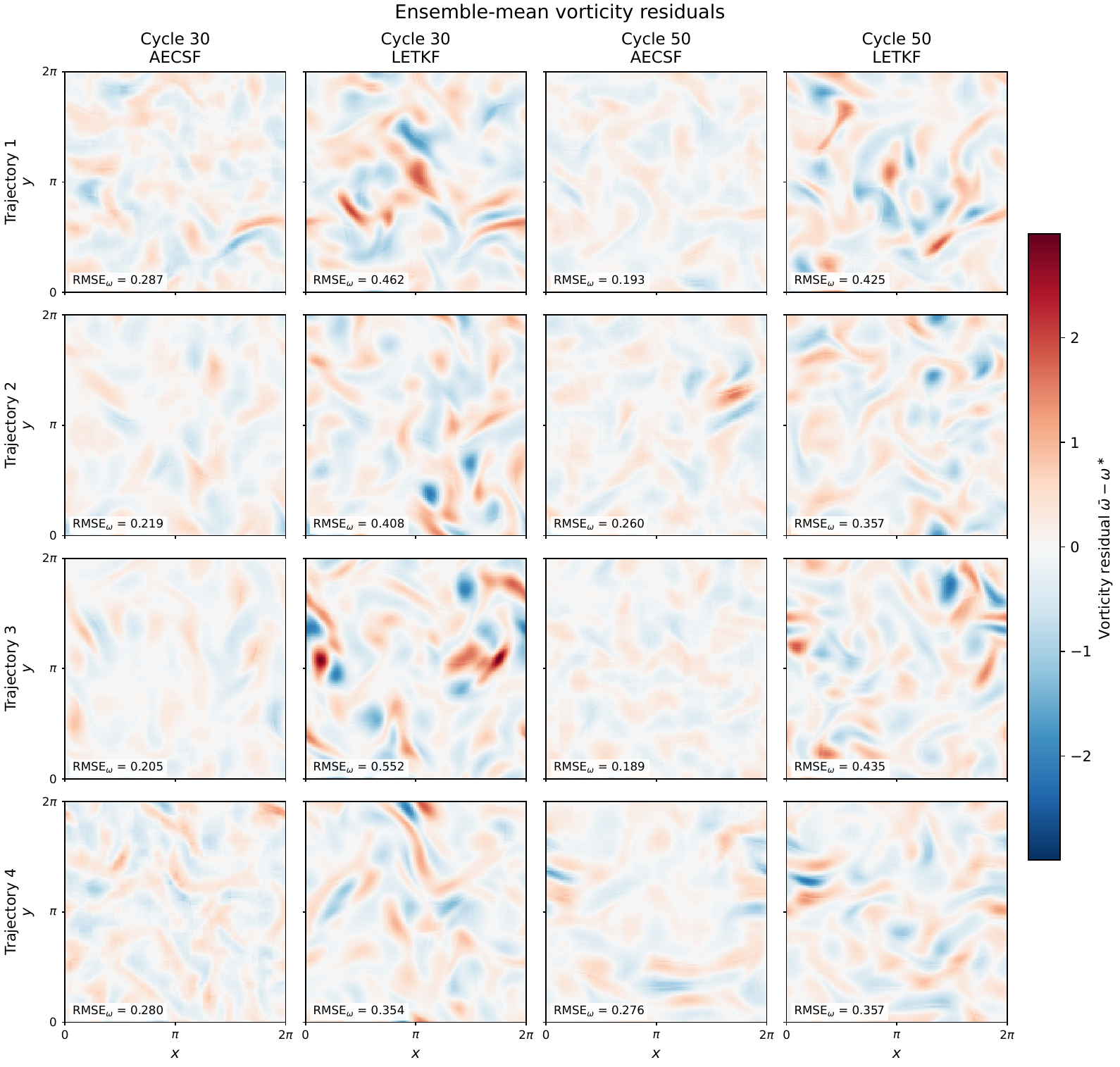}
\caption{Ensemble-mean vorticity minus truth for four trajectories (rows). The first two columns show AECSF and LETKF at cycle 30, and the last two show them at cycle 50. All panels share a colour scale. Inset values give vorticity RMSE.}
\label{fig:ns50_vorticity_residuals}
\end{figure}

\clearpage
\paragraph{Spatial reconstruction}
Figures~\ref{fig:ns50_vorticity_fields}--\ref{fig:ns50_vorticity_residuals} compare the
reconstructed vorticity fields, their evolution, and their errors. Both filters recovered the
main flow structures missing from the forecast-only ensemble mean. Across the four
trajectories at cycles 30 and 50, AECSF had lower vorticity RMSE than LETKF in all eight
comparisons.

\clearpage

\clearpage

\end{document}